\documentclass[10pt,twocolumn,letterpaper]{article}

\usepackage{cvpr}              % To produce the CAMERA-READY version
\usepackage{graphicx}
\usepackage{amsmath}
\usepackage{amssymb}
\usepackage{booktabs}
\usepackage{multirow}
\usepackage{xcolor}

\usepackage[pagebackref,breaklinks,colorlinks]{hyperref}

\usepackage[capitalize]{cleveref}
\crefname{section}{Sec.}{Secs.}
\Crefname{section}{Section}{Sections}
\Crefname{table}{Table}{Tables}
\crefname{table}{Tab.}{Tabs.}

\def\cvprPaperID{3658} % *** Enter the CVPR Paper ID here
\def\confName{CVPR}
\def\confYear{2022}

\newcommand{\myparagraph}[1]{\vspace{3pt}\noindent\textbf{#1.}}
\newcommand{\suppmat}{Appendix}

\begin{document}

\title{Deep Visual Geo-localization Benchmark}

\author{%
  \textbf{Gabriele Berton} \\
  Politecnico di Torino\\
  \and
  \textbf{Riccardo Mereu}\\
  Politecnico di Torino\\
  \and
  \textbf{Gabriele Trivigno}\\
  Politecnico di Torino\\
  \and
  \textbf{Carlo Masone}\\
  CINI
  %Consorzio Interuniversitario Nazionale per l'Informatica\\
  \and
  \textbf{Gabriela Csurka}\\
  NAVER LABS Europe\\
  \and
  \textbf{Torsten Sattler}\\
  %Czech Institute of Informatics,\\ Robotics and Cybernetics, \\
  CIIRC, Czech Technical \\ 
  University in Prague
  \and
  \textbf{Barbara Caputo}\\
  Politecnico di Torino\\
}

%% omit the following lines up until the closing ``}''.
%% Additional authors and addresses can be added with ``\and'',
%% just like the second author.
%% To save space, use either the email address or home page, not both
%\and
\maketitle

%%%%%%%%% ABSTRACT
\begin{abstract}
    In this paper, we propose a new open-source benchmarking framework for Visual Geo-localization (VG) that allows to build, train, and test a wide range of commonly used architectures, with the flexibility to change individual components of a geo-localization pipeline.
    The purpose of this framework is twofold: i) gaining insights into how different components and design choices in a VG pipeline impact the final results, both in terms of performance (recall@N metric) 
    and system requirements (such as execution time and memory consumption);
    ii) establish a systematic evaluation protocol for comparing different methods.
    Using the proposed framework, we perform a large suite of experiments which provide criteria for choosing backbone, aggregation and negative mining depending on the use-case and requirements. We also assess the impact of engineering techniques like pre/post-processing, data augmentation and image resizing, showing that better performance can be obtained through somewhat simple procedures: for example, downscaling the images' resolution to 80\% can lead to similar results with a 36\% savings in extraction time and dataset storage requirement.
    Code and trained models are available at {\small{\url{https://github.com/gmberton/deep-visual-geo-localization-benchmark}}}.
\end{abstract}
%%%%%%%%% BODY TEXT
\section{Introduction}
\label{sec:intro}
The task of coarsely estimating the place where a photo was taken based on a set of previously visited locations is called \emph{Visual (Image) Geo-localization} (VG) \cite{Kim-2017,Liu-2019,Zamir-2021} or \emph{Visual Place Recognition} (VPR) \cite{Lowry-2016, Garg-2021} and it is addressed  using image matching and retrieval methods on a database of images of known locations. 
We are witnessing a rapid growth of this field of research,
as demonstrated by the increasing number of publications \cite{Torii-2015, Lowry-2016, Kim-2017, Arandjelovic-2018, Piasco-2018, Torii-2018, Liu-2019, Radenovic-2019, Khaliq-2020, Ge-2020, Warburg-2020, Hausler-2021, Zhang-2021, Masone-2021, Berton-2021, Hong_2019_ICCV, Garg-2021-seqNet, Vysotska_2019E_seqMaps, Garg-2020, Wang-2019, Ward-2019, xin-17, chen-17}, but this expansion is accompanied by two major limitations:\\
i) \textbf{A focus on single metric optimization}, as it is common practice to compare results solely based on the recall on chosen datasets and ignoring other factors such as execution time, hardware requirements, and scalability.
All these aspects are important constraints in the design of a real-world VG system. 
For instance, one might gladly accept a 5\% drop in accuracy if this leads to a 90\% decrease of descriptors size as the resulting reduction in memory requirements enables a better scalability. 
Similarly, computational time and descriptor dimensionality are crucial constraints in real-time applications, given a target hardware platform.\\
ii) \textbf{A lack of a standardized framework} to train and test VG models. It is common practice to perform direct comparisons among off-the-shelf methods that use different setups (\eg, data augmentation, initialization, training dataset, \etc) \cite{Seymour-2018, Kim-2017, Zaffar-2021}, which can hide the improvement (or lack thereof) obtained by  algorithmic changes and it does not allow to pinpoint the impact of each individual component. \Cref{tab:vanilla_vs_modifications} shows how some simple engineering choices can have big effects on the recall metric.

\begin{table}[tb]
  \centering
  \resizebox{\columnwidth}{!}{
  \begin{tabular}{lcccccc}
    \toprule
    &  \multirow{2}{*}{Vanilla} & Resize & Data augm. & Pred. refinement & PCA & \multirow{2}{*}{CRN \cite{Kim-2017}} \\
    &  & (80\%) & (brightness = 2) &  (\it{nearest crop}) & (2048) &\\
    \midrule
    R@1 & 63.4 & 64.3 & 68.6 & 67.0 & 56.6 & 68.8 \\
    \bottomrule
  \end{tabular}}
  \vspace{-0.2cm}
  \caption{Example of how results can be influenced by little train or test time changes to the VG pipeline. Recall@1 for a ResNet-18 with NetVLAD trained on Pitts30k and tested on Tokyo24/7. Results are thoroughly discussed in later sections.}
  \label{tab:vanilla_vs_modifications}
   \vspace{-0.4cm}
\end{table}

Although previous benchmarks for VPR \cite{Zaffar-2021} and the related task of Visual Localization \cite{Pion-2020, Sattler-2018} offer interesting insights, they do not address the aforementioned issues.
For these reasons, we propose a new open-source benchmark that provides researchers with an all-inclusive tool to build, train, and test a wide range of commonly used VG architectures, offering the flexibility to change each component of a geo-localization pipeline.
This allows to rigorously examine how each element of the system influences the final results while providing information computed on-the-fly regarding the number of parameters, FLOPs, descriptors dimensionality, \etc.

Using our framework, we run numerous experiments aiming to understand which components are the most suitable for a real-world application, and derive good  practices depending on the target dataset and one's hardware availability. 
For example, we find that ResNet-50 \cite{He-2016}  provides a  good trade-off between accuracy, FLOPs and model size, and that Visual Transformers can successfully replace the CNN backbones and achieve better geo-localization performances when  trained on larger datasets. Furthermore, we observed that partial negative mining and reduced resolution yield important decrease in computations without significantly compromising the performance, or even yielding gains in some cases.

The benchmark's software and models are hosted at {\small{\url{https://deep-vg-bench.herokuapp.com/}}}.

\section{Related Work}
\label{sec:related}
\myparagraph{Representation learning for visual retrieval and localization} 
Visual Geo-localization (VG), Visual Localization (VL), and Landmark Retrieval (LR) are three well-known Computer Vision tasks that try to establish a mapping between an image and a spatial location, albeit with some nuances.
In VG the goal is to find the geographical location of a given query image and the predicted coordinates are considered correct if they are roughly close to ground truth position \cite{Arandjelovic-2018, Kim-2017, Warburg-2020, Liu-2019, Liu-2021, Ge-2020, Berton-2021, Wang-2019}. 
VL focuses on precisely estimating the 6~DoF camera pose of a query image within a known scene.
VG methods can be used as a part of a VL pipeline, combined with other processing stages that reduce the differences when used in a VL task. Therefore the evaluation papers on VL~\cite{Sattler-2018,  Torii-2021, Pion-2020} might not be indicative of VG performance, justifying a separate benchmark on the latter.
LR is a particular case of Image Retrieval (IR) in which queries contain some landmark, and the goal is to identify all database instances depicting the same landmark, regardless of their visually overlap with the query photo.
Since VG is usually addressed as a retrieval problem where the query position is estimated using the GPS tags of the top retrieved image, several methods originally proposed for LR (or IR in general) have carried over to VG.
LR datasets, both on a city-scale (Oxford and Paris Buildings \cite{Philbin-2007, Philbin-2008}) and on a global scale (Google Landmarks \cite{Noh-2017, Weyand-2020}), consists of a discrete set of landmarks, whereas VG datasets usually cover a continuous geographical area.

IR \cite{Stumm-2013, Cummins-2009, Arandjelovic-2014} is traditionally performed via nearest neighbors search using fixed-size image representations \cite{Csurka-2003,SchindlerCVPR07CityScale,Perronnin-2010,Jegou-2008,Jegou-2011,Torii-2018,Jegou-2014} obtained from the aggregation of highly informative local \cite{Lowe-2004, Bay-2008, Arandjelovic-2012} or global \cite{Oliva-2006, Zaffar-2020} features. 
Convolutional neural networks (CNNs) have become the de-facto standard to extract the features for IR, using various methods to concatenate them \cite{Babenko-2014, Razavian-2014} or pool them \cite{Azizpour-2015, Babenko-2015, Tolias-2016} to create image descriptors. 
Among the deep learning representation methods, one that has proven very effective for VG is NetVLAD \cite{Arandjelovic-2018}, a differentiable implementation of VLAD~\cite{Jegou-2011}  trained end-to-end with the CNN backbone directly for place recognition. The layer has since been used in numerous works \cite{Berton-2021, Gadd-2020, Garg-2021-seqNet, Ge-2020, Hausler-2021, Liu-2019, Wang-2019, Warburg-2020}.
One downside of NetVLAD is that it outputs high-dimensional  descriptors, leading to steep memory requirements for VG systems.
This problem has inspired research on more compact descriptors, either using dimensionality reduction techniques \cite{Babenko-2014,Ong2017,Radenovic-2016,Gordo-2016,Zhu-2018,Cao-2020} or replacing NetVLAD with lighter pooling layers, such as GeM \cite{Radenovic-2019} and R-MAC \cite{Gordo-2017}.
It has also been shown that attention modules can be used to focus feature extraction and aggregation towards the most salient parts of the scene for the geo-localization task \cite{Kim-2017, Mohedano-2018, Liu-2021, Cao-2020}. The Contextual Reweighting Network (CRN) \cite{Kim-2017} is a variation of NetVLAD that adds a contextual modulation to produce a weighting mask based on semi-global context. 
Visual Transformers based on self-attention such as ViT~\cite{Dosovitskiy-2021} and DeiT~\cite{Touvron-2021} have also been used in IR~\cite{Caron-2021,ElNouby-2021}, but not yet in VG.
All these architectures used in VG to learn image representations are trained with metric embedding objectives commonly used in learning-to-rank problems, such as the contrastive loss \cite{Ong2017, Radenovic-2016, Radenovic-2019}, the triplet loss \cite{Arandjelovic-2018, Gordo-2017, Kim-2017} and the SARE 
loss \cite{Liu-2019}.

Our benchmark analyzes how the combination of popular backbone networks, pooling strategies, data augmentation, and engineering choices impacts geo-localization performance and other aspects, such as memory and computational requirements. 

\myparagraph{Benchmarking}
The only available benchmark focused specifically on VG/VPR is VPR-Bench \cite{Zaffar-2021}.
In contrast to our work, \cite{Zaffar-2021} (as well as \cite{Pion-2020} for VL) directly compares off-the-shelf models because it is mainly concerned with the performance of VG in practical settings, where one would likely prefer using a pre-trained model rather than having to fine-tune or train it.
On the other hand, we are more interested in measuring the impact of algorithmic changes, which requires performing comparisons where all other factors are the same.
To this end, we propose a modular framework that allows a fair evaluation of each element of a VG system under identical conditions, ensuring clarity and reliability of the results.

While \cite{Zaffar-2021} also provides insights on descriptors dimensionality and retrieval time, we focus on more general hardware-agnostic statistics, such as FLOPs and model size (Sec. \ref{sec:backbones}), training complexity (Sec. \ref{sec:mining}), storage requirements (Sec. \ref{sec:resolution}).

\begin{figure*}[t!]
    \centering
    \includegraphics[width=0.7\linewidth]{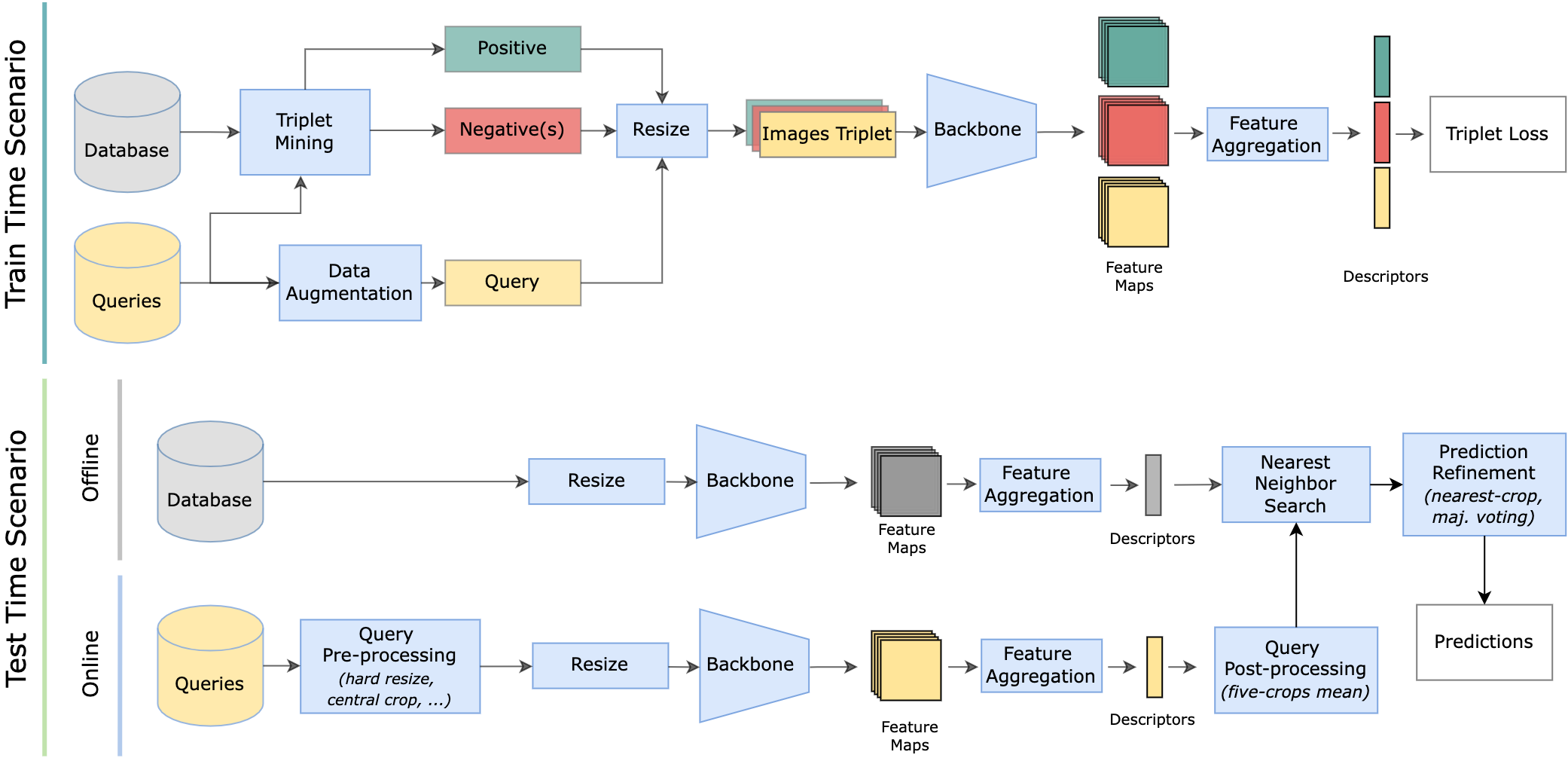}
    \caption{\textbf{Diagram of a visual geo-localization system.}
    Throughout this work, we rigorously and fairly analyze each component of a visual geo-localization system (the light blue blocks) comparing a variety of different implementations, both for train and test time.
    }
    \label{fig:diagram}
    \vspace{-0.4cm}
\end{figure*}

\section{Methodology} 
\label{sec:methodology}

This section describes the VG pipeline used in our benchmark (\cf Fig.~\ref{fig:diagram}) and our experimental setup.
\subsection{Visual Geo-localization System}
The VG task is commonly tackled using an image retrieval pipeline: given a new photo (\emph{query}) to be geo-localized, its location is estimated by matching it to a database of geo-tagged images. 
A VG system is thus an algorithm that first extracts descriptors for the database images (offline) and for the query photo (online), 
then it applies a nearest neighbors search in the descriptor space.
The orange blocks in Fig.~\ref{fig:diagram} show that a VG system is built through several design choices, including network architectures, negative mining methods, and engineering aspects such as image sizes and data augmentation. 
All of these choices impact the behavior of the system, both in terms of performance and  required resources. 
We propose a new benchmark to systematically investigate the impact of the components of VG systems, using the modular architecture shown in \cref{fig:diagram} as a canvas to reproduce most VG methods based on CNN backbones and to develop new models based on
Visual Transformers.

This abstract model contains  several components that can be modified, both during training and test time: the {\bf backbone} (Sec.~\ref{sec:backbones}); {\bf feature aggregation} (Sec.~\ref{sec:aggregation_methods}); {\bf mining} training examples (Sec.~\ref{sec:mining}); {\bf image resizing} (Sec.~\ref{sec:resolution}); {\bf data augmentation} (Sec.~\ref{sec:data_augmentation}).
We conduct a series of tests focused individually on each of these elements, to systematically show each component's influence.
Due to limited space, we only summarize here the results of some experiments, while detailed results and additional experiments on {\bf pre/post-processing methods and predictions refinement}, {\bf effect of pre-training} and many other aspects are provided in the \suppmat.

The code of the benchmark follows the modular structure shown in \cref{fig:diagram}, where each component can be modified.
We further provide scripts to download and format a number of datasets, and to train and test the models making easy to perform a large number of experiments while ensuring consistency and reproducibility of results.
Our codebase allows to easily reproduce the architectures used in a wide range of works \cite{Arandjelovic-2018, Kim-2017, Radenovic-2019, Tolias-2016, Revaud-2019, Gordo-2017, Liu-2019, Warburg-2020} and commonly used training protocols \cite{Arandjelovic-2018, Warburg-2020, Liu-2019}.
More details on the software are provided in \cref{sec:supp_software}. 

%%%%%%%%%%%%%%%%%%%%%%%%%%%%%%%% DATASETS
%%%%%%%%%%%%%%%%%%%%%%%%%%%%%%%% DATASETS
%%%%%%%%%%%%%%%%%%%%%%%%%%%%%%%% DATASETS

\subsection{Datasets}
We use six highly heterogeneous datasets (see \cref{tab:datasets} and maps in \cref{sec:supp_datasets}), which together cover a variety of real-world scenarios: different scales, degree of inter-image variability, different camera types.
For training, we use Pitts30k \cite{Arandjelovic-2018} and Mapillary Street-Level Sequences 
(MSLS) \cite{Warburg-2020} datasets, as they provide a small and large amount of images, respectively.
While Pitts30k is very homogeneous, 
\ie all images share the same resolution, weather conditions and camera, MSLS represents a wide range of conditions from very diverse cities.
Regarding MSLS, given the lack of labels for the test set, we follow \cite{Hausler-2021} and report validation recalls computed on the validation set.
To assess inter-dataset robustness, we also test all models on four other datasets: Tokyo 24/7 \cite{Torii-2018}, Revisited San Francisco (R-SF) \cite{Chen-2011,Li-2012,Torii-2021}, 
Eynsham \cite{Cummins-2009} and St Lucia \cite{Milford-2008}.
Further details on these datasets, such as their geographical coverage, are included in \cref{sec:supp_datasets}. 

\begin{table}[t!]
  \centering
  \resizebox{\columnwidth}{!}{
  \begin{tabular}{lcccllc}
    \toprule
    \textbf{} & \multicolumn{1}{l}{\textbf{\begin{tabular}[c]{@{}c@{}}\# train/val \\ datab./queries\end{tabular}}} 
    & \multicolumn{1}{l}{\textbf{\begin{tabular}[c]{@{}c@{}}\# test \\ datab./queries\end{tabular}}} 
    & \multicolumn{1}{l}{\textbf{\begin{tabular}[c]{@{}c@{}}Dataset \\ size\end{tabular}}} 
    & \multicolumn{1}{l}{\textbf{\begin{tabular}[c]{@{}c@{}}Database \\ type\end{tabular}}} 
    & \multicolumn{1}{l}{\textbf{\begin{tabular}[c]{@{}c@{}}Database \\ img. size\end{tabular}}} 
    & \multicolumn{1}{l}{\textbf{\begin{tabular}[c]{@{}c@{}}Queries \\ type\end{tabular}}} \\
    \midrule
    Pitts30k & 20K / 15K & 10K / 6.8K & 2.0 GB & panorama & 480$\times$640 & panorama  \\
    MSLS & 934K / 514K & 19K / 11K & 56 GB & front-view & 480$\times$640& front-view  \\
    Tokyo 24/7 & 0 / 0 &  75K / 315 & 4.0 GB & panorama & 480$\times$640 & phone$^{\ast}$ \\
    R-SF & 0 / 0 &  1.05M / 598 & 36 GB & panorama & 480$\times$640 & phone$^{\ast}$  \\
    Eynsham & 0 / 0 & 24K / 24K & 1.2 GB & panorama & 512$\times$384 & panorama  \\
    St Lucia & 0 / 0 & 1.5K / 1.5K & 124 MB & front-view & 480$\times$640 & front-view \\
    \bottomrule
  \end{tabular}}
  \caption{\textbf{Summary of the datasets:}
  "panorama" means images are cropped from a 360° panorama (including undistortion); "front-view" means that only one (forward facing) view is available; "phone" means photos were collected with a smartphone. "panorama" and "front-view" images were taken with car-rooftop cameras.
  $^{\ast}$ Variable resolution.}
  \label{tab:datasets}
  \vspace{-0.4cm}
\end{table}

\subsection{Benchmark Protocol}
In all experiments, unless otherwise specified, we use the metric of recall@N (R@N) measuring the percentage of queries for which one of the top-N retrieved images was taken within a certain distance of the query location. 
We mostly focus on R@1 and, following common practice in the literature
\cite{Arandjelovic-2018, Kim-2017, Liu-2019, Peng_2021_appsvr, Peng_2021_sralNet, Berton-2021, BertonICCV21, Warburg-2020, Hausler-2021}, use 25 meters as a distance threshold, but we also investigate how results change varying thresholds and top-N (\cf. \cref{sec:supp_metrics}).
For reliability, all results are averaged over three repetitions of experiments.
To avoid overloading the tables, standard deviations are shown in the \suppmat, where the reported experiments are a superset of the ones in this manuscript.
Training is performed until recall@5 on the validation set does not improve for 3 epochs.
Given the variability in datasets size (see \cref{tab:datasets}), we define an epoch as a pass over 5,000 queries.
We use the Adam optimizer \cite{Kingma-2014} for training, as in general it leads to faster convergence and better performance than SGD.
Following the widely used training protocol defined in \cite{Arandjelovic-2018}, we use a batch size of 4 triplets, where each triplet is composed of an anchor (the query), a positive and 10 negatives.
Following standard practice \cite{Arandjelovic-2018, Warburg-2020, Hausler-2021, Wang-2019, Berton-2021, Liu-2019}, at train time, the positive is selected as the nearest database image in features space among those within a 10 meters radius from the query and  negative images selected from those further than 25. 
Due to the size of each dataset, we use full database mining when training on the Pitts30k, and partial  mining when training on the MSLS (cf. Sec. \ref{sec:mining} for details). 

\vspace{-0.4cm}
\section{Results}
\label{sec:results}
Throughout this section, we explore how each block from \cref{fig:diagram} influences the results.
Specifically, we first investigate the use of different architectures, with a focus on backbones (\cref{sec:backbones}), aggregation methods (\cref{sec:aggregation_methods}) and Transformers-based networks (\cref{sec:transformers}).
We then move to train-time components (\ie negative mining \cref{sec:mining} and data augmentation \cref{sec:data_augmentation}), to understanding how the resolution of the images influences a VG system (\cref{sec:resolution}), and finally we explore the use of efficient nearest neighbor search algorithms (\cref{sec:nearest_neighbor}).
Given the limited amount of space in the manuscript, a thorough extension over each one of these sections can be found in the \suppmat, as well as further experiments on various metrics and more.

%%%%%%%%%%%%%%%%%%%%%%%%%%%%%%%% BACKBONES
%%%%%%%%%%%%%%%%%%%%%%%%%%%%%%%% BACKBONES
%%%%%%%%%%%%%%%%%%%%%%%%%%%%%%%% BACKBONES

\begin{table*}[tb!]
  \centering
  \resizebox{\textwidth}{!}{
  \begin{tabular}{llllllllllllllllll}
    \toprule
    \multirow{2}{*}{Backbone} & \multirow{2}{*}{\begin{tabular}[c]{@{}l@{}}Aggregation\\ Method\end{tabular}} 
    & \multirow{2}{*}{\begin{tabular}[c]{@{}l@{}}Features\\ Dim\end{tabular}} 
    & \multirow{2}{*}{\begin{tabular}[c]{@{}l@{}}FLOPs\\(GF)\end{tabular}}
    & \multirow{2}{*}{\begin{tabular}[c]{@{}l@{}}Model\\Size\\(MB)\end{tabular}}
    & \multirow{2}{*}{\begin{tabular}[c]{@{}l@{}}Extraction\\Time\\(ms)\end{tabular}}
    & \multicolumn{6}{l}{Training on Pitts30k} & \multicolumn{6}{l}{Training on MSLS} \\
    & & & & & & \begin{tabular}[c]{@{}l@{}}R@1\\ Pitts30k\end{tabular} & \begin{tabular}[c]{@{}l@{}}R@1\\ MSLS\end{tabular} & \begin{tabular}[c]{@{}l@{}}R@1 \\ Tokyo 24/7\end{tabular} & \begin{tabular}[c]{@{}l@{}}R@1\\ R-SF\end{tabular} & \begin{tabular}[c]{@{}l@{}}R@1\\ Eynsham\end{tabular} & \begin{tabular}[c]{@{}l@{}}R@1\\ St Lucia\end{tabular} & \begin{tabular}[c]{@{}l@{}}R@1\\ Pitts30k\end{tabular} & \begin{tabular}[c]{@{}l@{}}R@1\\ MSLS\end{tabular} & \begin{tabular}[c]{@{}l@{}}R@1 \\ Tokyo 24/7\end{tabular} & \begin{tabular}[c]{@{}l@{}}R@1\\ R-SF\end{tabular} & \begin{tabular}[c]{@{}l@{}}R@1\\ Eynsham\end{tabular} & \begin{tabular}[c]{@{}l@{}}R@1\\ St Lucia\end{tabular} \\
    \midrule
    VGG-16     &GeM     &512   &188.01 &56.13 &12.3&78.5 &43.4 &39.9 &40.4 &70.2 &46.4 &70.2 &66.7 &43.6 &32.1 &80.4 &79.9  \\
    ResNet-18  &GeM     &256   &17.29  &10.63  &4.1&77.8 &35.3 &35.3 &34.2 &64.3 &46.2 &71.6 &65.3 &42.8 &30.5 &80.3 &83.2  \\
    ResNet-50  &GeM     &1024  &40.61  &32.71  &6.7&82.0 &38.0 &41.5 &45.4 &66.3 &\textbf{59.0} &\textbf{77.4} &72.0 &\textbf{55.4} &45.7 &\textbf{83.9} &91.2  \\
    ResNet-101 &GeM     &1024  &86.29  &105.36 &9.6&\textbf{82.4} &\textbf{39.6} &\textbf{44.0} &\textbf{52.5} &\textbf{69.0} &57.6 &77.2 &\textbf{72.5} &51.0  &\textbf{46.9} &83.6 &\textbf{91.6}  \\
    \midrule
    VGG-16     &NetVLAD &32768 &188.09 &56.38 &13.0&83.2 &50.9 &61.4 &64.6 &74.4 &50.1 &79.0 &74.6 &61.9 &57.1 &84.2 &86.7  \\
    ResNet-18  &NetVLAD &16384 &17.27  &10.76  &4.4&86.4 &47.4 &63.4 &61.4 &76.8 &57.6 &\textbf{81.6} &75.8 &62.3 &55.1 &87.1 &92.1  \\
    ResNet-50  &NetVLAD &65536 &40.51  &33.21  &8.5&86.0 &50.7 &69.8 &67.1 &\textbf{77.7} &60.2 &80.9 &76.9 &\textbf{62.8} &51.5 &\textbf{87.2} &93.8  \\
    ResNet-101 &NetVLAD &65536 &86.06  &105.86 &11.5&\textbf{86.5} &\textbf{51.8} &\textbf{72.2} &\textbf{67.5} &74.0 &\textbf{63.6} &80.8 &\textbf{77.7} &59.0 &\textbf{56.1} &86.7 &\textbf{95.1}  \\
    \bottomrule
  \end{tabular}}
\vspace{-0.1cm}
  \caption{Results and computational requirements with different convolutional \textbf{backbones}. Extraction time is the average over a 1000 forward passes.}
  \label{tab:t1_backbones}
\vspace{-0.3cm}
\end{table*}
\subsection{CNN Backbones}
\label{sec:backbones}
Tasked with extracting highly informative feature maps from images, the CNN backbone represents a fundamental component of any VG system.
To understand its impact, we experiment with four  CNN backbones (VGG16 \cite{Simonyan-2015}, ResNet-18, ResNet-50 and ResNet-101  \cite{He-2016}), combined with two popular aggregation methods, GeM \cite{Radenovic-2019} and NetVLAD \cite{Arandjelovic-2018}. 
Note that this seemingly limited number of backbones covers several state-of-the-art architectures in VG and image retrieval \cite{Arandjelovic-2018, Kim-2017, Radenovic-2019, Tolias-2016, Revaud-2019, Gordo-2017, Liu-2019, Warburg-2020, Peng_2021_appsvr, Peng_2021_sralNet}.
For all ResNets, we use the feature maps extracted from the \textit{conv4\_x} layer\footnote{Preliminary results have shown on average better recall and efficiency rather than using until \textit{conv5\_x} (see \cref{tab:t1_backbones_resnets} in the \suppmat)}. 
For VGG16, we use all the convolutional layers, excluding the last pooling before the classifier part. 
\Cref{tab:t1_backbones} shows the results of our experiments. 

\myparagraph{Discussion}
We can see that deeper ResNets, such as ResNet-50 and ResNet-101, achieve better results w.r.t. their shallower counterparts.
In particular, ResNet-50 shows recalls on par with ResNet-101, but with the advantage of less than half the FLOPs and model size, making the former a more practically relevant option than the latter.
ResNet-18 performs worse, but allows for much faster and lighter computation, making it the most efficient, lightweight backbone.
Moreover, results considerably depend on the training data: as an example, training the same network on Pitts30k or MSLS yields a 30\% gap testing the model on St. Lucia, as well as a noticeable difference on other datasets too.
This effect demonstrates that comparing models trained on different datasets, as done in \cite{Zaffar-2021}, can be misleading.

%%%%%%%%%%%%%%%%%%%%%%%%%%%%%%%% AGGREGATIONS
%%%%%%%%%%%%%%%%%%%%%%%%%%%%%%%% AGGREGATIONS
%%%%%%%%%%%%%%%%%%%%%%%%%%%%%%%% AGGREGATIONS

\begin{table*}[t]
  \centering
  \resizebox{\textwidth}{!}{
  \begin{tabular}{llllllllllllllllll}
    \toprule
    \multirow{2}{*}{Backbone} & \multirow{2}{*}{\begin{tabular}[c]{@{}l@{}}Aggregation\\ Method\end{tabular}} & \multirow{2}{*}{\begin{tabular}[c]{@{}l@{}}Features\\ Dim\end{tabular}} & \multicolumn{6}{l}{Training on Pitts30k} & \multicolumn{6}{l}{Training on MSLS} \\
    &  &  & \begin{tabular}[c]{@{}l@{}}R@1\\ Pitts30k\end{tabular} & \begin{tabular}[c]{@{}l@{}}R@1\\ MSLS\end{tabular} & \begin{tabular}[c]{@{}l@{}}R@1 \\ Tokyo 24/7\end{tabular} & \begin{tabular}[c]{@{}l@{}}R@1\\ R-SF\end{tabular} & \begin{tabular}[c]{@{}l@{}}R@1\\ Eynsham\end{tabular} & \begin{tabular}[c]{@{}l@{}}R@1\\ St Lucia\end{tabular} & \begin{tabular}[c]{@{}l@{}}R@1\\ Pitts30k\end{tabular} & \begin{tabular}[c]{@{}l@{}}R@1\\ MSLS\end{tabular} & \begin{tabular}[c]{@{}l@{}}R@1 \\ Tokyo 24/7\end{tabular} & \begin{tabular}[c]{@{}l@{}}R@1\\ R-SF\end{tabular} & \begin{tabular}[c]{@{}l@{}}R@1\\ Eynsham\end{tabular} & \begin{tabular}[c]{@{}l@{}}R@1\\ St Lucia\end{tabular} & \begin{tabular}[c]{@{}l@{}}R@1\\ Average\end{tabular}\\
    \midrule
    ResNet-50   &GeM                 &1024  &82.0 &38.0 &41.5 &45.4 &66.3 &59.0 &\textbf{77.4} &72.0 &\textbf{55.4} &\textbf{45.7} &83.9 &91.2  & 63.2 \\
    ResNet-50   &NetVLAD + PCA 1024  &1024  &83.9 &46.5 &59.4 &53.2 &72.5 &57.7 &\textbf{77.4} &74.8 &51.3 &39.0 &85.2 &92.9  & 66.2\\
    ResNet-50   &CRN + PCA 1024      &1024  &\textbf{84.1} &\textbf{49.9} &\textbf{64.6} &\textbf{58.8} &\textbf{74.3} &\textbf{63.4} &77.3 &\textbf{75.6} &51.8 &38.8 &\textbf{85.7} &\textbf{94.1} & \textbf{68.2}  \\
    \midrule
    ResNet-50   &GeM + FC 2048       &2048  &80.1 &33.7 &43.6 &48.2 &70.0 &56.0 &\textbf{79.2} &73.5 &\textbf{64.0} &\textbf{55.1} &86.1 &90.3  & 65.0 \\
    ResNet-50   &NetVLAD + PCA 2048  &2048  &84.4 &47.9 &62.6 &56.0 &74.1 &58.9 &78.5 &75.4 &52.8 &42.6 &85.8 &93.4 & 67.7 \\
    ResNet-50   &CRN + PCA 2048      &2048  &\textbf{84.7} &\textbf{51.2} &\textbf{67.1} &\textbf{62.3} &\textbf{75.8} &\textbf{65.0} &78.3 &\textbf{76.3} &54.3 &42.8 &\textbf{86.2} &\textbf{94.4} & \textbf{69.9} \\
    \midrule
    ResNet-50   &GeM + FC 65536      &65536 &80.8 &35.8 &45.6 &49.0 &72.5 &59.6 &79.0 &74.4 &\textbf{69.2} &\textbf{58.4} &86.2 &90.8  & 66.8 \\
    ResNet-50   &NetVLAD             &65536 &\textbf{86.0} &50.7 &69.8 &67.1 &77.7 &60.2 &\textbf{80.9} &76.9 &62.8 &51.5 &87.2 &93.8 & 72.1\\
    ResNet-50   &CRN                 &65536 &85.8 &\textbf{54.0} &\textbf{73.1} &\textbf{70.9} &\textbf{79.7} &\textbf{65.9} &80.8 &\textbf{77.8} &63.6 &53.4 &\textbf{87.5} &\textbf{94.8}  & \textbf{73.9}\\
    \bottomrule
  \end{tabular}}
  \caption{\textbf{Aggregation methods:} we report results with different aggregation methods downscaled or upscaled to equivalent dimensionality. }
  \label{tab:t2_aggregation_methods}
\vspace{-0.4cm}
\end{table*}
\subsection{Aggregation and Descriptor Dimensionality}
\label{sec:aggregation_methods}

Aggregations methods are layers tasked with processing the output features of the backbone.
Over the years, a number of such methods have been proposed, from shallow pooling layers ~\cite{Babenko-2015, Razavian-2015} to more complex modules \cite{Arandjelovic-2018, Kim-2017}.
Our framework allows to compute results with a number of them, namely SPOC~\cite{Babenko-2015}, MAC~\cite{Razavian-2015}, R-MAC~\cite{Tolias-2016}, RRM~\cite{Kordopatis-2021}, GeM \cite{Radenovic-2019}, NetVLAD \cite{Arandjelovic-2018} and CRN \cite{Kim-2017}.
While a complete list of results with all aggregation methods is shown in \cref{sec:supp_aggregation_methods}, in \cref{tab:t2_aggregation_methods} we report the performance of the best performing aggregators: GeM, NetVLAD and CRN.
Given the difference in size of the outputted descriptors, we apply PCA or a fully connected (FC) layer to even their dimensionality.

\myparagraph{Discussion}
The results in \cref{tab:t2_aggregation_methods} show that performance strongly depends on the training set. When training on the small Pitts30k, the best results are obtained globally with CRN, even when reducing its dimension to be the same as GeM. However, when training on the much larger MSLS, the advantage of CRN is reduced, and both CRN and NetVLAD end up being significantly outclassed on Tokyo and R-SF\footnote{The reason could be that these two datasets have different query and database image types, \ie phone-taken and panorama images, respectively.} 
by GeM, making it a more compelling choice. 
Furthermore, the dimensionality reduction via PCA yields a significant drop in performance for NetVLAD and CRN, while adding a fully connected layer on top of GeM gives best results when trained on a large scale dataset, which is the type of scenario for which GeM was proposed \cite{Radenovic-2019}.
Note that while the CRN aggregator yields the most robust results, it has the drawbacks of requiring a two-stage training process that almost doubles the training time and three times more hyperparameters w.r.t. NetVLAD. In addition, depending on the initialization of its modulation layer, training does not always converge.

%%%%%%%%%%%%%%%%%%%%%%%%%%%%%%%% TRANSFORMERS
%%%%%%%%%%%%%%%%%%%%%%%%%%%%%%%% TRANSFORMERS
%%%%%%%%%%%%%%%%%%%%%%%%%%%%%%%% TRANSFORMERS

\subsection{Visual Transformers}
\label{sec:transformers}
In this section we investigate how Visual Transformers compare to more traditional CNN-based methods in VG.
For this analysis we use two popular Transformer architectures, the Vision Transformer (ViT) \cite{Dosovitskiy-2021}, which processes the images by splitting them into sequences of flattened 2D patches, and the Compact Convolutional Transformer (CCT) \cite{Hassani-2021}, which incorporates convolutional layers to insert the inductive bias of CNNs. 
Following \cite{ElNouby-2021}, we use as a global descriptor the CLS token, which is the output state of the prepended learnable embedding to the sequence of patches \cite{Dosovitskiy-2021}. Moreover, we test the use of CCT in conjunction with traditional aggregation methods, such as  GeM \cite{Radenovic-2019} and NetVLAD \cite{Arandjelovic-2018}, and with  SeqPool, which was specifically introduced in~\cite{Hassani-2021} for Transformers.
\begin{table}[tb!]
  \centering
  \resizebox{\columnwidth}{!}{
  \begin{tabular}{llllllllllll}
    \toprule
    \multirow{2}{*}{Backbone} & \multirow{2}{*}{\begin{tabular}[c]{@{}l@{}}Aggreg.\\ Method\end{tabular}} & \multirow{2}{*}{\begin{tabular}[c]{@{}l@{}}Feat.\\ Dim\end{tabular}} & \multirow{2}{*}{\begin{tabular}[c]{@{}l@{}}FLOPs\\ (GF)\end{tabular}} & \multicolumn{6}{l}{Training on MSLS} \\
    & & & & \begin{tabular}[c]{@{}l@{}}R@1\\ Pitts30k\end{tabular} & \begin{tabular}[c]{@{}l@{}}R@1\\ MSLS\end{tabular} & \begin{tabular}[c]{@{}l@{}}R@1 \\ Tok. 24/7\end{tabular} & \begin{tabular}[c]{@{}l@{}}R@1\\ R-SF\end{tabular} & \begin{tabular}[c]{@{}l@{}}R@1\\ Eyns.\end{tabular} & \begin{tabular}[c]{@{}l@{}}R@1\\ St Lucia\end{tabular} \\
    \midrule
    ResNet-18 & GeM   & 256  & 17.29 &   71.6 & 65.3 & 42.8 & 30.5 & 80.3 & 83.2 \\
    ResNet-50       &GeM      &1024  &40.61   & 77.4  &72.0   &55.4   &45.7   &83.9  &91.2     \\
    ViT     & CLS     & 768  & 82.31 &  \textbf{82.9} & \textbf{73.5} & \textbf{59.9} & \textbf{65.0} & 84.5 & 93.6 \\
    CCT     & CLS     & 384  & 22.34 & 79.6 & 71.1 & 52.0 & 49.9 & 85.6 & \textbf{94.0} \\
    CCT     & SeqPool & 384  & 26.19 &  81.4 & 71.0 & 59.1 & 60.5 & \textbf{86.1} & 92.4 \\
    CCT     & GeM     & 384  & 22.36 &  78.7 & 72.0 & 48.8 & 48.6 & 83.9 & 92.9 \\
    \midrule
    ResNet-18 & NetVLAD & 16384 & 17.27 &   81.6 & 75.8 & 62.3 & 55.1 & 87.1 & 92.1 \\
    ResNet-50       &NetVLAD  &65536 &40.51  & 80.9   &76.9 &62.3  &51.5   &87.2   &93.8   \\
    CCT       & NetVLAD & 24576 & 18.53 &  \textbf{85.1} & \textbf{79.9} & \textbf{70.3} & \textbf{65.9} & \textbf{87.4} & \textbf{98.4} \\
    \bottomrule
  \end{tabular}
  }
\vspace{-0.1cm}
  \caption{\textbf{Transformers} Comparison of traditional CNN architectures with novel Transformers-based approaches. 
  }
  \label{tab:transformers}
\vspace{-0.5cm}
\end{table}
%%%%%%%%%%%%%%%%%% 
\begin{table*}[!ht]
  \centering
  \resizebox{\textwidth}{!}{
  \begin{tabular}{lllllllllllllllll}
    \toprule
    \multirow{2}{*}{Backbone} &
    \multirow{2}{*}{\begin{tabular}[c]{@{}l@{}}Aggregation\\ Method\end{tabular}} &
    \multirow{2}{*}{\begin{tabular}[c]{@{}l@{}}Mining\\ Method\end{tabular}} &
    \multirow{2}{*}{\begin{tabular}[c]{@{}l@{}}Space\\\& Time\\Complexity\end{tabular}} &
    \multicolumn{6}{l}{Training on Pitts30k} &
    \multicolumn{6}{l}{Training on MSLS} \\
    &  &  &  & \begin{tabular}[c]{@{}l@{}}R@1\\ Pitts30k\end{tabular} & \begin{tabular}[c]{@{}l@{}}R@1\\ MSLS\end{tabular} & \begin{tabular}[c]{@{}l@{}}R@1 \\ Tokyo 24/7\end{tabular} & \begin{tabular}[c]{@{}l@{}}R@1\\ R-SF\end{tabular} & \begin{tabular}[c]{@{}l@{}}R@1\\ Eynsham\end{tabular} & \begin{tabular}[c]{@{}l@{}}R@1\\ St Lucia\end{tabular} & \begin{tabular}[c]{@{}l@{}}R@1\\ Pitts30k\end{tabular} & \begin{tabular}[c]{@{}l@{}}R@1\\ MSLS\end{tabular} & \begin{tabular}[c]{@{}l@{}}R@1 \\ Tokyo 24/7\end{tabular} & \begin{tabular}[c]{@{}l@{}}R@1\\ R-SF\end{tabular} & \begin{tabular}[c]{@{}l@{}}R@1\\ Eynsham\end{tabular} & \begin{tabular}[c]{@{}l@{}}R@1\\ St Lucia\end{tabular} \\
    \midrule
    ResNet-18    &GeM      &Random           & $\mathcal{O}(1)$ &73.7           &30.5 &31.3 &24.0 &58.2 &41.0 &62.2 &50.6 &28.8 &17.1 &70.2 &71.4  \\
    ResNet-18    &GeM      &Full database    & $\mathcal{O}(\textit{\#db} + \textit{\#q})$          &\bf{77.8} &\bf{35.3} &\bf{35.3} &\bf{34.2} &\bf{64.3} &\bf{46.2} &70.1 &61.8 &\bf{42.8} &\bf{31.3} &79.3 &81.0  \\
    ResNet-18    &GeM      &Partial database & $\mathcal{O}(k_{db}+k_{q}+\textit{\#pos})$ &76.5 &34.2 &33.9 &32.9 &64.0 &45.6 &\bf{71.6} &\bf{65.3} &\bf{42.8} &30.5 &\bf{80.3} &\bf{83.2}  \\
    \midrule
    ResNet-18    &NetVLAD  &Random           &  $\mathcal{O}(1)$ &83.9 &43.6 &55.1 &53.8 &76.3 &53.5 &73.3 &61.5 &45.0 &34.8 &84.9 &79.7  \\
    ResNet-18    &NetVLAD  &Full database    & $\mathcal{O}(\textit{\#db} + \textit{\#q})$          &\bf{86.4} &\bf{47.4} &\bf{63.4} &61.4 &\bf{76.8} &\bf{57.6} &-&-&-&-&-&- \\
    ResNet-18    &NetVLAD  &Partial database & $\mathcal{O}(k_{db}+k_{q}+\textit{\#pos})$ &86.2 &47.3 &61.2 &\bf{62.9} &76.6 &57.1 &\bf{81.6} &\bf{75.8} &\bf{62.3} &\bf{55.1} &\bf{87.1} &\bf{92.1}  \\
    \bottomrule
  \end{tabular}
  }
  \vspace{-1mm}
  \caption{\textbf{Negative mining methods. }
 "Space \& Time Complexity" refers to the complexity of building the cache, which normally is done after iterating over 1000 triplets \cite{Arandjelovic-2018, Warburg-2020}.
  $\#db$ and $\#q$ are the numbers of database and query images, $k_{db}$ and $k_{q}$ are chosen constants (usually set to 1000), and $\#pos$ is the number of positives for the considered queries, which depends on the queries and database density.
  }
  \label{tab:t3_mining}
  \vspace{-0.2cm}
\end{table*}

\begin{figure*}[!ht]
    \centering
    \begin{minipage}{.16\textwidth}
        \begin{subfigure}{\textwidth}
            \includegraphics[width=\textwidth]{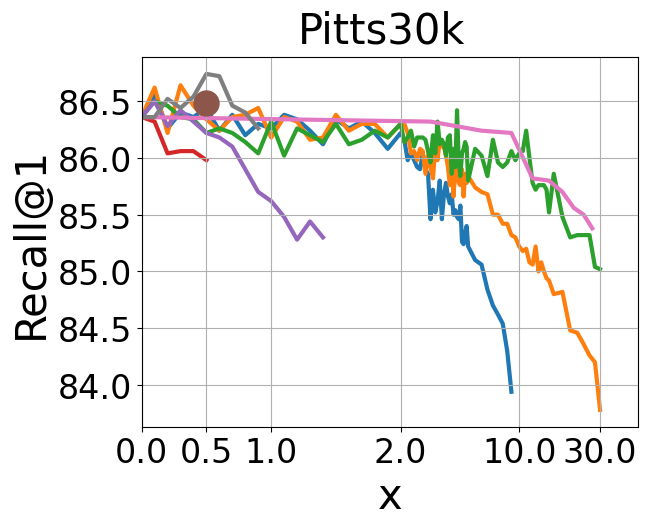}
        \end{subfigure}
    \end{minipage}
    \begin{minipage}{.16\textwidth}
        \begin{subfigure}{\textwidth}
            \includegraphics[width=\textwidth]{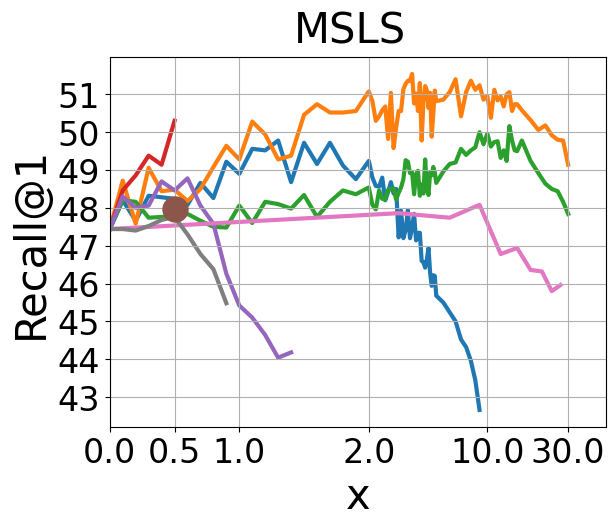}
        \end{subfigure}
    \end{minipage}
    \begin{minipage}{.16\textwidth}
        \begin{subfigure}{\textwidth}
            \includegraphics[width=\textwidth]{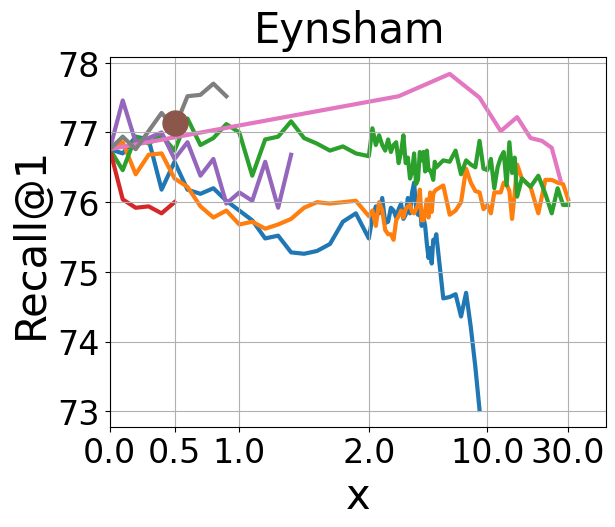}
        \end{subfigure}
    \end{minipage}
    \begin{minipage}{.16\textwidth}
        \begin{subfigure}{\textwidth}
            \includegraphics[width=\textwidth]{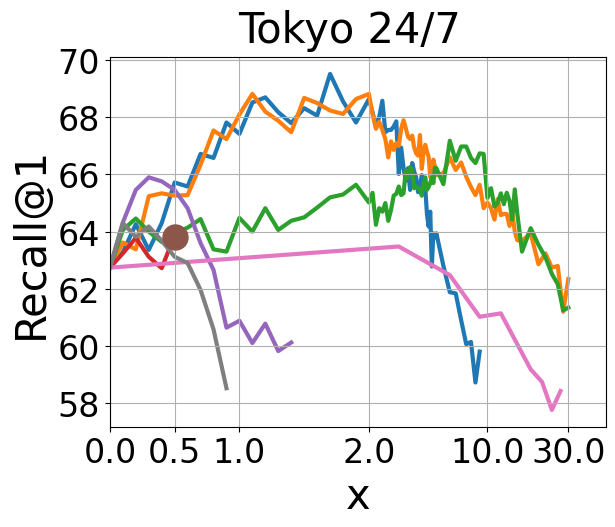}
        \end{subfigure}
    \end{minipage}
    \begin{minipage}{.16\textwidth}
        \begin{subfigure}{\textwidth}
            \includegraphics[width=\textwidth]{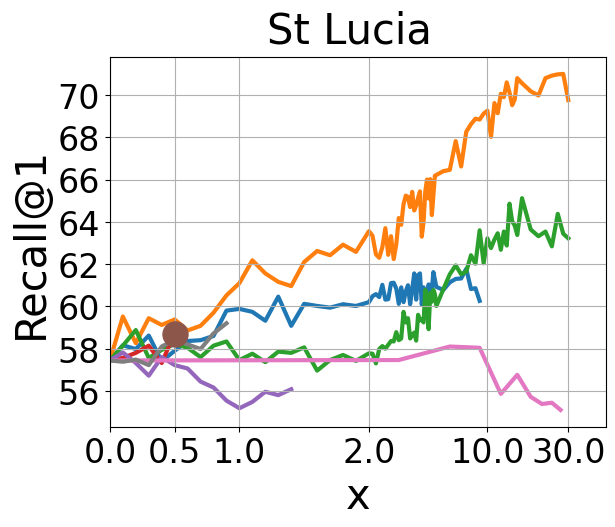}
        \end{subfigure}
    \end{minipage}
    \begin{minipage}{.16\textwidth}
        \begin{subfigure}{\textwidth}
            \includegraphics[width=\textwidth]{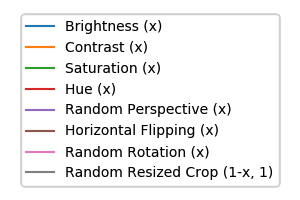}
        \end{subfigure}
    \end{minipage}
    \caption{\textbf{Data Augmentation.} 
    Results obtained applying popular augmentation techniques during training.
    We used PyTorch's transforms, and the x axis relates to the parameter passed to the class; the higher the parameter, the heavier the transform effect (\emph{i.e.} $x=0$ equals to the identity transformation). Refer to {\suppmat} for further details on the transforms.
    }
    \label{fig:data_augmentation}
  \vspace{-0.4cm}
\end{figure*}

\myparagraph{Discussion}
\Cref{tab:transformers} compares traditional CNN-based methods with novel Visual Transformer based 
approaches,
never used before  specifically 
for VG. The main findings of this set of experiments is that they represent a viable alternative to CNN-based backbones
even without an additional aggregation steps using directly the compact and robust representation provided by the  CLS token. Further improvements can be obtained when combined with aggregators such as GeM, SeqPool, NetVLAD, as shown in the table. 
Overall, the results show that these architectures possess better generalization capabilities than their CNN counterparts , and ViT proves to be competitive even with the much bigger NetVLAD descriptors, albeit with higher computational requirements. 
As for CCT, despite  being incredibly lightweight, with a cost comparable to a ResNet-18, consistently outperforms the ResNet-18 and, in many cases, also the ResNet-50, which has roughly double the computational cost. 
Concluding, it seems that the SeqPool aggregator enhances the robustness of the CCT descriptors, providing better generalization and that NetVLAD coupled with CCT outperforms CNN-based methods.  We observe similar 
behaviors when trained on Pitts30k (see \cref{tab:supp_transformers}).
The main limitations of these architectures is the lack of an all-around best configuration. In other words, for each use case, an additional tuning on where to truncate/freeze the network was required, unlike the CNNs which were consistently used up to their \emph{conv4} layer.

\begin{figure*}[t!]
    \centering
    \begin{minipage}{.16\textwidth}
        \begin{subfigure}{\textwidth}
            \includegraphics[width=\textwidth]{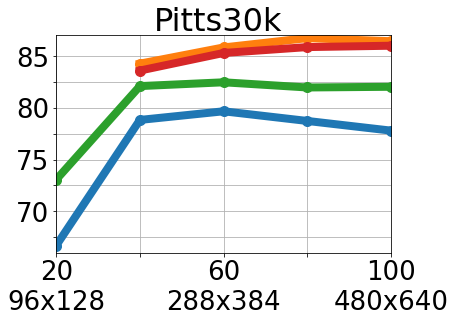}
        \end{subfigure}
    \end{minipage}
    \begin{minipage}{.16\textwidth}
        \begin{subfigure}{\textwidth}
            \includegraphics[width=\textwidth]{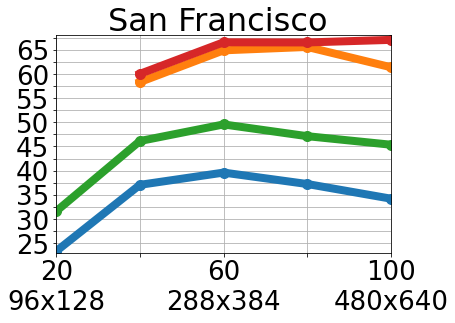}
        \end{subfigure}
    \end{minipage}
    \begin{minipage}{.16\textwidth}
        \begin{subfigure}{\textwidth}
            \includegraphics[width=\textwidth]{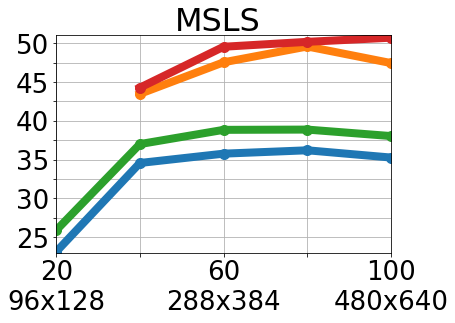}
        \end{subfigure}
    \end{minipage}
    \begin{minipage}{.16\textwidth}
        \begin{subfigure}{\textwidth}
            \includegraphics[width=\textwidth]{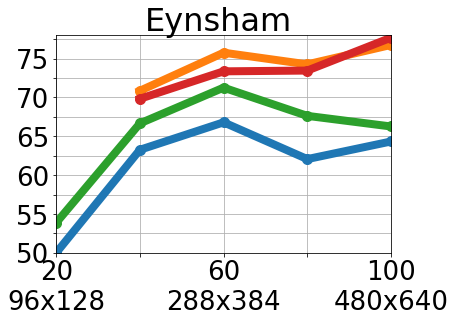}
        \end{subfigure}
    \end{minipage}
    \begin{minipage}{.16\textwidth}
        \begin{subfigure}{\textwidth}
            \includegraphics[width=\textwidth]{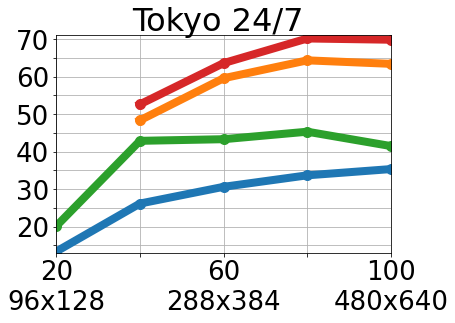}
        \end{subfigure}
    \end{minipage}
    \begin{minipage}{.16\textwidth}
        \begin{subfigure}{\textwidth}
            \includegraphics[width=\textwidth]{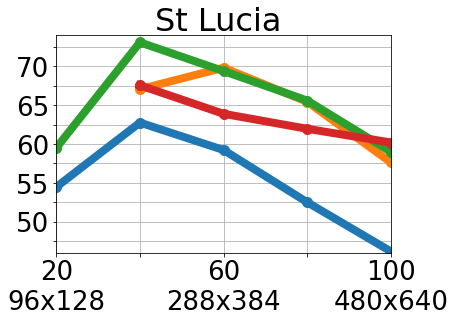}
        \end{subfigure}
    \end{minipage}
    \caption{\textbf{Changing the images' resolution.} On the x-axis is the train and test resolution (N\%), on the y-axis is the recall@1.
    Regarding the curves, red refers to ResNet-50 + NetVLAD, orange to ResNet18 + NetVLAD, green to ResNet-50 + GeM, and blue to ResNet-18 + GeM.
    In many cases, full resolution is not the optimal choice. NetVLAD's initial clusters computation breaks with low resolutions.}
    \label{fig:resolution_train_and_test}
  \vspace{-0.4cm}
\end{figure*}

%%%%%%%%%%%%%%%%%%%%%%%%%%%%%%%% MINING
%%%%%%%%%%%%%%%%%%%%%%%%%%%%%%%% MINING
%%%%%%%%%%%%%%%%%%%%%%%%%%%%%%%% MINING
\subsection{Negative Mining}
\label{sec:mining}
An important step in a VG pipeline is the mining of negatives: ideally, we want to select images of different scenes that appear visually similar to the query to ensure that the model learns highly informative features for the task.
We extensively compare three main mining strategies: full
database mining~\cite{Arandjelovic-2018}, partial database mining~\cite{Warburg-2020}
where only a reasonable subset of images is ranked,  and random negative sampling. 
Details about the mining strategies, full set of results and their analyses can be  found in the \cref{sec:supp_mining}, here we present in \cref{tab:t3_mining} only a subset of our results as illustration and summarize our main findings. 

\myparagraph{Discussion} As expected, both full and partial database mining outperform the random negative sampling. The latter, in spite of its low cost  yields in average 5\% lower results on Pitts30k, due to the low variability of the dataset. Indeed, on the larger MSLS results drop of 10\% or more. 
On the other hand, full database mining does not provide always best  performance and on average its gain over 
partial mining is around 1\%. Furthermore,
on large scale datasets such as MSLS full mining is not feasible in a reasonable time. These results clearly show that 
partial mining is, in general, a great compromise between cost and accuracy.

%%%%%%%%%%%%%%%%%%%%%%%%%%%%%%%% DATA AUGMENTATION
%%%%%%%%%%%%%%%%%%%%%%%%%%%%%%%% DATA AUGMENTATION
%%%%%%%%%%%%%%%%%%%%%%%%%%%%%%%% DATA AUGMENTATION

\subsection{Data Augmentation}
\label{sec:data_augmentation}
Here we investigate if and which data augmentation are beneficial for VG methods, and if the improvements are domain-specific or can generalize to diverse datasets.
We apply data augmentation to the query, with the sole exception of random horizontal flipping, for which we either flip or not flip the whole triplet.
We run experiments with many popular augmentation techniques, training a ResNet-18 with NetVLAD on Pitts30k.

\myparagraph{Discussion}
Plots of the results are in \cref{fig:data_augmentation} 
(shown in higher resolution in the \cref{fig:supp_data_augmentation}). 
Depending on the test dataset, we observe different   impact of these augmentations. On one hand, 
on Pitts30k augmentation only worsens results, probably due to dataset homogeneity between train and test.
On the other hand, we see that some techniques can improve robustness on unseen datasets,
in particular color jittering methods that change brightness, contrast and saturation.
As an example, setting contrast\footnote{This refers to PyTorch's \texttt{ColorJittering()} function.} up to 2 can improve recall@1 by more than 3\% on MSLS, 5\% on Tokyo 24/7, 5\% on St Lucia, with a less than 1\% drop on Pitts30k and Eynsham.
Although most augmentations fail to produce consistent improvements, two notable exceptions are random horizontal flipping (with probability 50\%) and random resized cropping, where crops are as small as 50\% of the image size (and then resized to full resolution).

%%%%%%%%%%%%%%%%%%%%%%%%%%%%%%%% RESIZE
%%%%%%%%%%%%%%%%%%%%%%%%%%%%%%%% RESIZE
%%%%%%%%%%%%%%%%%%%%%%%%%%%%%%%% RESIZE

\subsection{Resize}
\label{sec:resolution}
While common VG datasets have images of resolutions around 480x640 pixels, it is interesting to investigate how resizing them can affect the results.
To this end, we perform experiments by training and testing models on images of lower resolution, by reducing both sides of the images from 80\% to 20\% of their original size, both at train and test time on Pitts30k.  
We conduct this analysis with CNNs followed by GeM or NetVLAD, since such architectures do not require a fixed input image resolution.

\myparagraph{Discussion}
Interestingly, it can be seen in \Cref{fig:resolution_train_and_test} that using the highest available resolution is in most cases superfluous, and often even detrimental.
On average, NetVLAD's descriptors seem to better handle higher resolutions than their GeM counterparts.
Lower resolutions, as low as 40\%, show improved results especially when there is a wide domain gap between train and test sets: this is exemplified by the results on the St Lucia dataset, which is very different from Pitts30k (the former has only forward views) and shows best R@1 performance when using 40\% of the original resolution.
This behaviour can be explained by the disappearance of domain specific low-level patterns (\eg, texture and foliage) when the size of the image is reduced.
In general 60\% is a good compromise, suggesting that for geo-localization, which is strongly related to appearance-based retrieval, fine details are not too important.

Finally, note that 40\% resolution means reducing it to 192x256, with FLOPs going down to (40\%)\textsuperscript{2} = 16\% w.r.t. full resolution images. 
Storage needs also decrease in the same fashion as FLOPs, and although images are not directly needed in a retrieval system (only descriptors and coordinates are used for kNN), they can be used useful for post-processing, \eg, spatial verification, or to generate a visual response for users.

%%%%%%%%%%%%%%%%%%%%%%%%%%%%%%%% NEAREST NEIGHBOR
%%%%%%%%%%%%%%%%%%%%%%%%%%%%%%%% NEAREST NEIGHBOR
%%%%%%%%%%%%%%%%%%%%%%%%%%%%%%%% NEAREST NEIGHBOR

\begin{figure}[tb]
    \centering
    \begin{minipage}{.49\linewidth}
        \begin{subfigure}{\linewidth}
            \includegraphics[width=\linewidth]{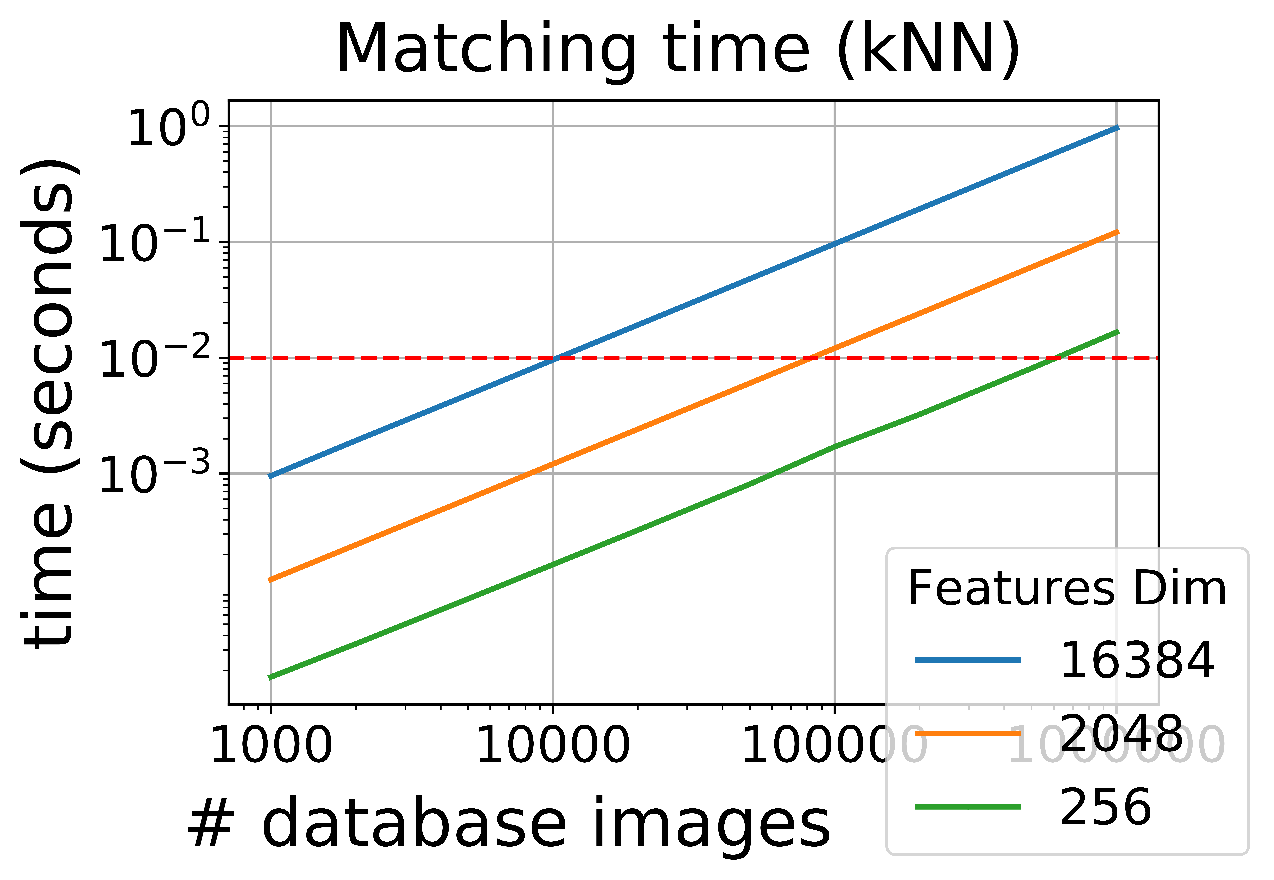}
            \centering\subcaption{}
            \label{fig:matching_time}
        \end{subfigure}
    \end{minipage}
    \begin{minipage}{.49\linewidth}
        \begin{subfigure}{\linewidth}
            \includegraphics[width=\linewidth]{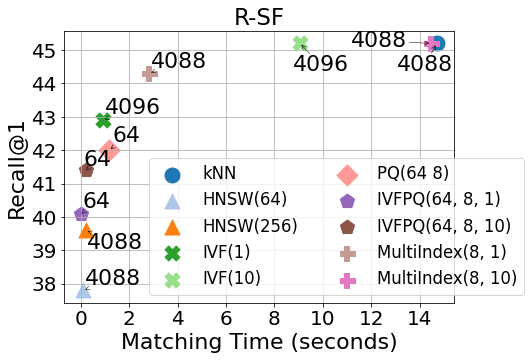}
            \centering\subcaption{}
        \label{fig:knn_indexing}
        \end{subfigure}
    \end{minipage}
    \caption{
        (a) \textbf{Matching time for one query.} 
        The plot shows, with exact search, linear dependency on database size and features dimensionality. The red line marks the extraction time of an image for ResNet-101 + GeM; above, the bottleneck is matching time, below it is extraction time. As a rule of thumb, kNN is the bottleneck if database size times the features dimension exceeds 200M.\\
        (b) Analysis of the \textbf{Recall-Speed-Memory trade-off} using optimized indexing techniques for neighbor search. Dots refer to a ResNet-50 + GeM (feat. dim. 1024) trained on Pitts30k. On the x axis is matching time in seconds for all queries in the dataset, on the y axis recall@1. The numbers next to the dots represent the RAM requirements in MB.
    }
    \vspace{-0.4cm}
\end{figure}

\subsection{Nearest Neighbor Search and Inference Time}
\label{sec:nearest_neighbor}
In practical applications, one of the most relevant factors for a VG system is inference time ($t_i$). Once the application is deployed and has to serve the user's needs, the perceived delay depends only on $t_i$. 
Inference time can be divided into: i) \textbf{extraction time} ($t_e$), defined as the elapsed time to extract the features of an image, which solely depends on model and resolution; 
ii) \textbf{matching time} ($t_m$), \ie, duration of the kNN to find the best matches in the database, which depends on the parameter $k$ (\ie number of candidates), the size of the database, the dimension of the descriptors, and the type of searching algorithm.

In \cref{fig:matching_time}, we report a plot on how matching time linearly depends on the sizes of the database and descriptors. \Cref{fig:knn_indexing} shows how the use of efficient nearest neighbor search algorithm impacts computation and memory footprint.
Besides exhaustive kNN, we investigate the use of inverted file indexes (IVF) \cite{Sivic-2003}, product quantization without and with IVF (PQ and IVFPQ) \cite{Jegou-2011_productQ}, inverted multi index (MultiIndex) \cite{BabenkoL12} and hierarchical navigable small world graphs (HNSW) \cite{Malkov2020EfficientAR}.
In \cref{fig:knn_indexing} we report results computed with a ResNet-50 + GeM descriptors on R-SF.
See more experiments and thorough discussions in \cref{sec:supp_knn_indexing}.

\paragraph{Discussion}
\Cref{fig:matching_time} shows that as the database grows, inference time is dominated by matching time whereas the extraction time is generally fixed at around 10 milliseconds (see \cref{tab:t1_backbones}).
On the other hand, \cref{fig:knn_indexing} shows that the choice of  neighbor search algorithm can bring huge benefits on time and memory footprint, with little to no loss in recalls.
Among the most interesting results, IVFPQ reduces both matching time and memory footprint by 98.5\%, with a drop in accuracy from 45.4\% to 41.4\%. 
Note that memory footprint is an important factor in image retrieval, since for fast computation all vectors should be kept in RAM, making large scale VG application expensive in terms of memory.
For example, R-SF dataset's descriptors, with a ResNet-50 + NetVLAD, require roughly $1.05\textrm{M} \cdot 65536 \cdot 4\textrm{B} = 256\textrm{GB}$ of memory, thus making a RAM-efficient search technique (\eg product quantization) very useful.
When memory is not a critical constraint, using a MultiIndex yields the same RAM occupancy but provides an 80\% saving in matching time, losing only a 0.9 \% of recall. These observations make the use of exact search hardly justified and prove that (i) recall should not be the only metric considered and (ii) for practical applications, the optimization of the neighbor search is a crucial factor that cannot be ignored.

\section{Discussions and Findings}

\begin{table}[t!]
  \centering
  \resizebox{\columnwidth}{!}{
  \begin{tabular}{lcccc}
    \toprule
    \multicolumn{1}{c}{\begin{tabular}[c]{@{}l@{}}Method \end{tabular}}
    & \multicolumn{1}{c}{\begin{tabular}[c]{@{}l@{}} Feat. Dim.\end{tabular}} 
    & \begin{tabular}[c]{@{}l@{}}R@1 \\ Pitts30k \end{tabular} & \begin{tabular}[c]{@{}l@{}}R@1 \\ Pitts250k\end{tabular} & \begin{tabular}[c]{@{}l@{}}R@1 \\ Tokyo 24/7\end{tabular} \\
    \hline
    VGG16 + NetVLAD + PCA \cite{Peng_2021_appsvr} &  4096 & 85.2 & 86.5 & 68.9 \\
    VGG16 + NetVLAD \cite{Peng_2021_appsvr} &  32768 & - & 84.1 & 60.0 \\
    SRALNet (ICRA21) \cite{Peng_2021_sralNet} &  4096 &   -  & 87.8 & 72.1  \\
    SRALNet (ICRA21) \cite{Peng_2021_sralNet} & 32768 & 85.1 & 85.8 & 68.6  \\
    APPSVR (ICCV21) \cite{Peng_2021_appsvr}  &  4096 & 87.4 & 88.8 & 77.1  \\
    APPSVR (ICCV21) \cite{Peng_2021_appsvr}  & 32768 &   -  & 86.6 & 68.3  \\\midrule
    ResNet-18 + NetVLAD + PCA (Ours) &  4096 & 86.8 & 87.9 & 72.2 \\
    ResNet-18 + NetVLAD  (Ours) & 16384 & 87.2 & 88.1 & 73.7 \\
    \bottomrule
  \end{tabular}}
  \caption{Comparison between recent SOTA methods, and a simple ResNet-18+NetVLAD where we use all the insight gained from the benchmark to find its optimal configuration: 
  training with data augmentation, resize 80\%, and majority voting post-processing for Tokyo 24/7 (since queries have different resolutions). 
  }
  \label{tab:comparison_with_sota}
  \vspace{-0.4cm}
\end{table}

This work introduces a modular framework that allows to build, train and test a wide range of  VG architectures, with the flexibility to change each component of a geo-localization pipeline.
Our experiments provide valuable insights on how different engineering choices implemented at training and test time can affect both the performance and the required resources (FLOPs, storage, time). 

\textbf{Architecture.}
We found that ResNet-50 is an excellent choice as a CNN backbone, yielding close to the best results at a reasonable cost. We also demonstrate for the first time the use of Visual Transformers for VG and find that they provide compelling results compared to their CNN counterparts.
Among them, CCT is particularly interesting because it is incredibly lightweight, with a cost comparable to a ResNet-18, but it performs better than a heavier ResNet-50. 
Regarding the feature aggregation layers, the best performance is generally obtained with CRN, nevertheless requiring a significant training cost. At the same time, the GeM pooling, which is much more efficient, has shown a better generalization power, especially when training the model on a large and heterogeneous dataset. The best results overall are obtained with CCT combined with NetVLAD.

\textbf{Negative mining.}  In general for metric learning for retrieval, negative mining is a crucial element. This was confirmed by our experiments, where we have additionally shown that partial mining can yield similar or sometimes even better performance than full mining, but at a fraction of the (computational) cost.

\textbf{Training dataset.} Unsurprisingly, using a large-scale training set, with a wide range of conditions and collected from very diverse cities, leads to significantly better results. This confirms the importance of the training set and the evidence that comparisons amongst models trained on different datasets, as commonly done in many papers \cite{Kim-2017, Zaffar-2021}, are not fair and should be avoided if possible.

\textbf{Image size and data augmentation.} 
As usually observed for deep models, data augmentation generally helps. In our case  we found that the effectiveness of the color jittering augmentations are highly dependent on the dataset, while  horizontal flipping and resized cropping provide a slight but consistent boost in all cases. Finally,  a surprising finding is that using the full resolution images (usually 480x640) is often superfluous -- scaling down the images to 60\% not only reduces the FLOPs, but on average yields comparable (and sometimes better) results.

\textbf{Inference time and kNN search.} 
We  propose an extensive  study for VG, unique in its kind, comparing  advanced kNN search algorithms and compact representations. This study  has shown that  the choice of a good neighbor search algorithm can have a huge impact on time and memory footprint, with little impact on the performance. Furthermore, we observe that advanced kNN methods might nullify the gap in terms of both memory footprint and matching time between larger and smaller descriptors.

\textbf{Final remarks} All the above insights are important to design and optimize VG architectures depending on one's use case and requirements.
For instance, consider again the example from \cref{tab:vanilla_vs_modifications}. In light of the lessons learned, we can carefully optimize the same simple architecture to get results that are comparable with much more complex (yet not optimized) methods (see \cref{tab:comparison_with_sota}).

\textbf{Limitations}
Despite its modularity and versatility, our framework has also some limitations, \eg, it is focused on VG methods in outdoor urban environments, it only addresses the task of Visual Geo-localization from a single image, it does not try to analyze the viewpoint and luminosity invariance of the methods (as done in \cite{Zaffar-2021}).
Furthermore, some recent SOTA works \cite{Ge-2020, Peng_2021_appsvr} are not implemented yet, and some newer losses  not yet compared \cite{Liu-2019}.
However, we plan to continue supporting the software and website, expanding them to evaluate more techniques and use-cases and investigate additional elements in a VG pipeline. 

\noindent\textbf{Acknowledgements}
%\small{
We acknowledge the CINECA award under the ISCRA initiative, for the availability of high performance computing resources and support.
Also, computational resources were provided by HPC@POLITO, a project of Academic Computing within the Department of Control and Computer Engineering at the Politecnico di Torino (http://www.hpc.polito.it).
This work was partially supported by CINI, the European Regional Development Fund under project IMPACT (reg. no. CZ.02.1.01/0.0/0.0/15\_003/0000468), and the EU Horizon 2020 project RICAIP (grant agreement No 857306).
%}

% %%%%%%%%% REFERENCES
{\small
\bibliographystyle{ieee_fullname}
\bibliography{benchmark}
}

\appendix

\section*{Appendix}
This appendix contains additional information that could not fit within the main paper due to a lack of space:
\begin{itemize}
    \item[-] Section~\ref{sec:supp_datasets} describes in detail the datasets used in the benchmark.
    \item[-] Section~\ref{sec:supp_software} explains the organization of the open-source software that implements the benchmark.
    \item[-] Section~\ref{sec:supp_ext_results} provides extended results and discussions for the experiments presented in the main paper.
    \item[-] Section~\ref{sec:supp_additional_experiments} contains additional experiments and discussions that complement the tests presented in the main paper.
\end{itemize}

\section{Datasets}
\label{sec:supp_datasets}

% pitts30k
\textbf{Pitts30k} \cite{Arandjelovic-2018} is a subset of Pitts250k \cite{Torii-2015}, split in train, val and test set. It is collected from Google Street View imagery from the city of Pittsburgh cropping equirectangular panoramas into tiles, and applying a gnomonic projection to the tiles. Database and queries are collected two years apart, and there are no noticeable weather variations.

% msls
\textbf{Mapillary Street Level Sequence (MSLS)} \cite{Warburg-2020} spans multiple cities across six continents, covering a large variety of domains, cameras and seasons. As for Pitts30k, it is split in train, val and test set, although the test set's ground truths are not currently released. We therefore report the validation recalls, following previous works \cite{Hausler-2021}.
Only Pitts30k and MSLS provide a train set with temporal variability, which is necessary for training a VG model~\cite{Arandjelovic-2018}.

% tokyo 24/7
\textbf{Tokyo 24/7} \cite{Torii-2018} presents a relatively large database (from Google Street View) against a smaller number of queries, which are split into three equally sized sets: day, sunset and night.
The latter are manually collected with phones.
In some cases \cite{Arandjelovic-2018, Liu-2021, Yu-2020} Tokyo Time Machine (Tokyo TM) is used as a training set for Tokyo 24/7.

% san francisco
\textbf{San Francisco} \cite{Chen-2011}, similarly to Tokyo 24/7, is composed of a large database collected by a car-mounted camera and orders of magnitude less queries taken by phone.
Among the multiple Structure from Motion reconstructions available, we use the one from \cite{Torii-2021,Li-2012} as it offers the most accurate query 6~DoF coordinates, thus referring to it as Revisited San Francisco.

% eynsham
\textbf{Eynsham} \cite{Cummins-2009} consists of grayscale images from cameras mounted on a car going around around a loop twice, in the city and countryside of Oxford. We use the first loop as database, and second as queries. The cameras collected equirectangular panoramas, and each panorama was split in five crops.

% st lucia
\textbf{St Lucia} \cite{Milford-2008} is collected by driving a car with a forward facing camera around the riverside suburb of St Lucia, Brisbane. Of the nine drives, we use the first and the last one as database and queries. Given the high density of the images (extracted from videos), we select only one frame every 5 meters. Note that all these pre-processing steps (as well as downloading) are performed automatically with our open source codebase (see Section \ref{sec:supp_software}).

\begin{figure*}[tb]
    \centering
    \begin{minipage}{.32\textwidth}
        \begin{subfigure}{\textwidth}
            \includegraphics[width=0.85\textwidth]{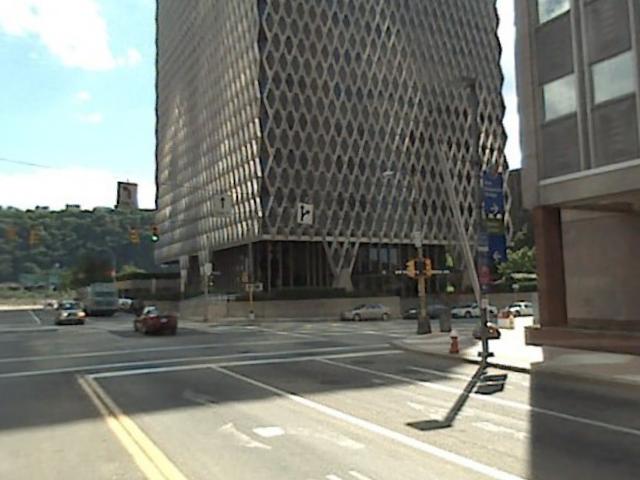}
            \includegraphics[width=0.85\textwidth]{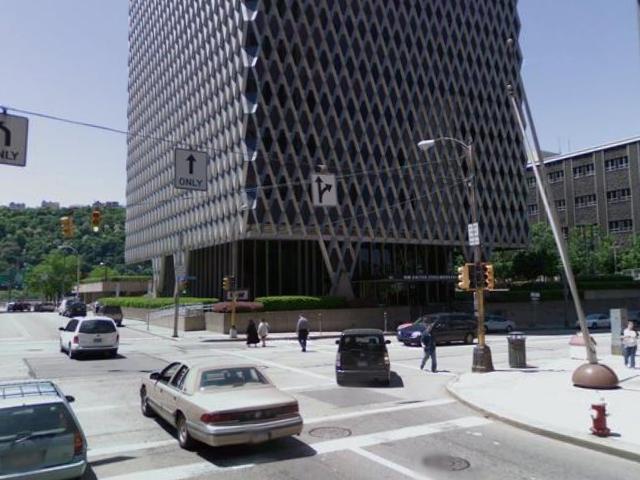}
            \centering\subcaption{Pitts30k}
            \vspace*{60px}
        \end{subfigure}
        \begin{subfigure}{\textwidth}
            \includegraphics[width=0.85\textwidth]{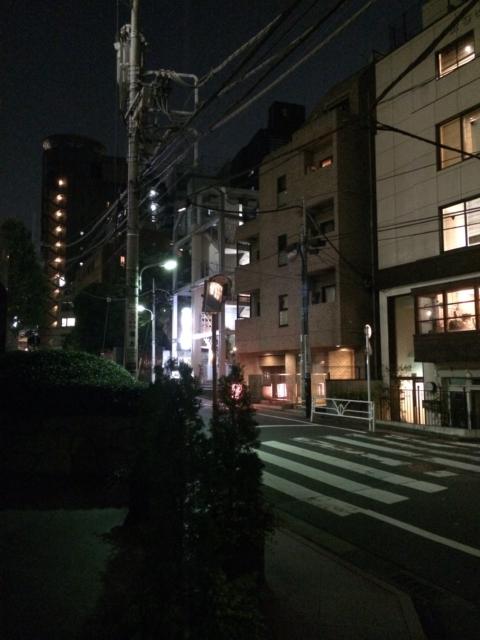}
            \includegraphics[width=0.85\textwidth]{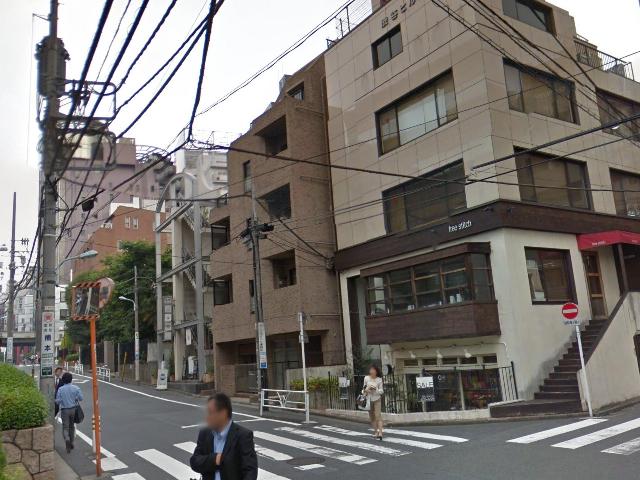}
            \centering\subcaption{Tokyo 24/7}
        \end{subfigure}
       
    \end{minipage}
    \begin{minipage}{.32\textwidth}
        \begin{subfigure}{\textwidth}
            \includegraphics[width=0.85\textwidth]{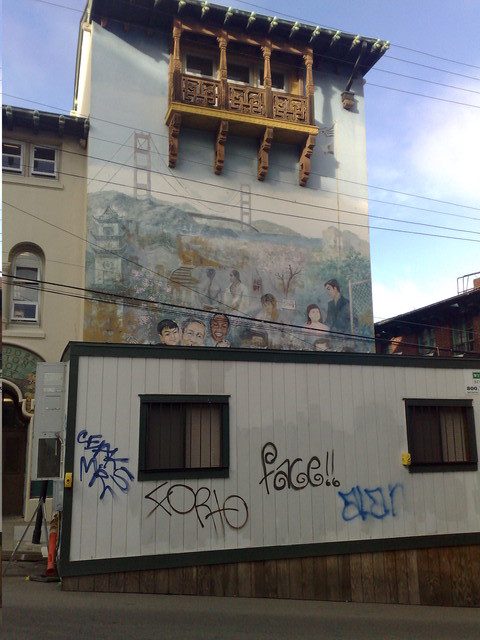} 
            \includegraphics[width=0.85\textwidth]{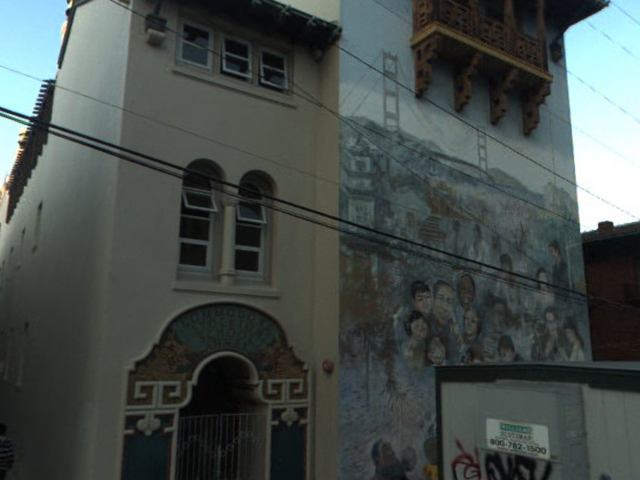} 
            \centering\subcaption{San Francisco}
            \vspace*{61px}
        \end{subfigure}
        \begin{subfigure}{\textwidth}
            \includegraphics[width=0.85\textwidth]{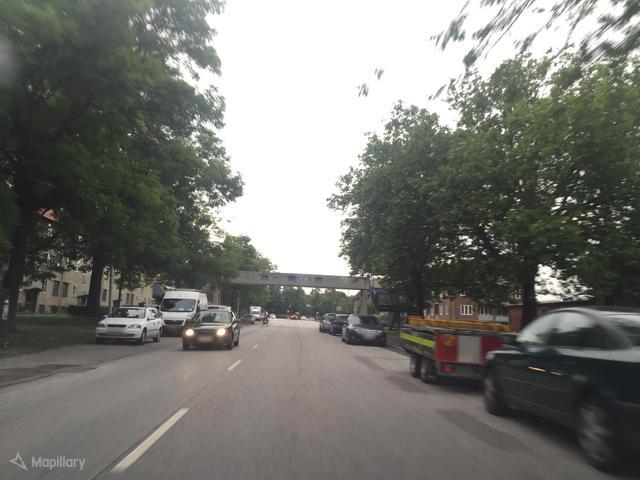} 
            \includegraphics[width=0.85\textwidth]{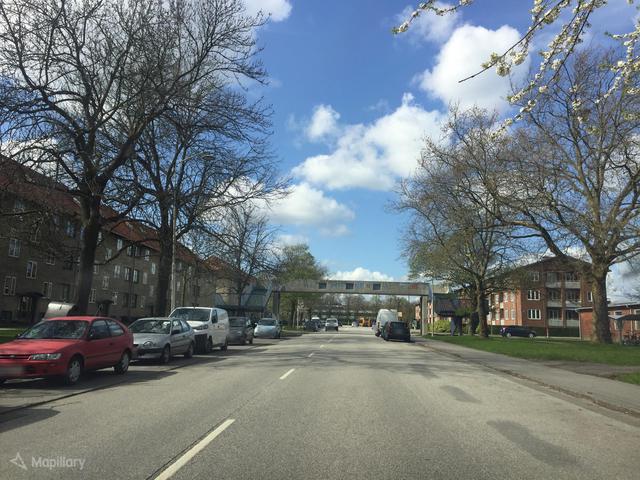} 
            \centering\subcaption{MSLS}
        \end{subfigure}
    \end{minipage}
    \begin{minipage}{.32\textwidth}
        \begin{subfigure}{\textwidth}
            \includegraphics[width=0.85\textwidth]{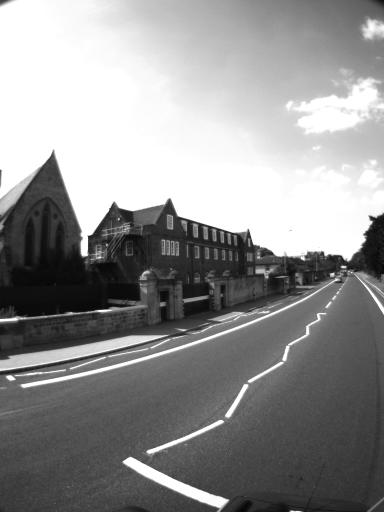} 
            \includegraphics[width=0.85\textwidth]{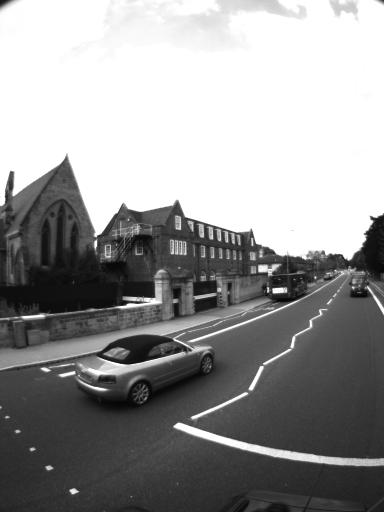} 
            \centering\subcaption{Eynsham}
          \vspace*{1px}
        \end{subfigure}
         \begin{subfigure}{\textwidth}
            \includegraphics[width=0.85\textwidth]{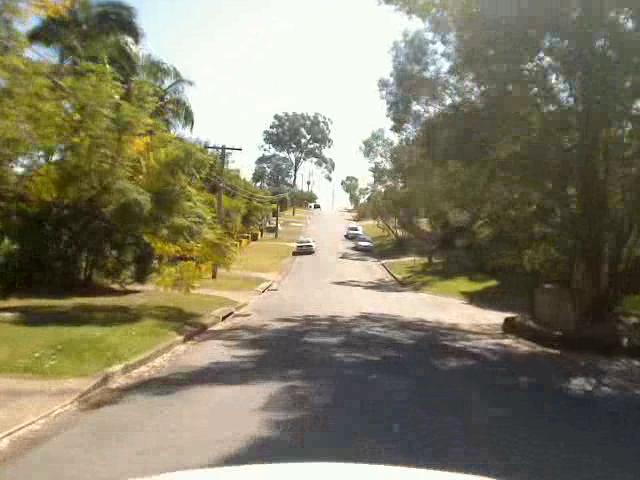}
            \includegraphics[width=0.85\textwidth]{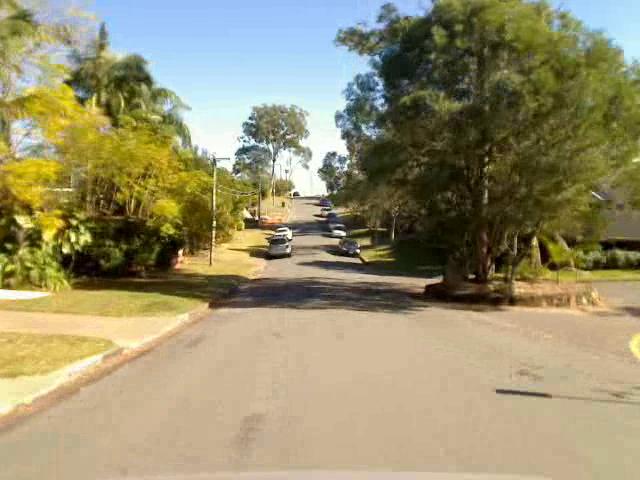}
            \centering\subcaption{St Lucia}
        \end{subfigure}
    \end{minipage}
    \caption{\textbf{Examples of a query and a positive} for each of the used dataset.}
    \label{fig:examples}
\end{figure*}

\begin{figure*}[tb]
    \centering
    \resizebox{0.85\textwidth}{!}{
        \begin{minipage}{.3\textwidth}
            \begin{subfigure}{\textwidth}
                \includegraphics[width=0.85\textwidth]{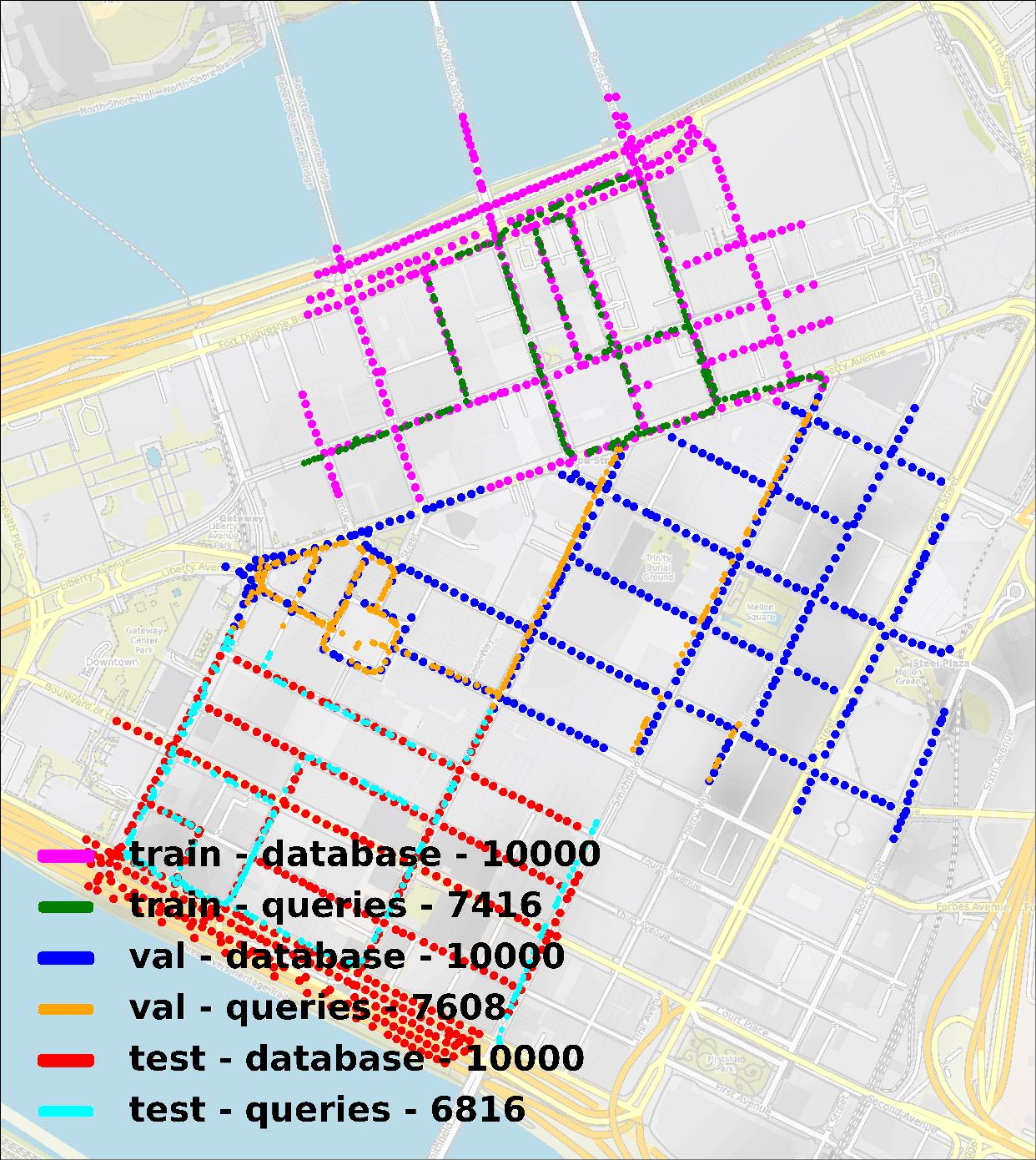}
                \centering\subcaption{Pitts30k}
                \vspace*{25px}
            \end{subfigure}
            \begin{subfigure}{\textwidth}
                \includegraphics[width=0.85\textwidth]{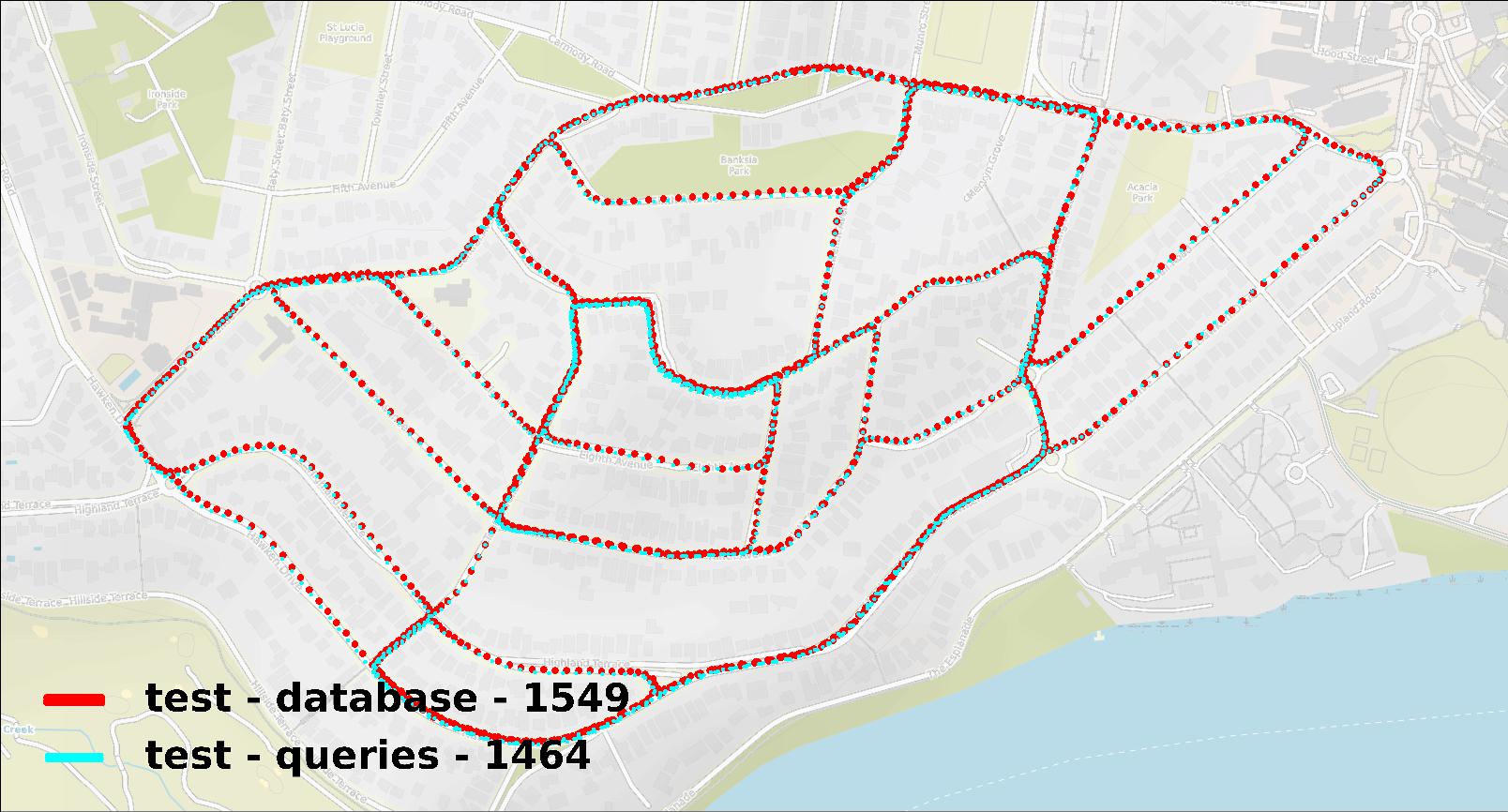}
                \centering\subcaption{St Lucia}
            \end{subfigure}
        \end{minipage}
        \begin{minipage}{.3\textwidth}
            \begin{subfigure}{\textwidth}
                \includegraphics[width=0.85\textwidth]{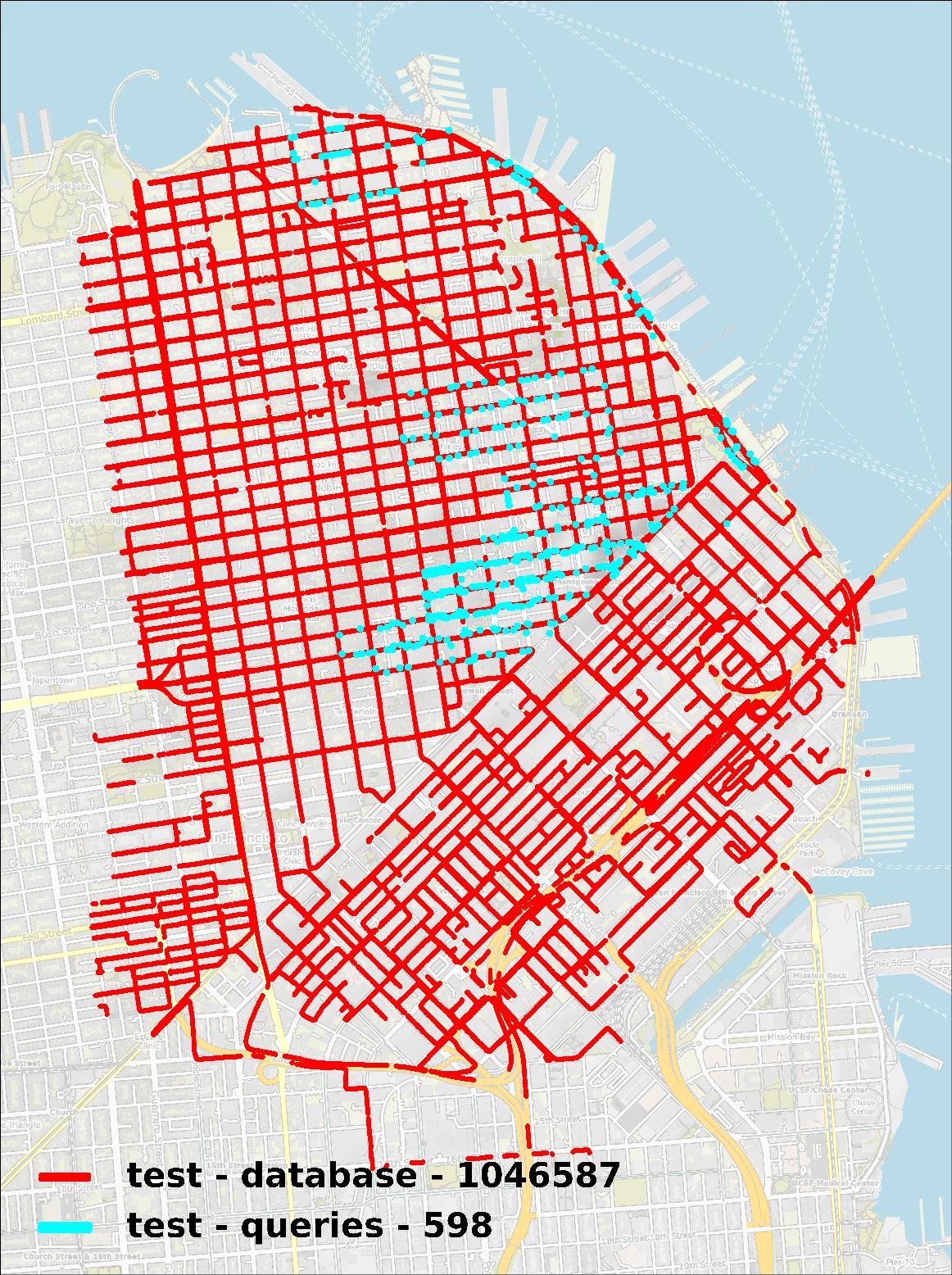} 
                \centering\subcaption{San Francisco}
            \end{subfigure}
            \begin{subfigure}{\textwidth}
                \includegraphics[width=0.85\textwidth]{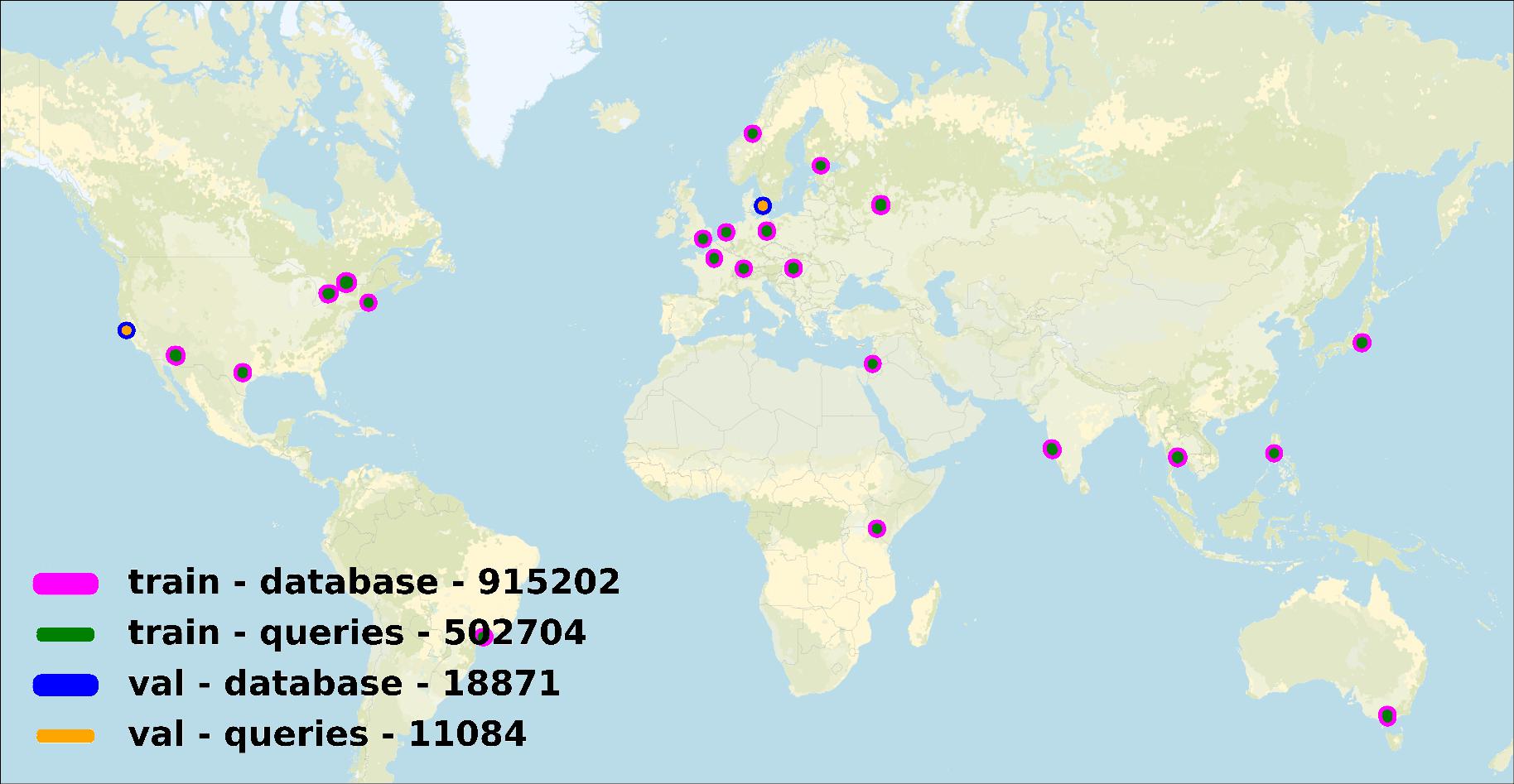} 
                \centering\subcaption{MSLS}
            \end{subfigure}
        \end{minipage}
        \begin{minipage}{.3\textwidth}
            \begin{subfigure}{\textwidth}
                \includegraphics[width=0.85\textwidth]{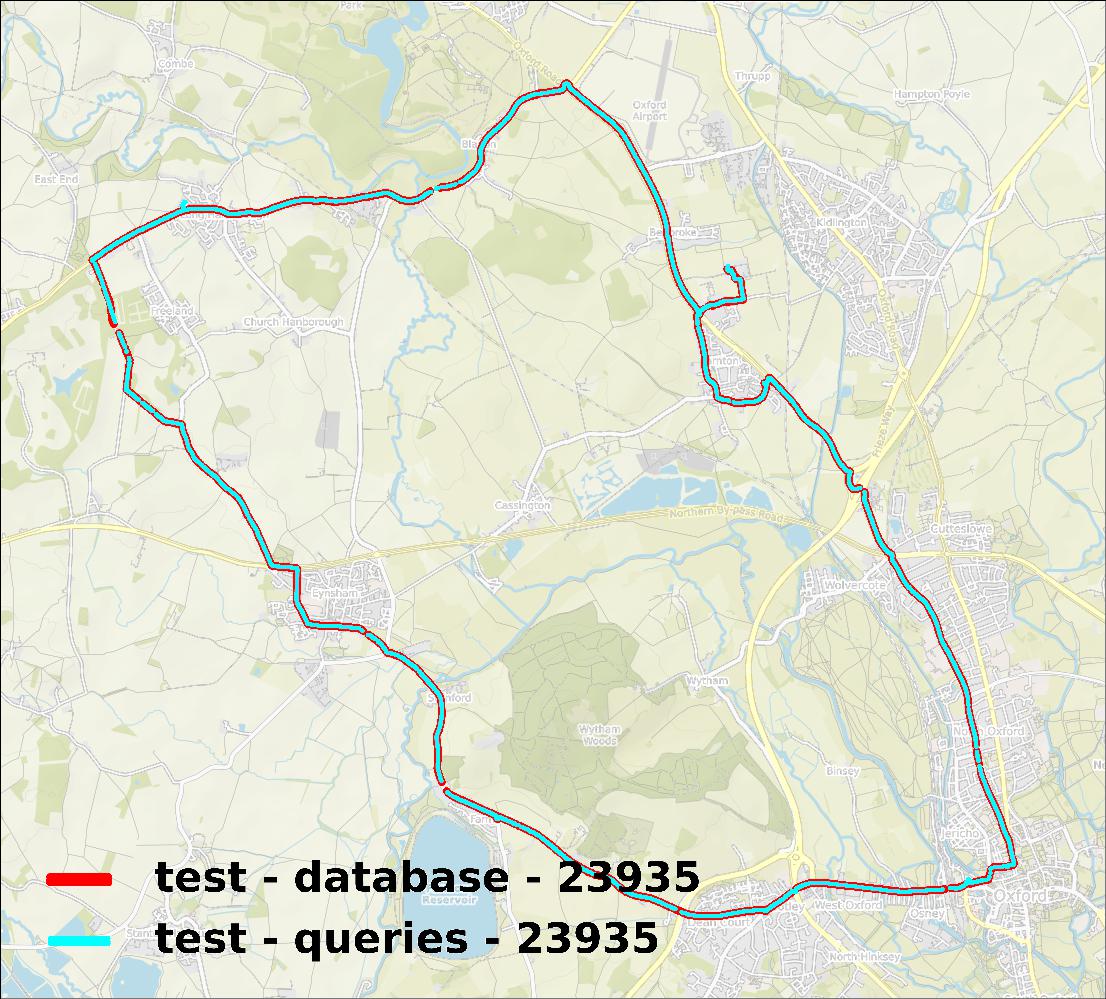} 
                \centering\subcaption{Eynsham}
              \vspace*{5px}
            \end{subfigure}
            \begin{subfigure}{\textwidth}
                \includegraphics[width=0.85\textwidth]{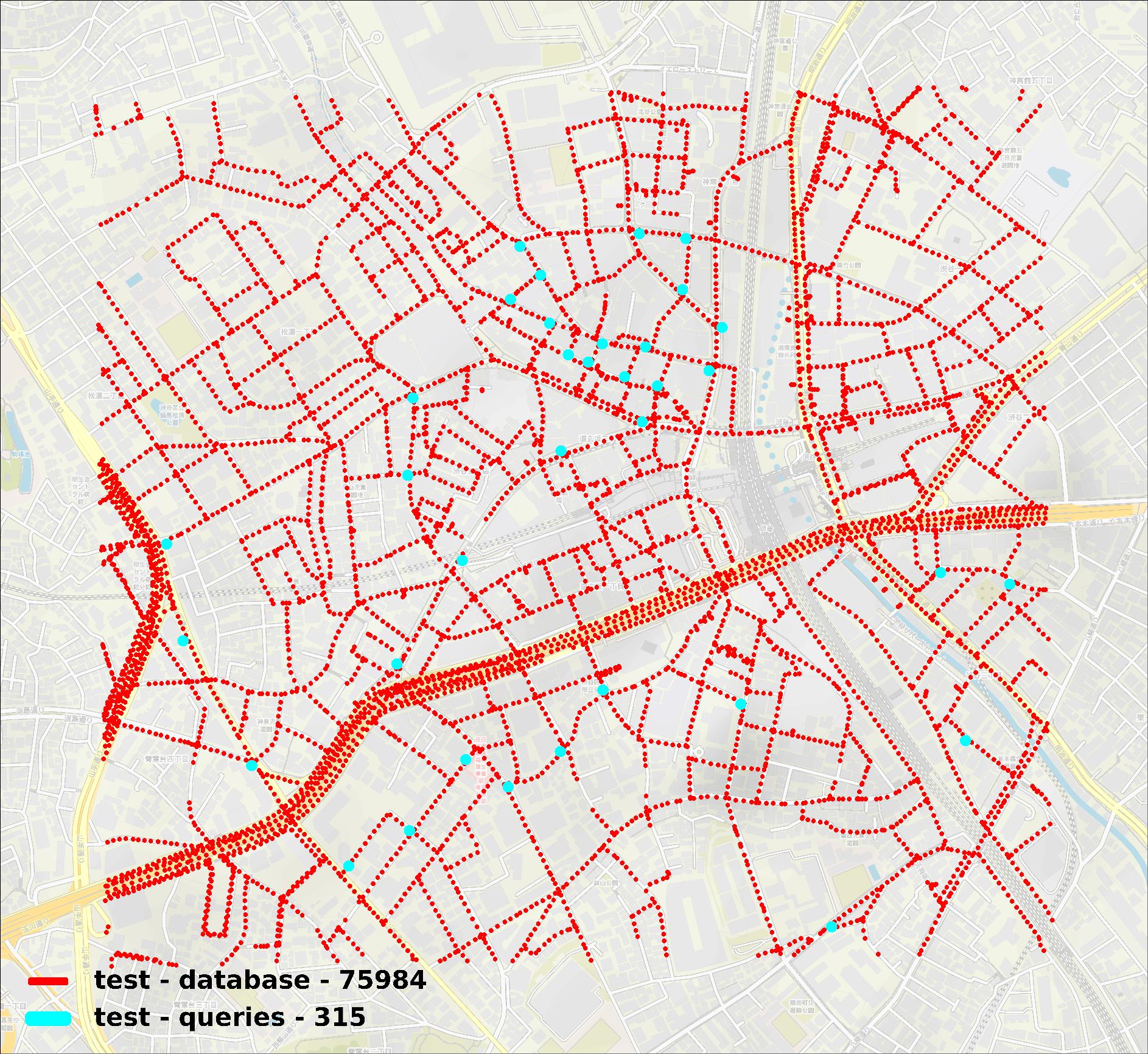}
                \centering\subcaption{Tokyo 24/7}
            \end{subfigure}
        \end{minipage}
    }
    \caption{\textbf{Maps of used datasets,} self-generated with our open source codebase.}
    \label{fig:maps}
     \vspace{-0.2cm}
\end{figure*}

In Fig. \ref{fig:examples} we show relevant query-database image  pairs from each of the used datasets. These examples illustrate view, environmental, and acquisition condition variability between query and database images as well as across the datasets making the generalization between datasets hard. In Tab. \ref{tab:datasets_size} we provide a summary of the number of database and query images as well as the area and perimeter covered by the respective datasets. \Cref{fig:maps} shows the density of the images in the respective geographical areas.   

\begin{table*}[htb]
  \centering
  \resizebox{.8\textwidth}{!}{
  \begin{tabular}{lccccccccccccc}
    \toprule
    \textbf{} & \multicolumn{1}{l}{\textbf{\begin{tabular}[c]{@{}c@{}}\# database \\ \end{tabular}}} 
    & \multicolumn{1}{l}{\textbf{\begin{tabular}[c]{@{}c@{}}\# queries \\ \end{tabular}}} 
    & \multicolumn{1}{l}{\textbf{\begin{tabular}[c]{@{}c@{}}Dataset \\ size\end{tabular}}} 
    & \multicolumn{1}{l}{\textbf{\begin{tabular}[c]{@{}c@{}}Area\\(Km\textsuperscript{2}) \end{tabular}}} 
    & \multicolumn{1}{l}{\textbf{\begin{tabular}[c]{@{}c@{}}Perimeter\\(Km) \end{tabular}}}
    & \multicolumn{1}{c}{\textbf{\begin{tabular}[c]{@{}c@{}}Environment\\ \end{tabular}}}
    & \multicolumn{1}{l}{\textbf{\begin{tabular}[c]{@{}c@{}}Day/night\\changes \end{tabular}}}
    & \multicolumn{1}{l}{\textbf{\begin{tabular}[c]{@{}c@{}}Long-term\\variations \end{tabular}}}
    \\
    \midrule
    Pitts30k   & 30K   & 21.8K & 2.0 GB & 0.615 & 3.42 & Urban            & N & Y \\
    MSLS       & 973K  & 541K  & 56 GB  & N/A   & N/A  & Urban + Suburban & Y & Y \\
    Tokyo 24/7 & 75K   & 315   & 4.0 GB & 2.1   & 5.8  & Urban            & Y & Y \\
    R-SF       & 1.05M & 598   & 36 GB  & 13.6  & 14.0 & Urban            & N & Y \\
    Eynsham    & 24K   & 24K   & 1.2 GB & N/A   & N/A  & Urban + Suburban & N & N \\
    St Lucia   & 1.5K  & 1.5K  & 124 MB & 0.69  & 3.5  & Suburban         & N & N \\
    \bottomrule
  \end{tabular}}
  \caption{\textbf{Summary of datasets used.} Long-term variations refers to images taken at least one year apart.}
  \label{tab:datasets_size}
\end{table*}

\section{Software} \label{sec:supp_software}
We aim to create and maintain an organized open-source repository where existing and new VG methods will be integrated in the future. Our site \footnote{\url{https://deep-vg-bench.herokuapp.com/}}  will be used to show the performances of these methods with different VG datasets.

Following these main motivations, we designed the software aiming to create a modular and easy expandable framework that provides the users with a common playground (i) to train, test, and fairly compare the impact of different components of a VG model, (ii) to ease the reproducibility of the results, and (iii) to evaluate the performances with datasets of different scales.

We organized the software into three distinct modules:
\begin{itemize}
    \item[-]\texttt{ benchmarking\_vg}:  a general and expandable template for training and evaluating VG models; 
    \item[-]\texttt{ dataset\_vg}: a dataset utility to automatically download most of the datasets described in Section \ref{sec:supp_datasets} and format them according to a standardized methodology;
    \item[-]\texttt{ pretrain\_vg}: a template to pretrain neural networks backbones used in the VG task.
\end{itemize}

We mainly consider VG techniques that tackle the Visual Geo-localization problem through an image retrieval approach using Deep Learning (DL).
For this reason, the first and main module (\texttt{benchmarking\_vg}) of our framework follows a common structure for all the models, which are composed of a neural network backbone and a pooling layer on top. We integrated existing PyTorch open-source implementations of VG models or self-implemented them when they were unavailable. For the similarity search we use the implementations from the highly optimized FAISS~\cite{Johnson-2017} library. Unless otherwise specified, we use an exhaustive kNN search.
Further techniques can be easily integrated and work under our environment. 

The \texttt{benchmarking\_vg} module further allows the user to choose which VG model, training dataset, and mining techniques to use to evaluate its performance. Even external trained models can be loaded and evaluated on VG datasets. \Cref{sec:supp_role_training_dataset} shows the results obtained by integrating the models of \cite{Radenovic-2019} into our framework.

The \texttt{pretrain\_vg} module constitutes a template to pretrain backbones on the Landmark Recognition and Classification datasets. In the current version, the Google Landmark v2 dataset \cite{Weyand-2020} and  Places 365 \cite{Zhou-2017} are available.

\section{Extended Results} 
\label{sec:supp_ext_results}

%%%%%%%%%%%%%%%%%%%%%%%%%%%%%%%% BACKBONES
%%%%%%%%%%%%%%%%%%%%%%%%%%%%%%%% BACKBONES
%%%%%%%%%%%%%%%%%%%%%%%%%%%%%%%% BACKBONES

\begin{table*}[htb]
  \centering
  \resizebox{\textwidth}{!}{
   \begin{tabular}{llllllllllll}
    \toprule
    \multirow{2}{*}{Backbone} & 
    \begin{tabular}[c]{@{}l@{}}Aggregation\\ Method\end{tabular} & \begin{tabular}[c]{@{}l@{}}Features\\ Dim\end{tabular} & 
    \multirow{2}{*}{FLOPs} &
    \begin{tabular}[c]{@{}l@{}}Model\\ Size\end{tabular} &
    \begin{tabular}[c]{@{}l@{}}Training \\ Dataset\end{tabular} & \begin{tabular}[c]{@{}l@{}}R@1\\ Pitts30k\end{tabular} & \begin{tabular}[c]{@{}l@{}}R@1\\ MSLS\end{tabular} & \begin{tabular}[c]{@{}l@{}}R@1 \\ Tokyo 24/7\end{tabular} & \begin{tabular}[c]{@{}l@{}}R@1\\ R-SF\end{tabular} & \begin{tabular}[c]{@{}l@{}}R@1\\ Eynsham\end{tabular} & \begin{tabular}[c]{@{}l@{}}R@1\\ St Lucia\end{tabular}  \\
    \midrule
ResNet-18 \emph{conv4\_x}        &GeM      &256   &17.29 GF  &10.63 MB  &Pitts30k & 77.8  \footnotesize{$\pm$ 0.2}&35.3  \footnotesize{$\pm$ 0.5}&35.3  \footnotesize{$\pm$ 1.1}&34.2  \footnotesize{$\pm$ 1.7}&64.3  \footnotesize{$\pm$ 1.2}&46.2  \footnotesize{$\pm$ 0.4}\\
ResNet-18 \emph{conv4\_x}        &NetVLAD  &16384 &17.27 GF  &10.76 MB  &Pitts30k & \bf{86.4}  \footnotesize{$\pm$ 0.3}&\bf{47.4}  \footnotesize{$\pm$ 1.2}&\bf{63.4}  \footnotesize{$\pm$ 1.2}&\bf{61.4}  \footnotesize{$\pm$ 1.5}&\bf{76.8}  \footnotesize{$\pm$ 1.2}&\bf{57.6}  \footnotesize{$\pm$ 3.3}\\
ResNet-18 \emph{conv5\_x}        &GeM      &512   &22.33 GF  &42.67 MB  &Pitts30k & 77.9  \footnotesize{$\pm$ 0.3}&34.4  \footnotesize{$\pm$ 0.4}&34.4  \footnotesize{$\pm$ 0.6}&36.9  \footnotesize{$\pm$ 0.3}&59.1  \footnotesize{$\pm$ 1.3}&51.2  \footnotesize{$\pm$ 1.3}\\
ResNet-18 \emph{conv5\_x}        &NetVLAD  &32768 &22.28 GF  &42.92 MB  &Pitts30k & 79.6  \footnotesize{$\pm$ 0.5}&47.1  \footnotesize{$\pm$ 1.8}&48.9  \footnotesize{$\pm$ 2.5}&49.1  \footnotesize{$\pm$ 3.6}&70.5  \footnotesize{$\pm$ 1.0}&54.4  \footnotesize{$\pm$ 2.7}\\
\midrule
ResNet-50 \emph{conv4\_x}        &GeM      &1024  &40.61 GF  &32.71 MB  &Pitts30k & 82.0  \footnotesize{$\pm$ 0.3}&38.0  \footnotesize{$\pm$ 0.1}&41.5  \footnotesize{$\pm$ 1.8}&45.4  \footnotesize{$\pm$ 2.0}&66.3  \footnotesize{$\pm$ 2.5}&59.0  \footnotesize{$\pm$ 1.4}\\
ResNet-50 \emph{conv4\_x}        &NetVLAD  &65536 &40.51 GF  &33.21 MB  &Pitts30k & \bf{86.0}  \footnotesize{$\pm$ 0.1}&\bf{50.7}  \footnotesize{$\pm$ 2.0}&\bf{69.8}  \footnotesize{$\pm$ 0.8}&\bf{67.1}  \footnotesize{$\pm$ 2.3}&\bf{77.7}  \footnotesize{$\pm$ 0.4}&\bf{60.2}  \footnotesize{$\pm$ 1.6}\\
ResNet-50 \emph{conv5\_x}        &GeM      &2048  &50.54 GF  &89.88 MB  &Pitts30k & 79.8  \footnotesize{$\pm$ 0.5}&41.5  \footnotesize{$\pm$ 0.7}&48.0  \footnotesize{$\pm$ 2.5}&44.3  \footnotesize{$\pm$ 1.0}&65.2  \footnotesize{$\pm$ 1.4}&57.5  \footnotesize{$\pm$ 1.5}\\
ResNet-50 \emph{conv5\_x}        &NetVLAD  &131072&50.35 GF  &90.88 MB  &Pitts30k & 79.6  \footnotesize{$\pm$ 0.2}&46.2  \footnotesize{$\pm$ 0.5}&54.7  \footnotesize{$\pm$ 2.6}&51.2  \footnotesize{$\pm$ 2.5}&69.8  \footnotesize{$\pm$ 1.0}&53.0  \footnotesize{$\pm$ 4.1}\\
\midrule\midrule
ResNet-18 \emph{conv4\_x}        &GeM      &256   &17.29 GF  &10.63 MB  &MSLS     & 71.6  \footnotesize{$\pm$ 0.1}&65.3  \footnotesize{$\pm$ 0.2}&42.8  \footnotesize{$\pm$ 1.1}&30.5  \footnotesize{$\pm$ 0.8}&80.3  \footnotesize{$\pm$ 0.1}&83.2   \footnotesize{$\pm$ 0.9}\\
ResNet-18 \emph{conv4\_x}        &NetVLAD  &16384 &17.27 GF  &10.76 MB  &MSLS     & \bf{81.6}  \footnotesize{$\pm$ 0.5}&\bf{75.8}  \footnotesize{$\pm$ 0.1}&\bf{62.3}  \footnotesize{$\pm$ 1.6}&\bf{55.1}  \footnotesize{$\pm$ 0.9}&\bf{87.1}  \footnotesize{$\pm$ 0.2}&\bf{92.1}   \footnotesize{$\pm$ 0.7}\\
ResNet-18 \emph{conv5\_x}        &GeM      &512   &22.33 GF  &42.67 MB  &MSLS     & 73.5  \footnotesize{$\pm$ 0.5}&68.4  \footnotesize{$\pm$ 0.8}&41.0  \footnotesize{$\pm$ 0.8}&38.6  \footnotesize{$\pm$ 1.8}&79.4  \footnotesize{$\pm$ 0.5}&84.7   \footnotesize{$\pm$ 0.7}\\
ResNet-18 \emph{conv5\_x}        &NetVLAD  &32768 &22.28 GF  &42.92 MB  &MSLS     & 75.7  \footnotesize{$\pm$ 0.7}&75.7  \footnotesize{$\pm$ 0.6}&49.9  \footnotesize{$\pm$ 1.6}&41.3  \footnotesize{$\pm$ 0.2}&84.1  \footnotesize{$\pm$ 0.4}&91.3   \footnotesize{$\pm$ 0.4}\\
\midrule
ResNet-50 \emph{conv4\_x}        &GeM      &1024  &40.61 GF  &32.71 MB  &MSLS     & 77.4  \footnotesize{$\pm$ 0.6}&72.0  \footnotesize{$\pm$ 0.5}&55.4  \footnotesize{$\pm$ 2.5}&45.7  \footnotesize{$\pm$ 1.0}&83.9  \footnotesize{$\pm$ 0.6}&91.2   \footnotesize{$\pm$ 0.7}\\
ResNet-50 \emph{conv4\_x}        &NetVLAD  &65536 &40.51 GF  &33.21 MB  &MSLS     & \bf{80.9}  \footnotesize{$\pm$ 0.0}&\bf{76.9}  \footnotesize{$\pm$ 0.2}&\bf{62.8}  \footnotesize{$\pm$ 0.9}&\bf{51.5}  \footnotesize{$\pm$ 1.2}&\bf{87.2}  \footnotesize{$\pm$ 0.3}&\bf{93.8}   \footnotesize{$\pm$ 0.2}\\
ResNet-50 \emph{conv5\_x}        &GeM      &2048  &50.54 GF  &89.88 MB  &MSLS     & 74.7  \footnotesize{$\pm$ 0.4}&70.6  \footnotesize{$\pm$ 0.6}&46.3  \footnotesize{$\pm$ 1.3}&42.1  \footnotesize{$\pm$ 0.5}&82.5  \footnotesize{$\pm$ 0.5}&89.8   \footnotesize{$\pm$ 0.4}\\
ResNet-50 \emph{conv5\_x}        &NetVLAD  &131072&50.35 GF  &90.88 MB  &MSLS     & 74.7  \footnotesize{$\pm$ 0.2}&75.2  \footnotesize{$\pm$ 0.5}&52.4  \footnotesize{$\pm$ 0.8}&44.0  \footnotesize{$\pm$ 1.1}&85.5  \footnotesize{$\pm$ 0.4}&91.3   \footnotesize{$\pm$ 0.7}\\
    \bottomrule
  \end{tabular}}
    \caption{\textbf{ResNets:} The advantages of cropping the ResNets at \emph{conv4\_x} for visual geo-localization.}
  \label{tab:t1_backbones_resnets}
\end{table*}
\subsection{CNN Backbones}
In Tab. \ref{tab:t1_backbones_resnets} we show comparative results obtained by cropping a ResNet-18 and a ResNet-50 to the \emph{conv4\_x} layer (used in the experiments in the main paper) or alternatively  cropping to the \emph{conv5\_x} (refer to the ResNets paper \cite{He-2016} for more details on the layers). 
We see that on average cropping the ResNet backbone at the lower level \emph{conv4\_x} leads to better results while being somewhat lighter in size.

\begin{table*}[htp!]
  \centering
  \resizebox{\textwidth}{!}{
  \begin{tabular}{llllllllll}
    \toprule
    \multirow{2}{*}{Backbone} & 
    \begin{tabular}[c]{@{}l@{}}Aggregation\\ Method\end{tabular} &
    \begin{tabular}[c]{@{}l@{}}Features\\ Dim\end{tabular} & 
    \begin{tabular}[c]{@{}l@{}}Training \\ Dataset\end{tabular} & \begin{tabular}[c]{@{}l@{}}R@1\\ Pitts30k\end{tabular} & \begin{tabular}[c]{@{}l@{}}R@1\\ MSLS\end{tabular} & \begin{tabular}[c]{@{}l@{}}R@1 \\ Tokyo 24/7\end{tabular} & \begin{tabular}[c]{@{}l@{}}R@1\\ R-SF\end{tabular} & \begin{tabular}[c]{@{}l@{}}R@1\\ Eynsham\end{tabular} & \begin{tabular}[c]{@{}l@{}}R@1\\ St Lucia\end{tabular}  \\
    \midrule
ResNet-18   &SPOC \cite{Babenko-2015}        &256   &Pitts30k & 60.6  \footnotesize{$\pm$ 0.9}&16.5  \footnotesize{$\pm$ 0.5}&15.2  \footnotesize{$\pm$ 1.1}&10.4  \footnotesize{$\pm$ 0.3}&41.0  \footnotesize{$\pm$ 2.0}&29.0  \footnotesize{$\pm$ 1.5}\\
ResNet-18   &MAC \cite{Razavian-2015}        &256   &Pitts30k & 57.3  \footnotesize{$\pm$ 0.5}&25.6   \footnotesize{$\pm$ 0.4}&15.2  \footnotesize{$\pm$ 1.3}&15.5  \footnotesize{$\pm$ 0.3}&49.6  \footnotesize{$\pm$ 0.7}&26.6  \footnotesize{$\pm$ 1.0}\\
ResNet-18   &RMAC \cite{Tolias-2016}         &256   &Pitts30k & 63.2  \footnotesize{$\pm$ 0.4}&28.7   \footnotesize{$\pm$ 0.6}&22.7  \footnotesize{$\pm$ 2.3}&30.5   \footnotesize{$\pm$ 1.4}&64.0  \footnotesize{$\pm$ 0.7}&42.8  \footnotesize{$\pm$ 1.3}\\
ResNet-18   &RRM   \cite{Kordopatis-2021}         &256   &Pitts30k & 68.2  \footnotesize{$\pm$ 0.5}&21.4  \footnotesize{$\pm$ 0.8}&25.4  \footnotesize{$\pm$ 1.4}&21.7  \footnotesize{$\pm$ 1.8}&51.9  \footnotesize{$\pm$ 0.8}&33.7  \footnotesize{$\pm$ 0.3}\\
ResNet-18   &GeM \cite{Radenovic-2019}       &256   &Pitts30k & 77.8  \footnotesize{$\pm$ 0.2}&35.3  \footnotesize{$\pm$ 0.5}&35.3  \footnotesize{$\pm$ 1.1}&34.2  \footnotesize{$\pm$ 1.7}&64.3  \footnotesize{$\pm$ 1.2}&46.2  \footnotesize{$\pm$ 0.4}\\
ResNet-18   &GeM + FC 256                     &256   &Pitts30k & 72.4  \footnotesize{$\pm$ 0.7}&26.4  \footnotesize{$\pm$ 0.5}&27.5  \footnotesize{$\pm$ 1.2}&29.0  \footnotesize{$\pm$ 1.2}&59.3  \footnotesize{$\pm$ 1.0}&39.1  \footnotesize{$\pm$ 0.8}\\
ResNet-18   &NetVLAD + PCA 256                &256   &Pitts30k & 80.7  \footnotesize{$\pm$ 0.7}&38.3  \footnotesize{$\pm$ 1.2}&41.7  \footnotesize{$\pm$ 0.8}&35.9  \footnotesize{$\pm$ 1.8}&68.9  \footnotesize{$\pm$ 1.1}&45.4  \footnotesize{$\pm$ 2.2}\\
ResNet-18	&CRN + PCA 256                    &256   &Pitts30k & \bf{82.0}  \footnotesize{$\pm$ 0.7}&\bf{43.6}  \footnotesize{$\pm$ 0.7}&\bf{47.7}  \footnotesize{$\pm$ 0.9}&\bf{45.1}  \footnotesize{$\pm$ 0.3}&\bf{71.3}  \footnotesize{$\pm$ 0.8}&\bf{51.3}  \footnotesize{$\pm$ 3.4}\\
\midrule
ResNet-18   &GeM + FC 2048                    &2048  &Pitts30k & 75.0  \footnotesize{$\pm$ 0.4}&29.9  \footnotesize{$\pm$ 0.6}&34.5  \footnotesize{$\pm$ 0.4}&36.1  \footnotesize{$\pm$ 0.2}&63.7  \footnotesize{$\pm$ 0.3}&45.1  \footnotesize{$\pm$ 2.1}\\
ResNet-18   &NetVLAD + PCA 2048               &2048  &Pitts30k & 85.0  \footnotesize{$\pm$ 0.4}&45.0  \footnotesize{$\pm$ 1.5}&56.6  \footnotesize{$\pm$ 0.7}&53.2  \footnotesize{$\pm$ 2.4}&75.4  \footnotesize{$\pm$ 1.1}&54.6  \footnotesize{$\pm$ 3.0}\\
ResNet-18	&CRN + PCA 2048                   &2048  &Pitts30k & \bf{85.7}  \footnotesize{$\pm$ 0.3}&\bf{50.6}  \footnotesize{$\pm$ 0.6}&\bf{61.0}  \footnotesize{$\pm$ 1.6}&\bf{62.8}  \footnotesize{$\pm$ 1.2}&\bf{77.4}  \footnotesize{$\pm$ 0.5}&\bf{61.1}  \footnotesize{$\pm$ 2.7}\\
\midrule
ResNet-18   &NetVLAD \cite{Arandjelovic-2018} &16384 &Pitts30k & 86.4  \footnotesize{$\pm$ 0.3}&47.4  \footnotesize{$\pm$ 1.2}&63.4  \footnotesize{$\pm$ 1.2}&61.4  \footnotesize{$\pm$ 1.5}&76.8  \footnotesize{$\pm$ 1.2}&57.6  \footnotesize{$\pm$ 3.3}\\
ResNet-18   &CRN \cite{Kim-2017}             &16384 &Pitts30k & \bf{86.8}  \footnotesize{$\pm$ 0.1}&\bf{53.2}  \footnotesize{$\pm$ 0.7}&\bf{68.8}  \footnotesize{$\pm$ 1.0}&\bf{69.0}  \footnotesize{$\pm$ 0.6}&\bf{79.1}  \footnotesize{$\pm$ 0.3}&\bf{64.8}  \footnotesize{$\pm$ 3.2}\\

\midrule

ResNet-50   &SPOC  \cite{Babenko-2015}       &1024  &Pitts30k & 60.9  \footnotesize{$\pm$ 0.5}&19.2  \footnotesize{$\pm$ 0.4}&14.0  \footnotesize{$\pm$ 0.5}& 9.0  \footnotesize{$\pm$ 0.7}&40.5  \footnotesize{$\pm$ 2.3}&27.1  \footnotesize{$\pm$ 1.5}\\
ResNet-50   &MAC \cite{Razavian-2015}        &1024  &Pitts30k & 77.6  \footnotesize{$\pm$ 0.2}&36.2  \footnotesize{$\pm$ 0.7}&36.2  \footnotesize{$\pm$ 1.4}&34.8  \footnotesize{$\pm$ 0.7}&72.9  \footnotesize{$\pm$ 0.3}&51.3  \footnotesize{$\pm$ 2.4}\\
ResNet-50   &RMAC \cite{Tolias-2016}         &1024  &Pitts30k & 74.9  \footnotesize{$\pm$ 1.0}&34.8  \footnotesize{$\pm$ 0.8}&41.8  \footnotesize{$\pm$ 0.6}&46.4  \footnotesize{$\pm$ 1.0}&73.1  \footnotesize{$\pm$ 0.7}&\bf{68.7}  \footnotesize{$\pm$ 0.5}\\
ResNet-50   &RRM \cite{Kordopatis-2021}           &1024  &Pitts30k & 72.8  \footnotesize{$\pm$ 0.2}&27.9  \footnotesize{$\pm$ 0.6}&28.3  \footnotesize{$\pm$ 0.8}&28.6  \footnotesize{$\pm$ 1.0}&65.9  \footnotesize{$\pm$ 0.9}&45.1  \footnotesize{$\pm$ 1.7}\\
ResNet-50   &GeM \cite{Radenovic-2019}       &1024  &Pitts30k & 82.0  \footnotesize{$\pm$ 0.3}&38.0  \footnotesize{$\pm$ 0.1}&41.5  \footnotesize{$\pm$ 1.8}&45.4  \footnotesize{$\pm$ 2.0}&66.3  \footnotesize{$\pm$ 2.5}&59.0  \footnotesize{$\pm$ 1.4}\\
ResNet-50   &NetVLAD + PCA 1024               &1024  &Pitts30k & 83.9  \footnotesize{$\pm$ 0.7}&46.5  \footnotesize{$\pm$ 2.0}&59.4  \footnotesize{$\pm$ 1.2}&53.2  \footnotesize{$\pm$ 3.8}&72.5  \footnotesize{$\pm$ 0.3}&57.7  \footnotesize{$\pm$ 2.0}\\
ResNet-50   &CRN + PCA 1024                   &1024  &Pitts30k & \bf{84.1}  \footnotesize{$\pm$ 0.4}&\bf{49.9}  \footnotesize{$\pm$ 0.8}&\bf{64.6}  \footnotesize{$\pm$ 1.2}&\bf{58.8}  \footnotesize{$\pm$ 0.1}&\bf{74.3}  \footnotesize{$\pm$ 0.2}&63.4  \footnotesize{$\pm$ 0.4}\\
\midrule
ResNet-50   &GeM + FC 2048                    &2048  &Pitts30k & 80.1  \footnotesize{$\pm$ 0.2}&33.7  \footnotesize{$\pm$ 0.3}&43.6  \footnotesize{$\pm$ 1.6}&48.2  \footnotesize{$\pm$ 1.2}&70.0  \footnotesize{$\pm$ 0.3}&56.0  \footnotesize{$\pm$ 1.7}\\
ResNet-50   &NetVLAD + PCA 2048               &2048  &Pitts30k & 84.4  \footnotesize{$\pm$ 0.4}&47.9  \footnotesize{$\pm$ 2.0}&62.6  \footnotesize{$\pm$ 1.7}&56.0  \footnotesize{$\pm$ 2.9}&74.1  \footnotesize{$\pm$ 0.4}&58.9  \footnotesize{$\pm$ 1.6}\\
ResNet-50   &CRN + PCA 2048                   &2048  &Pitts30k & \bf{84.7}  \footnotesize{$\pm$ 0.3}&\bf{51.2}  \footnotesize{$\pm$ 0.8}&\bf{67.1}  \footnotesize{$\pm$ 0.7}&\bf{62.3}  \footnotesize{$\pm$ 0.3}&\bf{75.8}  \footnotesize{$\pm$ 0.2}&\bf{65.0}  \footnotesize{$\pm$ 0.1}\\
\midrule
ResNet-50   &NetVLAD \cite{Arandjelovic-2018} &65536 &Pitts30k & \bf{86.0}  \footnotesize{$\pm$ 0.1}&50.7  \footnotesize{$\pm$ 2.0}&69.8  \footnotesize{$\pm$ 0.8}&67.1  \footnotesize{$\pm$ 2.3}&77.7  \footnotesize{$\pm$ 0.4}&60.2  \footnotesize{$\pm$ 1.6}\\
ResNet-50   &CRN \cite{Kim-2017}             &65536 &Pitts30k & 85.8  \footnotesize{$\pm$ 0.2}&\bf{54.0}  \footnotesize{$\pm$ 0.8}&\bf{73.1}  \footnotesize{$\pm$ 0.3}&\bf{70.9}  \footnotesize{$\pm$ 0.2}&\bf{79.7}  \footnotesize{$\pm$ 0.1}&\bf{65.9}  \footnotesize{$\pm$ 0.4}\\

\midrule\midrule

ResNet-18   &SPOC \cite{Babenko-2015}        &256   &MSLS     & 44.2  \footnotesize{$\pm$ 1.0}&39.5  \footnotesize{$\pm$ 0.5}&20.3  \footnotesize{$\pm$ 1.3}& 9.5  \footnotesize{$\pm$ 0.9}&62.3  \footnotesize{$\pm$ 0.6}&58.8  \footnotesize{$\pm$ 0.8}\\
ResNet-18   &MAC \cite{Razavian-2015}        &256   &MSLS     & 60.4  \footnotesize{$\pm$ 1.1}&54.7  \footnotesize{$\pm$ 1.8}&20.4  \footnotesize{$\pm$ 2.6}&18.9  \footnotesize{$\pm$ 2.0}&76.3  \footnotesize{$\pm$ 1.2}&69.2    \footnotesize{$\pm$ 1.2}\\
ResNet-18   &RMAC \cite{Tolias-2016}         &256   &MSLS     & 58.1  \footnotesize{$\pm$ 1.2}&48.9  \footnotesize{$\pm$ 2.0}&29.1  \footnotesize{$\pm$ 2.0}&34.3  \footnotesize{$\pm$ 1.4}&73.3  \footnotesize{$\pm$ 1.1}&63.7    \footnotesize{$\pm$ 2.7}\\
ResNet-18   &RRM   \cite{Kordopatis-2021}         &256   &MSLS     & 60.8  \footnotesize{$\pm$ 1.5}&54.9  \footnotesize{$\pm$ 2.6}&\bf{44.4}  \footnotesize{$\pm$ 2.1}&30.9  \footnotesize{$\pm$ 2.8}&75.7  \footnotesize{$\pm$ 1.5}&68.7    \footnotesize{$\pm$ 1.4}\\
ResNet-18   &GeM \cite{Radenovic-2019}       &256   &MSLS     & 71.6  \footnotesize{$\pm$ 0.1}&65.3  \footnotesize{$\pm$ 0.2}&42.8  \footnotesize{$\pm$ 1.1}&30.5  \footnotesize{$\pm$ 0.8}&80.3  \footnotesize{$\pm$ 0.1}&83.2  \footnotesize{$\pm$ 0.9}\\
ResNet-18   &GeM + FC 256                     &256   &MSLS     & 68.6  \footnotesize{$\pm$ 1.1}&59.6  \footnotesize{$\pm$ 2.6}&41.9  \footnotesize{$\pm$ 2.7}&31.3  \footnotesize{$\pm$ 0.5}&78.5  \footnotesize{$\pm$ 2.0}&76.1    \footnotesize{$\pm$ 3.4}\\
ResNet-18   &NetVLAD + PCA 256                &256   &MSLS     & 74.2  \footnotesize{$\pm$ 0.2}&70.6  \footnotesize{$\pm$ 0.3}&43.6  \footnotesize{$\pm$ 0.5}&34.7  \footnotesize{$\pm$ 1.7}&84.4  \footnotesize{$\pm$ 0.4}&89.8    \footnotesize{$\pm$ 0.5}\\
ResNet-18	&CRN + PCA 256                    &256   &MSLS     & \bf{74.5}  \footnotesize{$\pm$ 0.8}&\bf{72.1}  \footnotesize{$\pm$ 0.1}&44.1  \footnotesize{$\pm$ 1.4}&\bf{35.1}  \footnotesize{$\pm$ 2.4}&\bf{84.8}  \footnotesize{$\pm$ 0.3}&\bf{91.6}    \footnotesize{$\pm$ 0.4}\\
\midrule
ResNet-18   &GeM + FC 2048                    &2048  &MSLS     & 71.9  \footnotesize{$\pm$ 1.0}&64.0  \footnotesize{$\pm$ 1.2}&51.8  \footnotesize{$\pm$ 0.9}&37.6  \footnotesize{$\pm$ 1.3}&81.1  \footnotesize{$\pm$ 0.9}&79.2    \footnotesize{$\pm$ 0.9}\\
ResNet-18   &NetVLAD + PCA 2048               &2048  &MSLS     & \bf{80.4}  \footnotesize{$\pm$ 0.4}&74.6  \footnotesize{$\pm$ 0.2}&55.6  \footnotesize{$\pm$ 1.2}&47.4  \footnotesize{$\pm$ 1.1}&86.4  \footnotesize{$\pm$ 0.3}&92.2    \footnotesize{$\pm$ 0.3}\\
ResNet-18	&CRN + PCA 2048                   &2048  &MSLS     & 80.1  \footnotesize{$\pm$ 0.8}&\bf{75.8}  \footnotesize{$\pm$ 0.1}&\bf{57.2}  \footnotesize{$\pm$ 2.3}&\bf{47.8}  \footnotesize{$\pm$ 2.7}&\bf{86.8}  \footnotesize{$\pm$ 0.3}&\bf{93.2}    \footnotesize{$\pm$ 0.4}\\
\midrule
ResNet-18   &NetVLAD \cite{Arandjelovic-2018} &16384 &MSLS     & \bf{81.6}  \footnotesize{$\pm$ 0.5}&75.8  \footnotesize{$\pm$ 0.1}&62.3  \footnotesize{$\pm$ 1.6}&\bf{55.1}  \footnotesize{$\pm$ 0.9}&87.1  \footnotesize{$\pm$ 0.2}&92.1    \footnotesize{$\pm$ 0.7}\\
ResNet-18   &CRN \cite{Kim-2017}             &16384 &MSLS     & 81.3  \footnotesize{$\pm$ 0.7}&\bf{76.8}  \footnotesize{$\pm$ 0.0}&\bf{63.8}  \footnotesize{$\pm$ 1.4}&53.9  \footnotesize{$\pm$ 2.0}&\bf{87.5}  \footnotesize{$\pm$ 0.2}&\bf{93.7}    \footnotesize{$\pm$ 0.1}\\

\midrule

ResNet-50   &SPOC  \cite{Babenko-2015}       &1024  &MSLS     & 47.5  \footnotesize{$\pm$ 1.3}&47.9  \footnotesize{$\pm$ 1.5}&20.6  \footnotesize{$\pm$ 1.6}& 8.9  \footnotesize{$\pm$ 1.0}&68.3  \footnotesize{$\pm$ 0.5}&68.6    \footnotesize{$\pm$ 1.4}\\
ResNet-50   &MAC \cite{Razavian-2015}        &1024  &MSLS     & 76.0  \footnotesize{$\pm$ 0.2}&67.4  \footnotesize{$\pm$ 1.6}&45.3  \footnotesize{$\pm$ 1.0}&44.4  \footnotesize{$\pm$ 2.6}&84.6  \footnotesize{$\pm$ 0.4}&86.0   \footnotesize{$\pm$ 0.7}\\
ResNet-50   &RMAC \cite{Tolias-2016}         &1024  &MSLS     & 70.1  \footnotesize{$\pm$ 0.8}&62.0  \footnotesize{$\pm$ 0.5}&52.1  \footnotesize{$\pm$ 2.3}&\bf{54.3}  \footnotesize{$\pm$ 1.8}&80.6  \footnotesize{$\pm$ 0.5}&85.9   \footnotesize{$\pm$ 1.0}\\
ResNet-50   &RRM \cite{Kordopatis-2021}           &1024  &MSLS     & 69.3  \footnotesize{$\pm$ 1.0}&67.4  \footnotesize{$\pm$ 0.4}&53.7  \footnotesize{$\pm$ 0.8}&43.7  \footnotesize{$\pm$ 1.0}&84.3  \footnotesize{$\pm$ 0.5}&84.8   \footnotesize{$\pm$ 1.1}\\
ResNet-50   &GeM \cite{Radenovic-2019}       &1024  &MSLS     & \bf{77.4}  \footnotesize{$\pm$ 0.6}&72.0  \footnotesize{$\pm$ 0.5}&\bf{55.4}  \footnotesize{$\pm$ 2.5}&45.7  \footnotesize{$\pm$ 1.0}&83.9  \footnotesize{$\pm$ 0.6}&91.2   \footnotesize{$\pm$ 0.7}\\
ResNet-50   &NetVLAD + PCA 1024               &1024  &MSLS     & \bf{77.4}  \footnotesize{$\pm$ 0.2}&74.8  \footnotesize{$\pm$ 0.3}&51.3  \footnotesize{$\pm$ 1.3}&39.0  \footnotesize{$\pm$ 1.3}&85.2  \footnotesize{$\pm$ 0.3}&92.9    \footnotesize{$\pm$ 0.3}\\
ResNet-50   &CRN + PCA 1024                   &1024  &MSLS     & 77.3  \footnotesize{$\pm$ 0.3}&\bf{75.6}  \footnotesize{$\pm$ 0.0}&51.8  \footnotesize{$\pm$ 1.1}&38.8  \footnotesize{$\pm$ 1.0}&\bf{85.7}  \footnotesize{$\pm$ 0.3}&\bf{94.1}    \footnotesize{$\pm$ 0.2}\\
\midrule
ResNet-50   &GeM + FC 2048                    &2048  &MSLS     & \bf{79.2}  \footnotesize{$\pm$ 0.6}&73.5  \footnotesize{$\pm$ 0.8}&\bf{64.0}  \footnotesize{$\pm$ 3.9}&\bf{55.1}  \footnotesize{$\pm$ 2.4}&86.1  \footnotesize{$\pm$ 0.7}&90.3    \footnotesize{$\pm$ 1.0}\\
ResNet-50   &NetVLAD + PCA 2048               &2048  &MSLS     & 78.5  \footnotesize{$\pm$ 0.2}&75.4  \footnotesize{$\pm$ 0.2}&52.8  \footnotesize{$\pm$ 0.4}&42.6  \footnotesize{$\pm$ 1.3}&85.8  \footnotesize{$\pm$ 0.3}&93.4    \footnotesize{$\pm$ 0.4}\\
ResNet-50   &CRN + PCA 2048                   &2048  &MSLS     & 78.3  \footnotesize{$\pm$ 0.3}&\bf{76.3}  \footnotesize{$\pm$ 0.1}&54.3  \footnotesize{$\pm$ 0.7}&42.8  \footnotesize{$\pm$ 1.6}&\bf{86.2}  \footnotesize{$\pm$ 0.4}&\bf{94.4}   \footnotesize{$\pm$ 0.2}\\
\midrule
ResNet-50   &NetVLAD \cite{Arandjelovic-2018} &65536 &MSLS     & \bf{80.9}  \footnotesize{$\pm$ 0.0}&76.9  \footnotesize{$\pm$ 0.2}&62.8  \footnotesize{$\pm$ 0.9}&51.5  \footnotesize{$\pm$ 1.2}&87.2  \footnotesize{$\pm$ 0.3}&93.8   \footnotesize{$\pm$ 0.2}\\
ResNet-50   &CRN \cite{Kim-2017}             &65536 &MSLS     & 80.8  \footnotesize{$\pm$ 0.2}&\bf{77.8}  \footnotesize{$\pm$ 0.1}&\bf{63.6}  \footnotesize{$\pm$ 0.5}&\bf{53.4}  \footnotesize{$\pm$ 1.4}&\bf{87.5}  \footnotesize{$\pm$ 0.4}&\bf{94.8}  \footnotesize{$\pm$ 0.3}\\
	\bottomrule
  \end{tabular}}
    \caption{\textbf{Aggregation methods.} Full table of aggregation methods, grouped by backbone and features dimension.}
  \label{tab:t2_aggregation_methods_complete}
\end{table*}
%%%%%%%%%%%%%%%%%%%%%%%%%%%%%%%% AGGREGATIONS
%%%%%%%%%%%%%%%%%%%%%%%%%%%%%%%% AGGREGATIONS
%%%%%%%%%%%%%%%%%%%%%%%%%%%%%%%% AGGREGATIONS

\subsection{Aggregation and Descriptors Dimensionality}
%: full table}
\label{sec:supp_aggregation_methods}
In Tab. \ref{tab:t2_aggregation_methods_complete}, we show a more comprehensive set of results than in the main paper, comprising all the aggregation methods that can be attached to the different backbones using our software. As seen in the literature, GeM pooling \cite{Radenovic-2019}  outperforms in general   SPOC~\cite{Babenko-2015}, MAC~\cite{Razavian-2015}, R-MAC~\cite{Tolias-2016}, RRM~\cite{Kordopatis-2021}.

%%%%%%%%%%%%%%%%%%%%%%%%%%%%%%%% TRANSFORMERS
%%%%%%%%%%%%%%%%%%%%%%%%%%%%%%%% TRANSFORMERS
%%%%%%%%%%%%%%%%%%%%%%%%%%%%%%%% TRANSFORMERS
\begin{table*}[tb!]
  \centering
  \resizebox{\textwidth}{!}{
  \begin{tabular}{llllllllllll}
    \toprule
    \multirow{2}{*}{Backbone} & 
    \begin{tabular}[c]{@{}l@{}}Aggregation\\ Method\end{tabular} & \begin{tabular}[c]{@{}l@{}}Features\\ Dim\end{tabular} & 
    \multirow{2}{*}{FLOPs [GF]} &
    \begin{tabular}[c]{@{}l@{}}Model\\ Size [MB]\end{tabular} &
    \begin{tabular}[c]{@{}l@{}}Training \\ Dataset\end{tabular} & \begin{tabular}[c]{@{}l@{}}R@1\\ Pitts30k\end{tabular} & \begin{tabular}[c]{@{}l@{}}R@1\\ MSLS\end{tabular} & \begin{tabular}[c]{@{}l@{}}R@1 \\ Tokyo 24/7\end{tabular} & \begin{tabular}[c]{@{}l@{}}R@1\\ R-SF\end{tabular} & \begin{tabular}[c]{@{}l@{}}R@1\\ Eynsham\end{tabular} & \begin{tabular}[c]{@{}l@{}}R@1\\ St Lucia\end{tabular}  \\
    \midrule
    ResNet-18 & GeM   & 256 & 17.29  &10.63& Pitts30k&  77.8  \footnotesize{$\pm$ 0.2}&35.3  \footnotesize{$\pm$ 0.5}&35.3  \footnotesize{$\pm$ 1.1}&34.2  \footnotesize{$\pm$ 1.7}&64.3  \footnotesize{$\pm$ 1.2}&46.2  \footnotesize{$\pm$ 0.4}\\
    ResNet-50       &GeM      &1024  &40.61 & 32.71 &Pitts30k   & \bf{82.0}  \footnotesize{$\pm$ 0.3}&38.0  \footnotesize{$\pm$ 0.1}&41.5  \footnotesize{$\pm$ 1.8}&45.4  \footnotesize{$\pm$ 2.0}&66.3  \footnotesize{$\pm$ 2.5}&59.0  \footnotesize{$\pm$ 1.4}\\
    ViT     & CLS     & 768  & 82.31 & 350.96& Pitts30k&  79.2 \footnotesize{$\pm$ 1.5} &39.0 \footnotesize{$\pm$ 0.8} &44.5 \footnotesize{$\pm$ 3.2} &48.3 \footnotesize{$\pm$ 2.5} &67.6 \footnotesize{$\pm$ 1.2} &\bf{69.6} \footnotesize{$\pm$ 2.0} \\
    CCT     & CLS     & 384  & 22.34 & 190.39 & Pitts30k& 76.3 \footnotesize{$\pm$ 1.4} &39.5 \footnotesize{$\pm$ 0.4} &39.0 \footnotesize{$\pm$ 1.7} &44.4 \footnotesize{$\pm$ 0.4} &50.8 \footnotesize{$\pm$ 2.1} &57.3 \footnotesize{$\pm$ 2.6}  \\
    CCT     & SeqPool & 384  & 26.19  & 221.92 & Pitts30k& 81.1 \footnotesize{$\pm$ 1.0} &46.9 \footnotesize{$\pm$ 1.2} &51.5 \footnotesize{$\pm$ 0.8} &57.8 \footnotesize{$\pm$ 1.5} & \bf{75.2} \footnotesize{$\pm$ 1.1} &63.6 \footnotesize{$\pm$ 2.6}  \\
    CCT     & GeM     & 384  & 22.36 & 191.24 & Pitts30k&  79.6 \footnotesize{$\pm$ 0.3} & \bf{47.8} \footnotesize{$\pm$ 0.7} & \bf{52.3} \footnotesize{$\pm$ 2.0} & \bf{61.3} \footnotesize{$\pm$ 0.1} &71.0 \footnotesize{$\pm$ 0.8} &59.1 \footnotesize{$\pm$ 2.0} \\
    \midrule
    ResNet-18 & NetVLAD & 16384 & 17.27 &10.76& Pitts30k &   \bf{86.4}  \footnotesize{$\pm$ 0.3}&47.4 \footnotesize{$\pm$ 1.2}&63.4  \footnotesize{$\pm$ 1.2}&61.4  \footnotesize{$\pm$ 1.5}&76.8 \footnotesize{$\pm$ 1.2}&57.6  \footnotesize{$\pm$ 3.3}\\
    ResNet-50       &NetVLAD  &65536 &40.51 & 33.21 & Pitts30k  & 86.0  \footnotesize{$\pm$ 0.1}&50.7  \footnotesize{$\pm$ 2.0}&\bf{69.8}  \footnotesize{$\pm$ 0.8}&67.1  \footnotesize{$\pm$ 2.3}&\bf{77.7}  \footnotesize{$\pm$ 0.4}&\bf{60.2}  \footnotesize{$\pm$ 1.6}\\
    CCT       & NetVLAD & 24576 & 18.53 & 160.08 & Pitts30k&84.6 \footnotesize{$\pm$ 0.3} &\bf{52.5} \footnotesize{$\pm$ 1.9} &69.1 \footnotesize{$\pm$ 0.4} &\bf{73.5} \footnotesize{$\pm$ 1.4} &72.6 \footnotesize{$\pm$ 0.6} &56.1 \footnotesize{$\pm$ 3.3} \\
    \midrule
    \midrule
    ResNet-18 & GeM   & 256 & 17.29  &10.63& MSLS&  71.6  \footnotesize{$\pm$ 0.1}&65.3  \footnotesize{$\pm$ 0.2}&42.8  \footnotesize{$\pm$ 1.1}&30.5  \footnotesize{$\pm$ 0.8}&80.3  \footnotesize{$\pm$ 0.1}&83.2   \footnotesize{$\pm$ 0.9}\\
    ResNet-50       &GeM      &1024  &40.61 & 32.71 &MSLS   & 77.4  \footnotesize{$\pm$ 0.6}&72.0  \footnotesize{$\pm$ 0.5}&55.4  \footnotesize{$\pm$ 2.5}&45.7  \footnotesize{$\pm$ 1.0}&83.9  \footnotesize{$\pm$ 0.6}&91.2   \footnotesize{$\pm$ 0.7}\\
    ViT     & CLS     & 768  & 82.31 & 350.96& MSLS&  \bf{82.9} \footnotesize{$\pm$ 0.6} &\bf{73.5} \footnotesize{$\pm$ 0.6} &\bf{59.9} \footnotesize{$\pm$ 4.4} &\bf{65.0} \footnotesize{$\pm$ 1.1} &84.5 \footnotesize{$\pm$ 1.0} &93.6 \footnotesize{$\pm$ 0.7} \\
    CCT     & CLS     & 384  & 22.34 & 190.39 & MSLS& 79.6 \footnotesize{$\pm$ 0.3} &71.1 \footnotesize{$\pm$ 0.4} &52.0 \footnotesize{$\pm$ 1.1} &49.9 \footnotesize{$\pm$ 1.8} &85.6 \footnotesize{$\pm$ 0.1} & \bf{94.0} \footnotesize{$\pm$ 0.3} \\
    CCT     & SeqPool & 384  & 26.19  & 221.92 & MSLS& 
    81.4 \footnotesize{$\pm$ 0.8} &71.0 \footnotesize{$\pm$ 0.9} &59.1 \footnotesize{$\pm$ 3.2} &60.5 \footnotesize{$\pm$ 1.5} & \bf{86.1} \footnotesize{$\pm$ 0.6} &92.4 \footnotesize{$\pm$ 1.1}\\
    CCT     & GeM     & 384  & 22.36 & 191.24 & MSLS&  78.7 \footnotesize{$\pm$ 0.6} &72.0 \footnotesize{$\pm$ 0.6} &48.8 \footnotesize{$\pm$ 1.2} &48.6 \footnotesize{$\pm$ 2.9} &83.9 \footnotesize{$\pm$ 0.1} &92.9 \footnotesize{$\pm$ 0.7} \\
    \midrule
    ResNet-18 & NetVLAD & 16384 & 17.27 &10.76& MSLS & 81.6 \footnotesize{$\pm$ 0.5}&75.8  \footnotesize{$\pm$ 0.1}&62.3  \footnotesize{$\pm$ 1.6}&55.1  \footnotesize{$\pm$ 0.9}&87.1  \footnotesize{$\pm$ 0.2}&92.1   \footnotesize{$\pm$ 0.7}\\
    ResNet-50       &NetVLAD  &65536 &40.51 & 33.21 & MSLS  & 80.9  \footnotesize{$\pm$ 0.0}&76.9 \footnotesize{$\pm$ 0.2}&62.8 \footnotesize{$\pm$ 0.9}&51.5  \footnotesize{$\pm$ 1.2}&87.2 \footnotesize{$\pm$ 0.3}&93.8   \footnotesize{$\pm$ 0.2}\\
    CCT       & NetVLAD & 24576 & 18.53 & 160.08 & MSLS&  \bf{85.1} \footnotesize{$\pm$ 0.2} & \bf{79.9} \footnotesize{$\pm$ 0.3} &\bf{70.3} \footnotesize{$\pm$ 2.0} &\bf{65.9} \footnotesize{$\pm$ 1.3} &\bf{87.4} \footnotesize{$\pm$ 0.2} &\bf{98.4} \footnotesize{$\pm$ 0.2} \\
    \bottomrule
  \end{tabular}
  }
  \vspace{-0.1cm}
  \caption{\textbf{Transformers} Comparison of traditional CNN architectures with novel Transformers-based approaches. 
  }
  \label{tab:supp_transformers}
\vspace{-0.2cm}
\end{table*}

\subsection{Visual Transformers: full table}
Tab. \ref{tab:supp_transformers} includes results using Transformer-based backbones when trained on Pitts30k that could not fit into the main paper.
In general,  it can be seen that these architectures confer better generalization capabilities, outperforming both a ResNet-18 and a much more costly ResNet-50. Additionally,  directly using the CLS token yields worse results than SeqPool, GeM, or NetVLAD. 
A possible explanation is that using the CLS is the only strategy that does not consider the whole set of tokens. This consideration could indicate that the CLS token provides a less robust representation, especially when trained on small-scale datasets.

%%%%%%%%%%%%%%%%%%%%%%%%%%%%%%%% MINING
%%%%%%%%%%%%%%%%%%%%%%%%%%%%%%%% MINING
%%%%%%%%%%%%%%%%%%%%%%%%%%%%%%%% MINING

\subsection{Negative Mining}
\label{sec:supp_mining}
Since the inception of the triplet loss, considerable attention has been paid to finding the best possible negative images.
Using negatives too different from the query will cause a drop in the loss to low values (even to zero if using a triplet margin loss), severely hindering the learning process of the model.  
For this reason, mining for the hardest negatives w.r.t. a given query is an important step in learning representation in general and, hence, in Visual Geo-localization. Therefore several hard negative mining techniques were proposed in the literature.
In \cite{Arandjelovic-2018} the authors propose to compute offline features for all images (cache) periodically and to use such features to find the most difficult negatives. We refer to this as "full database mining".
While this has proven to produce good results, its time and space complexities grow linearly with the database size, making it extremely costly to use it with large dimensional descriptors and large-scale datasets.
In \cite{Warburg-2020} the authors presented a new large scale dataset, for which the mining proposed by \cite{Arandjelovic-2018} would be rather time-consuming, and they performed an approximation of it considering only a small subset of the database (1000 images), making it a more suitable choice for large scale datasets. We refer to this as "partial database mining".

\begin{table*}[htb]
  \centering
  \resizebox{\textwidth}{!}{
  \begin{tabular}{llllllllll}
    \toprule
    \multirow{2}{*}{Backbone} & 
    \begin{tabular}[c]{@{}l@{}}Aggregation\\ Method\end{tabular} & \begin{tabular}[c]{@{}l@{}}Mining\\ Method\end{tabular} & 
    \begin{tabular}[c]{@{}l@{}}Training \\ Dataset\end{tabular} & \begin{tabular}[c]{@{}l@{}}R@1\\ Pitts30k\end{tabular} & \begin{tabular}[c]{@{}l@{}}R@1\\ MSLS\end{tabular} & \begin{tabular}[c]{@{}l@{}}R@1 \\ Tokyo 24/7\end{tabular} & \begin{tabular}[c]{@{}l@{}}R@1\\ R-SF\end{tabular} & \begin{tabular}[c]{@{}l@{}}R@1\\ Eynsham\end{tabular} & \begin{tabular}[c]{@{}l@{}}R@1\\ St Lucia\end{tabular}  \\
    \midrule
ResNet-18    &GeM      &Random   					&Pitts30k & 73.7  \footnotesize{$\pm$ 0.7}&30.5  \footnotesize{$\pm$ 0.5}&31.3  \footnotesize{$\pm$ 0.8}&24.0  \footnotesize{$\pm$ 1.2}&58.2  \footnotesize{$\pm$ 1.4}&41.0  \footnotesize{$\pm$ 1.2}\\
ResNet-18    &GeM      &Full database mining  		&Pitts30k & \bf{77.8}  \footnotesize{$\pm$ 0.2}&\bf{35.3}  \footnotesize{$\pm$ 0.5}&\bf{35.3}  \footnotesize{$\pm$ 1.1}&\bf{34.2}  \footnotesize{$\pm$ 1.7}&\bf{64.3}  \footnotesize{$\pm$ 1.2}&\bf{46.2}  \footnotesize{$\pm$ 0.4}\\
ResNet-18    &GeM      &Partial database mining     &Pitts30k & 76.5  \footnotesize{$\pm$ 0.3}&34.2  \footnotesize{$\pm$ 1.3}&33.9  \footnotesize{$\pm$ 1.4}&32.9  \footnotesize{$\pm$ 0.7}&64.0  \footnotesize{$\pm$ 2.4}&45.6  \footnotesize{$\pm$ 0.9}\\
\midrule
ResNet-18    &NetVLAD  &Random                      &Pitts30k & 83.9  \footnotesize{$\pm$ 0.5}&43.6  \footnotesize{$\pm$ 0.5}&55.1  \footnotesize{$\pm$ 1.3}&53.8  \footnotesize{$\pm$ 1.1}&76.3  \footnotesize{$\pm$ 0.6}&53.5  \footnotesize{$\pm$ 1.4}\\
ResNet-18    &NetVLAD  &Full database mining        &Pitts30k & \bf{86.4}  \footnotesize{$\pm$ 0.3}&\bf{47.4}  \footnotesize{$\pm$ 1.2}&\bf{63.4}  \footnotesize{$\pm$ 1.2}&61.4  \footnotesize{$\pm$ 1.5}&\bf{76.8}  \footnotesize{$\pm$ 1.2}&\bf{57.6}  \footnotesize{$\pm$ 3.3}\\
ResNet-18    &NetVLAD  &Partial database mining     &Pitts30k & 86.2  \footnotesize{$\pm$ 0.3}&47.3  \footnotesize{$\pm$ 0.4}&61.2  \footnotesize{$\pm$ 0.5}&\bf{62.9}  \footnotesize{$\pm$ 0.3}&76.6  \footnotesize{$\pm$ 0.5}&57.1  \footnotesize{$\pm$ 1.6}\\
\midrule
ResNet-50    &GeM      &Random   					&Pitts30k & 77.9  \footnotesize{$\pm$ 1.0}&34.3  \footnotesize{$\pm$ 1.3}&40.1  \footnotesize{$\pm$ 1.0}&35.5  \footnotesize{$\pm$ 3.0}&63.8  \footnotesize{$\pm$ 0.9}&52.3  \footnotesize{$\pm$ 1.4}\\
ResNet-50    &GeM      &Full database mining  		&Pitts30k & 82.0  \footnotesize{$\pm$ 0.3}&38.0  \footnotesize{$\pm$ 0.1}&41.5  \footnotesize{$\pm$ 1.8}&45.4  \footnotesize{$\pm$ 2.0}&66.3  \footnotesize{$\pm$ 2.5}&59.0  \footnotesize{$\pm$ 1.4}\\
ResNet-50    &GeM      &Partial database mining     &Pitts30k & \bf{82.3}  \footnotesize{$\pm$ 0.0}&\bf{39.0}  \footnotesize{$\pm$ 0.4}&\bf{43.5}  \footnotesize{$\pm$ 0.2}&\bf{45.5}  \footnotesize{$\pm$ 1.7}&\bf{67.7}  \footnotesize{$\pm$ 1.4}&\bf{61.0}  \footnotesize{$\pm$ 2.0}\\
\midrule
ResNet-50    &NetVLAD  &Random                      &Pitts30k & 83.4  \footnotesize{$\pm$ 0.6}&45.0  \footnotesize{$\pm$ 0.3}&61.9  \footnotesize{$\pm$ 2.1}&55.8  \footnotesize{$\pm$ 1.5}&75.0  \footnotesize{$\pm$ 1.8}&52.6  \footnotesize{$\pm$ 1.2}\\
ResNet-50    &NetVLAD  &Full database mining        &Pitts30k & \bf{86.0}  \footnotesize{$\pm$ 0.1}&\bf{50.7}  \footnotesize{$\pm$ 2.0}&\bf{69.8}  \footnotesize{$\pm$ 0.8}&\bf{67.1}  \footnotesize{$\pm$ 2.3}&\bf{77.7}  \footnotesize{$\pm$ 0.4}&\bf{60.2}  \footnotesize{$\pm$ 1.6}\\
ResNet-50    &NetVLAD  &Partial database mining     &Pitts30k & 85.5  \footnotesize{$\pm$ 0.3}&48.6  \footnotesize{$\pm$ 3.1}&66.7  \footnotesize{$\pm$ 4.1}&65.0  \footnotesize{$\pm$ 4.3}&77.6  \footnotesize{$\pm$ 1.3}&59.0  \footnotesize{$\pm$ 4.1}\\
\midrule
\midrule
ResNet-18    &GeM      &Random   					&MSLS     & 62.2  \footnotesize{$\pm$ 0.3}&50.6  \footnotesize{$\pm$ 0.6}&28.8  \footnotesize{$\pm$ 0.8}&17.1  \footnotesize{$\pm$ 1.0}&70.2  \footnotesize{$\pm$ 0.6}&71.4   \footnotesize{$\pm$ 1.0}\\
ResNet-18    &GeM      &Full database mining  		&MSLS     & 70.1  \footnotesize{$\pm$ 1.1}&61.8  \footnotesize{$\pm$ 0.5}&\bf{42.8}  \footnotesize{$\pm$ 1.4}&\bf{31.3}  \footnotesize{$\pm$ 1.2}&79.3  \footnotesize{$\pm$ 0.2}&81.0   \footnotesize{$\pm$ 0.9}\\
ResNet-18    &GeM      &Partial database mining     &MSLS     & \bf{71.6}  \footnotesize{$\pm$ 0.1}&\bf{65.3}  \footnotesize{$\pm$ 0.2}&\bf{42.8}  \footnotesize{$\pm$ 1.1}&30.5  \footnotesize{$\pm$ 0.8}&\bf{80.3}  \footnotesize{$\pm$ 0.1}&\bf{83.2}   \footnotesize{$\pm$ 0.9}\\
\midrule
ResNet-18    &NetVLAD  &Random                      &MSLS     & 73.3  \footnotesize{$\pm$ 0.7}&61.5  \footnotesize{$\pm$ 1.4}&45.0  \footnotesize{$\pm$ 1.5}&34.8  \footnotesize{$\pm$ 0.2}&84.9  \footnotesize{$\pm$ 0.3}&79.7   \footnotesize{$\pm$ 1.7}\\
ResNet-18    &NetVLAD  &Full database mining        &MSLS     & -&- &- &- &- &- \\
ResNet-18    &NetVLAD  &Partial database mining     &MSLS     & \bf{81.6}  \footnotesize{$\pm$ 0.5}&\bf{75.8}  \footnotesize{$\pm$ 0.1}&\bf{62.3}  \footnotesize{$\pm$ 1.6}&\bf{55.1}  \footnotesize{$\pm$ 0.9}&\bf{87.1}  \footnotesize{$\pm$ 0.2}&\bf{92.1}   \footnotesize{$\pm$ 0.7}\\
\midrule
ResNet-50    &GeM      &Random   					&MSLS     & 69.5  \footnotesize{$\pm$ 1.2}&57.4  \footnotesize{$\pm$ 1.1}&43.5  \footnotesize{$\pm$ 3.3}&31.1  \footnotesize{$\pm$ 0.9}&78.8  \footnotesize{$\pm$ 0.5}&78.3   \footnotesize{$\pm$ 1.2}\\
ResNet-50    &GeM      &Full database mining  		&MSLS     & 77.3  \footnotesize{$\pm$ 0.3}&69.7  \footnotesize{$\pm$ 0.2}&52.4  \footnotesize{$\pm$ 1.7}&45.3  \footnotesize{$\pm$ 0.2}&\bf{84.2}  \footnotesize{$\pm$ 0.0}&91.0   \footnotesize{$\pm$ 0.2}\\
ResNet-50    &GeM      &Partial database mining     &MSLS     & \bf{77.4}  \footnotesize{$\pm$ 0.6}&\bf{72.0}  \footnotesize{$\pm$ 0.5}&\bf{55.4}  \footnotesize{$\pm$ 2.5}&\bf{45.7}  \footnotesize{$\pm$ 1.0}&83.9  \footnotesize{$\pm$ 0.6}&\bf{91.2}   \footnotesize{$\pm$ 0.7}\\
\midrule
ResNet-50    &NetVLAD  &Random                      &MSLS     & 74.9  \footnotesize{$\pm$ 0.4}&63.6  \footnotesize{$\pm$ 1.3}&41.9  \footnotesize{$\pm$ 1.6}&34.6  \footnotesize{$\pm$ 2.3}&85.5  \footnotesize{$\pm$ 0.2}&80.9   \footnotesize{$\pm$ 0.4}\\
ResNet-50    &NetVLAD  &Full database mining        &MSLS     & -&- &- &- &- &- \\
ResNet-50    &NetVLAD  &Partial database mining     &MSLS     & \bf{80.9}  \footnotesize{$\pm$ 0.0}&\bf{76.9}  \footnotesize{$\pm$ 0.2}&\bf{62.8}  \footnotesize{$\pm$ 0.9}&\bf{51.5}  \footnotesize{$\pm$ 1.2}&\bf{87.2}  \footnotesize{$\pm$ 0.3}&\bf{93.8}   \footnotesize{$\pm$ 0.2}\\
    \bottomrule
  \end{tabular}}
    \caption{\textbf{Mining methods.} }
  \label{tab:supp_t3_mining}
\end{table*}

\myparagraph{Discussion}
Tab. \ref{tab:supp_t3_mining} shows results when training with different mining methods.
Full database mining performs the best when training on Pitts30k, although the less expensive partial mining performs similarly.
Surprisingly, when training on Pitts30k, choosing random negatives without performing any mining operation results in only a 5\% drop in recall@1 (on average over all datasets) compared to the training with partial database mining, although the gap grows to 12\% when training on the MSLS dataset.
This is probably due to the huge variety in domains of MSLS (as it spans over multiple continents), making a random negative likely to be very different from the query.
On the other hand, Pitts30k is collected in a small area of Pittsburgh, with little to no weather variations, making random negatives a suitable choice for the triplet loss.

Training on MSLS, the results favour partial mining over full database mining because of the large scale of this dataset.
While all other experiments converge in less than 24 hours, training a network on the MSLS dataset using full database mining is computationally very expensive, and therefore we stopped training after 5 days. 
Moreover, training a model that outputs a descriptor with high dimensionality (such as the NetVLAD layer) converges slowly and also requires intractable amounts of RAM, as it requires all images' descriptors to be periodically computed and stored in RAM.
These results show that full database mining is impractical when working on large-scale problems.

%%%%%%%%%%%%%%%%%%%%%%%%%%%%%%%% DATA AUGMENTATION
%%%%%%%%%%%%%%%%%%%%%%%%%%%%%%%% DATA AUGMENTATION
%%%%%%%%%%%%%%%%%%%%%%%%%%%%%%%% DATA AUGMENTATION
\begin{figure*}[!ht]
    \centering
    \begin{minipage}{.32\textwidth}
        \begin{subfigure}{\textwidth}
            \includegraphics[width=\textwidth]{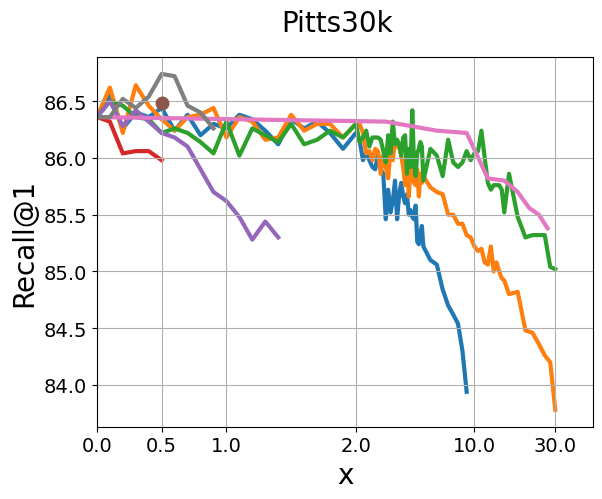}
        \end{subfigure}
        \begin{subfigure}{\textwidth}
            \includegraphics[width=\textwidth]{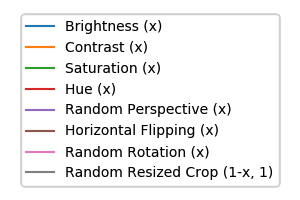} 
        \end{subfigure}
        \vspace*{33px}
    \end{minipage}
    \begin{minipage}{.32\textwidth}
        \begin{subfigure}{\textwidth}
            \includegraphics[width=\textwidth]{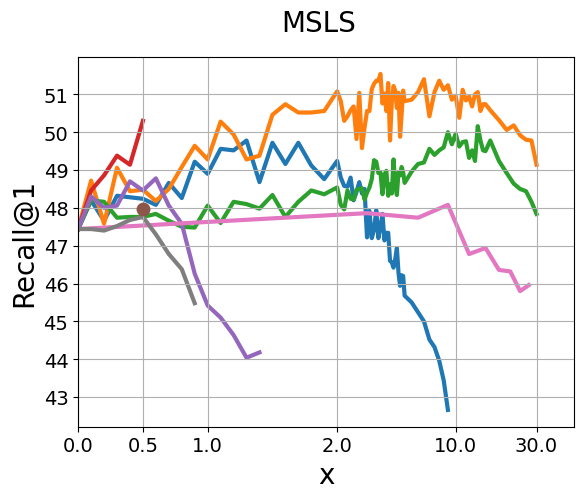} 
        \end{subfigure}
        \begin{subfigure}{\textwidth}
        \includegraphics[width=\textwidth]{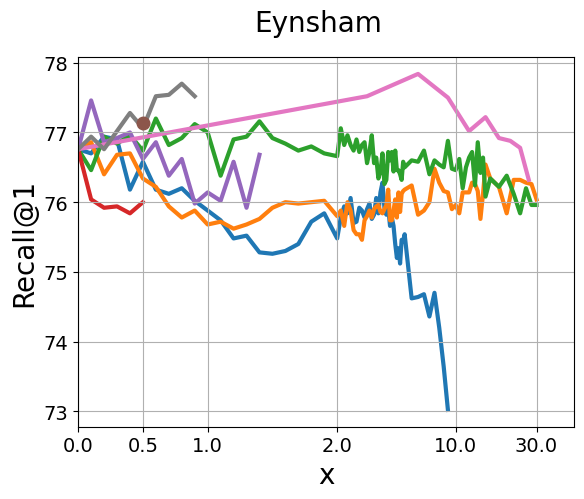} 
        \end{subfigure}
    \end{minipage}
    \begin{minipage}{.32\textwidth}
        \begin{subfigure}{\textwidth}
            \includegraphics[width=\textwidth]{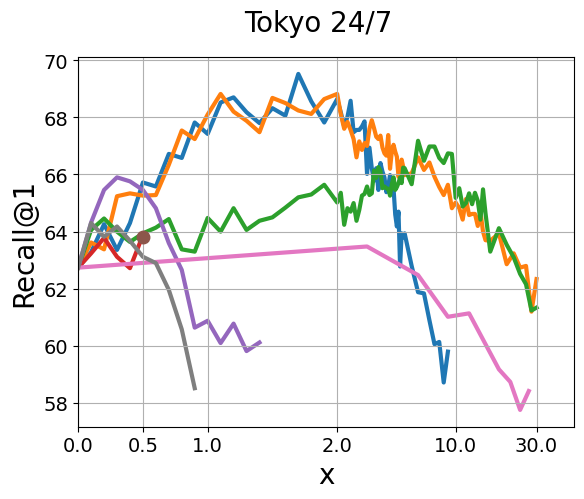} 
        \end{subfigure}
        \begin{subfigure}{\textwidth}
            \includegraphics[width=\textwidth]{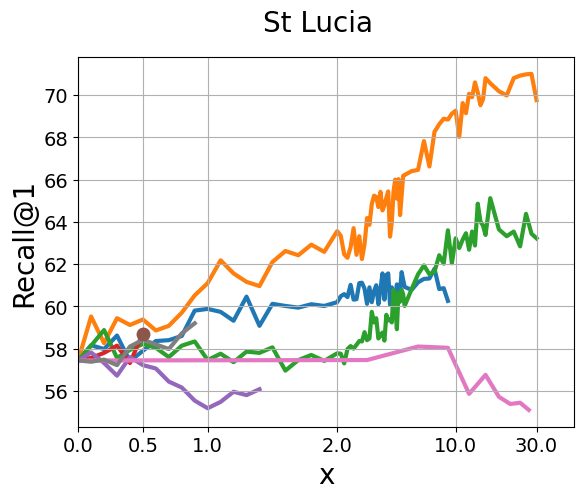} 
        \end{subfigure}
    \end{minipage}
    \caption{\textbf{Data Augmentation.} 
    Results obtained when applying  a number of popular data augmentation techniques during the training.
    We used PyTorch's transforms classes, and the x-axis relates to the parameter passed to the class. Brightness, contrast, saturation and hue are all performed with \texttt{ColorJittering()}. For \texttt{RandomPerspective()} and \texttt{RandomRotation()}, the parameter refers to the first argument (\texttt{distortion\_scale} and \texttt{degrees} respectively). Regarding \texttt{RandomResizedCrop()}, we use the value as $(1-x, 1)$ for \texttt{scale} so that all transformations have their origin in the same point (\emph{i.e.} $x=0$ equals to the identity transformation), and the crops are then resized to the original resolution. When used, \texttt{HorizontalFlipping()} is applied with a probability of 0.5. Please refer to the PyTorch documentation for further information.}
    \label{fig:supp_data_augmentation}
\end{figure*}
% add san francisco results (will be ready around the 20th)
\subsection{Data Augmentation}
\label{sec:supp_data_augmentation}
In Fig. \ref{fig:supp_data_augmentation} we report the same plots shown in the main paper at Fig. 2, at a bigger and more readable size.

%%%%%%%%%%%%%%%%%%%%%%%%%%%%%%%% PRE/POST PROCESSING
%%%%%%%%%%%%%%%%%%%%%%%%%%%%%%%% PRE/POST PROCESSING
%%%%%%%%%%%%%%%%%%%%%%%%%%%%%%%% PRE/POST PROCESSING

\subsection{Query pre/post-processing and Predictions Refinement}
\label{sec:supp_pre_post_processing}
\begin{table*}[!ht]
  \centering
  \resizebox{\textwidth}{!}{
  \begin{tabular}{lllllllllllll}
    \toprule
    \multirow{2}{*}{Backbone} &
    \begin{tabular}[c]{@{}l@{}}Aggregation\\ Method\end{tabular} & \begin{tabular}[c]{@{}l@{}}Pre/Post-\\ Processing\\Method\end{tabular} & \begin{tabular}[c]{@{}l@{}}Pre-\\Proc.\end{tabular} & \begin{tabular}[c]{@{}l@{}}Post-\\Proc.\end{tabular} & \begin{tabular}[c]{@{}l@{}}Batch\\ Parall.\end{tabular} &
    \begin{tabular}[c]{@{}l@{}}Training\\ Dataset.\end{tabular} &
    \begin{tabular}[c]{@{}l@{}}R@1\\ Pitts30k\end{tabular} & \begin{tabular}[c]{@{}l@{}}R@1\\ MSLS\end{tabular} & \begin{tabular}[c]{@{}l@{}}R@1 \\ Tokyo 24/7\end{tabular} & \begin{tabular}[c]{@{}l@{}}R@1\\ R-SF\end{tabular} & \begin{tabular}[c]{@{}l@{}}R@1\\ Eynsham\end{tabular} & \begin{tabular}[c]{@{}l@{}}R@1\\ St Lucia\end{tabular} \\
    \midrule
    ResNet-18 &GeM     &Hard Resize     &Y &N &Y &Pitts30k & \bf{77.8}  \footnotesize{$\pm$ 0.2}&35.3  \footnotesize{$\pm$ 0.5}&31.8  \footnotesize{$\pm$ 0.9}&33.2  \footnotesize{$\pm$ 2.1}&\bf{64.3}  \footnotesize{$\pm$ 1.2}&\bf{46.2}  \footnotesize{$\pm$ 0.4}\\
ResNet-18 &GeM     &Single Query    &Y &N &N &Pitts30k & \bf{77.8}  \footnotesize{$\pm$ 0.2}&\bf{35.6}  \footnotesize{$\pm$ 0.6}&35.3  \footnotesize{$\pm$ 1.1}&34.2  \footnotesize{$\pm$ 1.7}&\bf{64.3}  \footnotesize{$\pm$ 1.2}&\bf{46.2}  \footnotesize{$\pm$ 0.4}\\
ResNet-18 &GeM     &Central Crop    &Y &N &Y &Pitts30k & \bf{77.8}  \footnotesize{$\pm$ 0.2}&34.8  \footnotesize{$\pm$ 0.5}&\bf{36.4}  \footnotesize{$\pm$ 1.1}&32.6  \footnotesize{$\pm$ 1.4}&\bf{64.3}  \footnotesize{$\pm$ 1.2}&\bf{46.2}  \footnotesize{$\pm$ 0.4}\\
ResNet-18 &GeM     &Five Crops Mean &Y &Y &Y &Pitts30k & 75.4  \footnotesize{$\pm$ 0.3}&30.2  \footnotesize{$\pm$ 0.2}&35.9  \footnotesize{$\pm$ 0.5}&34.4  \footnotesize{$\pm$ 2.0}&59.1  \footnotesize{$\pm$ 0.7}&43.3  \footnotesize{$\pm$ 0.8}\\
ResNet-18 &GeM     &Nearest Crop    &Y &Y &Y &Pitts30k & 74.8  \footnotesize{$\pm$ 0.1}&28.3  \footnotesize{$\pm$ 0.3}&33.8  \footnotesize{$\pm$ 1.3}&\bf{35.7}  \footnotesize{$\pm$ 1.6}&55.5  \footnotesize{$\pm$ 0.8}&39.4  \footnotesize{$\pm$ 0.5}\\
ResNet-18 &GeM     &Majority Voting &Y &Y &Y &Pitts30k & 75.1  \footnotesize{$\pm$ 0.0}&29.1  \footnotesize{$\pm$ 0.4}&34.8  \footnotesize{$\pm$ 1.5}&35.3  \footnotesize{$\pm$ 1.3}&51.8  \footnotesize{$\pm$ 0.2}&41.3  \footnotesize{$\pm$ 0.5}\\

\midrule

ResNet-18 &NetVLAD &Hard Resize     &Y &N &Y &Pitts30k & \bf{86.4}  \footnotesize{$\pm$ 0.3}&47.4  \footnotesize{$\pm$ 1.2}&58.3  \footnotesize{$\pm$ 1.4}&58.9  \footnotesize{$\pm$ 1.1}&76.8  \footnotesize{$\pm$ 1.2}&\bf{57.6}  \footnotesize{$\pm$ 3.3}\\
ResNet-18 &NetVLAD &Single Query    &Y &N &N &Pitts30k & \bf{86.4}  \footnotesize{$\pm$ 0.3}&47.5  \footnotesize{$\pm$ 1.3}&63.4  \footnotesize{$\pm$ 1.2}&61.4  \footnotesize{$\pm$ 1.5}&76.8  \footnotesize{$\pm$ 1.2}&\bf{57.6}  \footnotesize{$\pm$ 3.3}\\
ResNet-18 &NetVLAD &Central Crop    &Y &N &Y &Pitts30k & \bf{86.4}  \footnotesize{$\pm$ 0.3}&\bf{48.0}  \footnotesize{$\pm$ 1.3}&63.2  \footnotesize{$\pm$ 0.2}&57.8  \footnotesize{$\pm$ 0.4}&76.8  \footnotesize{$\pm$ 1.2}&\bf{57.6}  \footnotesize{$\pm$ 3.3}\\
ResNet-18 &NetVLAD &Five Crops Mean &Y &Y &Y &Pitts30k & 85.1  \footnotesize{$\pm$ 0.2}&45.3  \footnotesize{$\pm$ 1.3}&63.0  \footnotesize{$\pm$ 0.7}&60.9  \footnotesize{$\pm$ 1.7}&\bf{78.9}  \footnotesize{$\pm$ 0.9}&54.6  \footnotesize{$\pm$ 2.8}\\
ResNet-18 &NetVLAD &Nearest Crop    &Y &Y &Y &Pitts30k & 84.8  \footnotesize{$\pm$ 0.2}&46.0  \footnotesize{$\pm$ 1.5}&\bf{67.0}  \footnotesize{$\pm$ 1.4}&\bf{64.8}  \footnotesize{$\pm$ 0.7}&75.7  \footnotesize{$\pm$ 1.4}&53.0  \footnotesize{$\pm$ 2.5}\\
ResNet-18 &NetVLAD &Majority Voting &Y &Y &Y &Pitts30k & 84.8  \footnotesize{$\pm$ 0.3}&45.2  \footnotesize{$\pm$ 1.4}&66.9  \footnotesize{$\pm$ 1.1}&64.7  \footnotesize{$\pm$ 0.7}&77.1  \footnotesize{$\pm$ 1.1}&53.4  \footnotesize{$\pm$ 2.3}\\

\midrule

ResNet-50 &GeM     &Hard Resize     &Y &N &Y &Pitts30k & \bf{82.0}  \footnotesize{$\pm$ 0.3}&38.0  \footnotesize{$\pm$ 0.1}&34.6  \footnotesize{$\pm$ 1.4}&40.7  \footnotesize{$\pm$ 1.8}&\bf{66.3}  \footnotesize{$\pm$ 2.5}&\bf{59.0}  \footnotesize{$\pm$ 1.4}\\
ResNet-50 &GeM     &Single Query    &Y &N &N &Pitts30k & \bf{82.0}  \footnotesize{$\pm$ 0.3}&\bf{38.2}  \footnotesize{$\pm$ 0.3}&41.5  \footnotesize{$\pm$ 1.8}&45.4  \footnotesize{$\pm$ 2.0}&\bf{66.3}  \footnotesize{$\pm$ 2.5}&\bf{59.0}  \footnotesize{$\pm$ 1.4}\\
ResNet-50 &GeM     &Central Crop    &Y &N &Y &Pitts30k & \bf{82.0}  \footnotesize{$\pm$ 0.3}&37.5  \footnotesize{$\pm$ 0.3}&40.4  \footnotesize{$\pm$ 0.9}&41.0  \footnotesize{$\pm$ 2.6}&\bf{66.3}  \footnotesize{$\pm$ 2.5}&\bf{59.0}  \footnotesize{$\pm$ 1.4}\\
ResNet-50 &GeM     &Five Crops Mean &Y &Y &Y &Pitts30k & 80.4  \footnotesize{$\pm$ 0.1}&33.2  \footnotesize{$\pm$ 0.1}&39.8  \footnotesize{$\pm$ 2.0}&43.8  \footnotesize{$\pm$ 0.9}&65.0  \footnotesize{$\pm$ 2.4}&54.4  \footnotesize{$\pm$ 1.3}\\
ResNet-50 &GeM     &Nearest Crop    &Y &Y &Y &Pitts30k & 79.2  \footnotesize{$\pm$ 0.2}&30.8  \footnotesize{$\pm$ 0.2}&\bf{43.5}  \footnotesize{$\pm$ 1.4}&\bf{46.9}  \footnotesize{$\pm$ 1.4}&63.5  \footnotesize{$\pm$ 2.2}&52.6  \footnotesize{$\pm$ 1.4}\\
ResNet-50 &GeM     &Majority Voting &Y &Y &Y &Pitts30k & 79.7  \footnotesize{$\pm$ 0.0}&31.5  \footnotesize{$\pm$ 0.1}&43.0  \footnotesize{$\pm$ 2.0}&44.8  \footnotesize{$\pm$ 1.2}&62.9  \footnotesize{$\pm$ 2.3}&52.8  \footnotesize{$\pm$ 0.9}\\

\midrule
ResNet-50 &NetVLAD &Hard Resize     &Y &N &Y &Pitts30k & \bf{86.0}  \footnotesize{$\pm$ 0.1}&50.7  \footnotesize{$\pm$ 2.0}&64.3  \footnotesize{$\pm$ 1.9}&64.3  \footnotesize{$\pm$ 1.2}&77.7  \footnotesize{$\pm$ 0.4}&\bf{60.2}  \footnotesize{$\pm$ 1.6}\\
ResNet-50 &NetVLAD &Single Query    &Y &N &N &Pitts30k & \bf{86.0}  \footnotesize{$\pm$ 0.1}&50.6  \footnotesize{$\pm$ 1.9}&69.8  \footnotesize{$\pm$ 0.8}&67.1  \footnotesize{$\pm$ 2.3}&77.7  \footnotesize{$\pm$ 0.4}&\bf{60.2}  \footnotesize{$\pm$ 1.6}\\
ResNet-50 &NetVLAD &Central Crop    &Y &N &Y &Pitts30k & \bf{86.0}  \footnotesize{$\pm$ 0.1}&\bf{50.9}  \footnotesize{$\pm$ 1.9}&68.3  \footnotesize{$\pm$ 1.4}&64.6  \footnotesize{$\pm$ 2.2}&77.7  \footnotesize{$\pm$ 0.4}&\bf{60.2}  \footnotesize{$\pm$ 1.6}\\
ResNet-50 &NetVLAD &Five Crops Mean &Y &Y &Y &Pitts30k & 84.7  \footnotesize{$\pm$ 0.1}&47.4  \footnotesize{$\pm$ 1.9}&68.0  \footnotesize{$\pm$ 2.2}&66.5  \footnotesize{$\pm$ 1.5}&\bf{78.6}  \footnotesize{$\pm$ 0.3}&54.3  \footnotesize{$\pm$ 2.8}\\
ResNet-50 &NetVLAD &Nearest Crop    &Y &Y &Y &Pitts30k & 84.2  \footnotesize{$\pm$ 0.2}&47.0  \footnotesize{$\pm$ 1.7}&72.3  \footnotesize{$\pm$ 1.3}&\bf{68.4}  \footnotesize{$\pm$ 0.8}&76.8  \footnotesize{$\pm$ 0.5}&52.3  \footnotesize{$\pm$ 2.3}\\
ResNet-50 &NetVLAD &Majority Voting &Y &Y &Y &Pitts30k & 84.3  \footnotesize{$\pm$ 0.2}&47.1  \footnotesize{$\pm$ 1.7}&\bf{72.8}  \footnotesize{$\pm$ 0.8}&68.1  \footnotesize{$\pm$ 1.3}&77.5  \footnotesize{$\pm$ 0.4}&53.4  \footnotesize{$\pm$ 2.2}\\
    \bottomrule
  \end{tabular}}
    \caption{\textbf{Query pre/post-processing.} Results with different pre/post-processing methods are shown in the table. The batch parallelization column indicates if images have to be processed one by one or if they can be stacked in a batch for parallel computation.}
  \label{tab:t6_pre_post_processing}
\end{table*}

In a real-world geo-localization system, the queries fed to the software at production time may have different resolutions than the database images. A handful of datasets (\eg, R-SF \cite{Torii-2021,Li-2012}, Tokyo 24/7 \cite{Torii-2018}) incorporates this variability,
and the solution often used in previous works to handle such cases  \cite{Arandjelovic-2018, Radenovic-2019, Tolias-2016, Gordo-2017, Revaud-2019} is to use a batch size of 1 when extracting query descriptors at inference time. This approach can give good results at the cost of slower computation, which may or may not be an issue depending on the application's scalability requirements.
Besides this common choice, we experiment with other engineering solutions that allow stacking multiple queries in a batch, investigating if it is possible to simultaneously also improve the recalls.
We group the methods into pre-processing, post-processing, and predictions refinement, according to where in the pipeline they are applied (see diagram in Fig. 1 of main paper).

In Tab. \ref{tab:t6_pre_post_processing} we report the full results of our pre/post-processing and predictions refinement experiments, while the following is a thorough explanation of how such methods are applied.
With respect to pre-processing approaches, in {\bf Hard Resize} we perform an anisotropic resize of the query to the same dimension as the database images (effectively leaving the query unchanged if the query and database images' dimensions already match);
for {\bf Single Query} we isotropically resize so that the query's shortest side is equal to the database images' shortest side, the aspect ratio is preserved, images are not padded, and if they are of varying resolutions, they cannot be stacked in a batch; in {\bf Central Crop} we isotropically resize to the smallest resolution that can accommodate a rectangular region of the size of the database images, and a central crop of such size is taken;
in {\bf Five Crops} we produce five square crops of the database images shortest side.

Regarding post-processing and predictions refinement methods, with {\bf Mean} we simply compute the mean of the descriptors of the five crops; in {\bf Nearest Crop} we choose the prediction with shortest descriptors distance from at least one crop; with {\bf Majority Voting} we implement a voting mechanism taking into account the distances from each crop's first 20 predictions.

\myparagraph{Discussion}
It can be seen in Tab. \ref{tab:t6_pre_post_processing} that, as expected, \emph{Hard Resize}, \emph{Single Query} and \emph{Central Crop} produce exactly the same results when queries have the same size as the database images (Pitts30k, Eynsham and St Lucia), as in these cases they correspond to an identity transformation.
On the other hand, in Tokyo 24/7 and R-SF, where roughly half of the queries are vertical (\ie, the height is longer than the width), more complex techniques, such as \emph{Nearest Crop} and \emph{Majority Voting}, on average yield better results. This is particularly noticeable with more robust networks.
Furthermore, these methods allow multiple queries to be stacked in a single batch, as the crops they operate on all have the same dimensions.
Finally, we can state that the ideal approach highly depends on the application:
for robotics, if all images come from the same devices (and have the same resolution as database images), simply applying \emph{Hard Resize} (which in this case results in no resize at all) leads to best results; 
in cases where queries can have unrestricted resolutions, \emph{Single Query} represents a simple approach with acceptable results, while \emph{Nearest Crop} produces the best R@1 and offers the possibility of batch parallelization, which is crucial for scalability.

%%%%%%%%%%%%%%%%%%%%%%%%%%%%%%%% NEAREST NEIGHBOR
%%%%%%%%%%%%%%%%%%%%%%%%%%%%%%%% NEAREST NEIGHBOR
%%%%%%%%%%%%%%%%%%%%%%%%%%%%%%%% NEAREST NEIGHBOR

\subsection{Nearest Neighbor Search and Inference Time}
\label{sec:supp_knn_indexing}
As stated in the main paper, matching time can significantly impact inference time (see Sec. 4.7 of the main paper) and memory footprint.
This section reports experiments with the different indexing techniques listed in the main paper, reporting extensive results on all datasets. The goal is to investigate how and if it is possible to make this computation more efficient.

\begin{figure*}[ht!]
    \centering
    \begin{minipage}{.32\textwidth}
        \begin{subfigure}{\textwidth}
            \includegraphics[width=\textwidth]{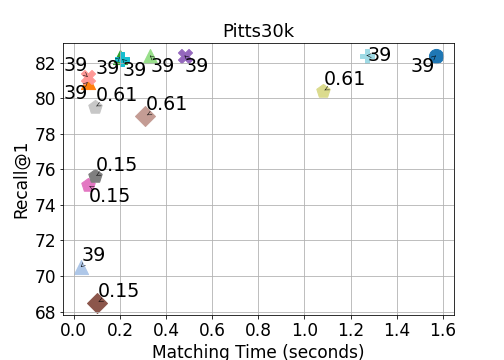}
        \end{subfigure}
        \begin{subfigure}{\textwidth}
            \includegraphics[width=\textwidth]{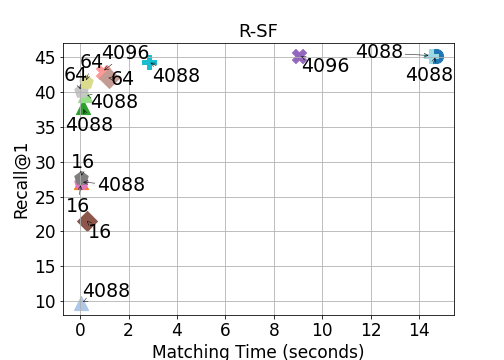} 
        \end{subfigure}
    \end{minipage}
    \begin{minipage}{.32\textwidth}
        \begin{subfigure}{\textwidth}
            \includegraphics[width=\textwidth]{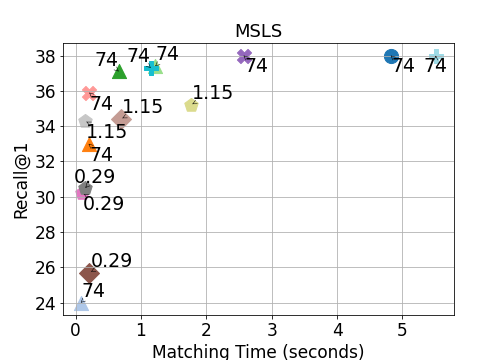} 
        \end{subfigure}
        \begin{subfigure}{\textwidth}
        \includegraphics[width=\textwidth]{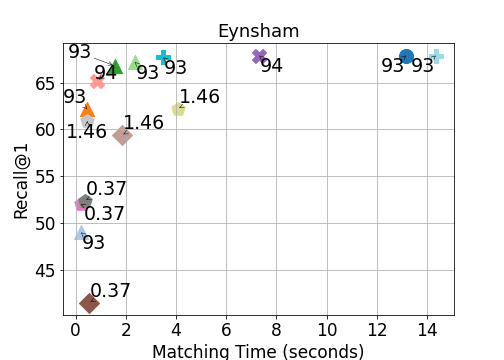} 
        \end{subfigure}
    \end{minipage}
    \begin{minipage}{.32\textwidth}
        \begin{subfigure}{\textwidth}
            \includegraphics[width=\textwidth]{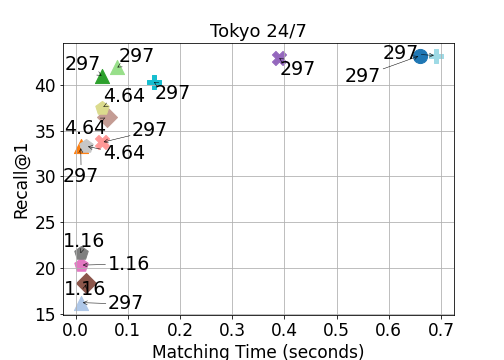}
            \vspace*{1px}
        \end{subfigure}
        \begin{subfigure}{\textwidth}
            \includegraphics[width=\textwidth]{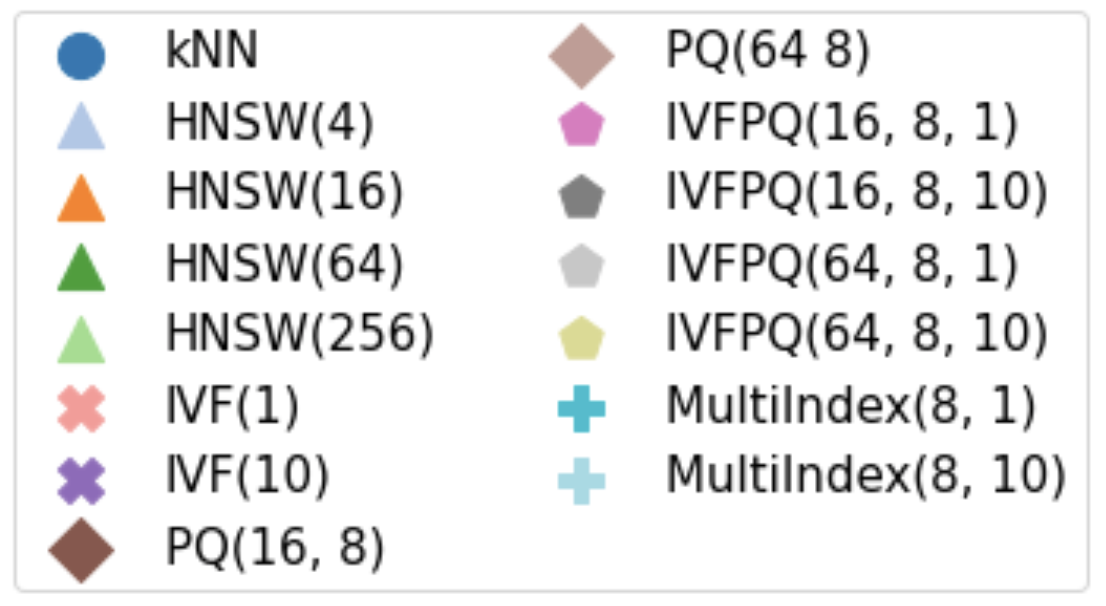} 
            \vspace*{2mm}
        \end{subfigure}
    \end{minipage}
    \caption{\textbf{Optimized kNN indexing: faster search \& lower memory footprint.} 
    The plots shows a number of efficient kNN variants, on different datasets, which are applied on features extracted with a ResNet-50 + GeM (features dimension 1024) trained on Pitts30k.
    On the x-axis is the matching time in seconds for all the queries in each dataset, while on the y-axis is the recall@1. The numbers next to the dots represent the RAM requirements of the method (memory footprint) in MB. Besides exhaustive kNN, we employ inverted file indexes (IVF) \cite{Sivic-2003}, product quantization (with and without inverted indexes, respectively PQ and IVFPQ) \cite{Jegou-2011_productQ}, the inverted multi index (MultiIndex) \cite{BabenkoL12} and hierarchical navigable small world graphs (HNSW) \cite{Malkov2020EfficientAR}. In the legend, the parameters are shown for each method. The last parameter of IVFPQ, MultiIndex and IVF, which is either 1 or 10, represents the percentage of Voronoi cells to search, given that the search space has been split into 1000 Voronoi cells.}
    \label{fig:supp_knn_indexing}
\end{figure*}

\myparagraph{Discussion}
In Fig. \ref{fig:supp_knn_indexing} we show results for various methods. Among the most outstanding results, using an Inverted File Index (IVF)~\cite{Sivic-2003} can lead to a reduction of matching time of roughly 20 times, while lowering the recall@1 of less than 2\% on average.
The Hierarchical Navigable Small World graphs (HNSW)~\cite{Malkov2020EfficientAR} and the Inverted Multi-Index (MultiIndex) \cite{BabenkoL12} bring similar achievements as the Inverted File Index, with slightly slower computation but higher recall.
Regarding memory footprint, the Inverted File Index with Product Quantization (IVFPQ)~\cite{Jegou-2011_productQ} can reduce it by a factor of 64, and in the largest dataset, namely Revisited San Francisco (R-SF), it reduces matching time by 98.5\%, with a drop in accuracy from 45.4\% to 41.4\%. However, the improvements become less obvious as the size of the database diminishes, as shown in the plots.
Compared to the Inverted File Index with Product Quantization, the simpler Product Quantization leads to the same memory savings but lower recall-speed ratio, making Inverted File Index with Product Quantization the Pareto optimal solution concerning the recall@1 and the matching time when memory efficiency is an issue.

\begin{table*}[!ht]
  \centering
  \resizebox{\textwidth}{!}{
  \begin{tabular}{lllllllllllll}
    \toprule
    \multirow{2}{*}{Source} &
    \multirow{2}{*}{Loss} &
    \begin{tabular}[c]{@{}l@{}}Training\\ Dataset\end{tabular} &
    \multirow{2}{*}{Backbone} &
    \begin{tabular}[c]{@{}l@{}}Aggregation\\ Method\end{tabular} &
    \begin{tabular}[c]{@{}l@{}}R@1\\ Pitts30k\end{tabular} &
    \begin{tabular}[c]{@{}l@{}}R@1\\ MSLS\end{tabular} &
    \begin{tabular}[c]{@{}l@{}}R@1\\ Tokyo 24/7\end{tabular} &
    \begin{tabular}[c]{@{}l@{}}R@1\\ R-SF\end{tabular} &
    \begin{tabular}[c]{@{}l@{}}R@1\\ Eynsham\end{tabular} &
    \begin{tabular}[c]{@{}l@{}}R@1\\ St Lucia\end{tabular}  \\
    \midrule
    \cite{Radenovic-2019} &Triplet   &GLDv1    &ResNet-50 &GeM + FC 2048 &\textbf{84.1} &69.5 &\textbf{77.8} &\textbf{76.4} &61.8 &77.3  \\
    \cite{Radenovic-2019} &Triplet   &Sfm120k &ResNet-50 &GeM + FC 2048 &83.4 &64.5 &75.2 &75.6 &68.8 &73.9  \\
    - & Triplet & Pitts30k & ResNet-50 &GeM + FC 2048 &80.1 &33.7 &43.6 &48.2 &70.0 &56.0  \\
    - & Triplet & MSLS     & ResNet-50 &GeM + FC 2048 &79.2 &\textbf{73.5} &64.0 &55.1 &\textbf{86.1} &\textbf{90.3}  \\
    \midrule
    \cite{Radenovic-2019} &Triplet   &GLDv1        &ResNet-101 &GeM + FC 2048 &\textbf{85.1} &72.4 &\textbf{77.8} &\textbf{79.8} &61.6 &83.4  \\
    \cite{Radenovic-2019} &Triplet   &Sfm120k     &ResNet-101 &GeM + FC 2048 &83.9 &64.7 &77.5 &78.3 &62.8 &76.3  \\
    - & Triplet & Pitts30k & ResNet-101 &GeM + FC 2048 & 82.4 &40.0 &47.2 &57.5 &75.9 &61.7 \\
    - & Triplet & MSLS     & ResNet-101 &GeM + FC 2048 & 79.1 &\textbf{75.3} &61.9 &54.9 &\textbf{86.0} &\textbf{92.5}  \\
    \bottomrule
  \end{tabular}}
   \caption{\textbf{The role of the training dataset.} The table shows results with models trained on large scale landmark retrieval datasets.}
  \label{tab:t2b_sota_lr}
\end{table*}

\section{Additional Experiments}
\label{sec:supp_additional_experiments}

While the different components of a Visual Geo-localization system, as shown in Fig. 1 of the main paper, have been thoroughly studied in Sec. 4 of the main paper and \cref{sec:supp_ext_results}, in this section, we investigate several other factors.
In \cref{sec:supp_role_training_dataset}, we aim to understand if models trained on large-scale landmark retrieval datasets can be reliably used for VG.
In \cref{sec:supp_distractors}, we use images from those datasets as distractors to increase the size of the database of various orders of magnitude.
The same landmark retrieval datasets are used in \cref{sec:supp_pretrain} to see if models pretrained on landmark retrieval can be easily fine-tuned for VG.
Finally, in \cref{sec:supp_metrics}, we explore the use of different metrics and how they relate to the final results.

%%%%%%%%%%%%%%%%%%%%%%%%%%%%%%%% ROLE OF TRAINING DATASET
%%%%%%%%%%%%%%%%%%%%%%%%%%%%%%%% ROLE OF TRAINING DATASET
%%%%%%%%%%%%%%%%%%%%%%%%%%%%%%%% ROLE OF TRAINING DATASET

\subsection{The role of the training dataset}
\label{sec:supp_role_training_dataset}

In this section, we further investigate the role of the training dataset.
To this end, we compute results with publicly-available state-of-the-art models for image retrieval\footnote{{\footnotesize\url{https://github.com/filipradenovic/cnnimageretrieval-pytorch}}}, which have been trained on large scale landmark retrieval datasets, and we compare them with analogous networks trained on Pitts30k and MSLS.
Note that higher recalls could have been achieved using a NetVLAD+PCA, but the GeM + FC aggregation was preferred to obtain a fair comparison with the models provided by third parties as state-of-the-art trained on the landmark datasets.
Our framework easily allows for retrieval models trained by \cite{Radenovic-2019} and \cite{Revaud-2019} to be automatically downloaded from their GitHub repositories and used to perform experiments on Visual Geo-localization datasets.

\myparagraph{Discussion}
Results are reported in Tab. \ref{tab:t2b_sota_lr}.
The experiments confirm that the choice of the training dataset plays a significant role in a network's robustness and generalization capabilities.
We notice that models trained on large-scale landmark retrieval datasets, especially on GLDv1 \cite{Noh-2017}, offer a decent off-the-shelf solution for many Visual Geo-localization datasets.
In general, we see that these models benefit from the wide variety of images present in the landmark retrieval datasets and are able to learn robust features that guarantee good generalization performances in the VG setting. 
As for the training directly on VG datasets, it is noticeable how models trained on MSLS, thanks to its bigger size and variability, achieve more robustness on all datasets except for R-SF and Tokyo 24/7 than the same models trained on Pitts30k. The reason behind this fact is that the images of these last two datasets are made up of 360° views, unlike the front view scenarios that the model sees during training on MSLS.
Finally, the poor generalization performances obtained training on Pitts30k can be understood in relation to the use of the rather big (in terms of the number of parameters) FC layer that inevitably leads to overfitting on the small size of the said dataset; in fact, Tab. \ref{tab:t2_aggregation_methods_complete} shows how using as aggregator a NetVLAD + PCA method, significantly reducing the number of parameters, leads to a better generalization.
The takeaway messages should be to use the training set that presents similar viewpoints, if known, or otherwise the more general one, and choose the model with a number of parameters proportional to the dataset size.

These results are directly comparable with results from Tab. \ref{tab:t2_aggregation_methods_complete}.

%%%%%%%%%%%%%%%%%%%%%%%%%%%%%%%% SCALING DATASETS WITH DISTRACTORS
%%%%%%%%%%%%%%%%%%%%%%%%%%%%%%%% SCALING DATASETS WITH DISTRACTORS
%%%%%%%%%%%%%%%%%%%%%%%%%%%%%%%% SCALING DATASETS WITH DISTRACTORS

\subsection{Scaling datasets with distractors}
\label{sec:supp_distractors}
While in the past few years, software and hardware improvements have allowed us to obtain better and faster results, common datasets are still on a small to medium scale, and their coverage is still shallow compared to a realistic real-world application (see Tab. \ref{tab:datasets_size}).
As an example, the San Francisco dataset, although being one of the largest with a database of 1 million images, still covers only 9\% of the city of San Francisco.
As in our work, we aim to investigate VG's possible applications, and then we built a large-scale dataset with up to 8 million distractors.
To this end, we used the 315 queries from Tokyo 24/7, and first built a small-scale database with their positives and several random images from Tokyo 24/7 database (to reach a total of 10.000 pictures).
A 10 times bigger set was then built using the whole Tokyo 24/7 database, as well as the ones from Pitts30k and MSLS test sets.
By using the database of San Francisco, we reached 1 million images, and, finally, we scaled it to 8 million by including the whole Google Landmark v2~\cite{Weyand-2020}, Places 365~\cite{Zhou-2017}, and MSLS train database.

\begin{figure}[t]
  \centering
  \includegraphics[width=.4\textwidth]{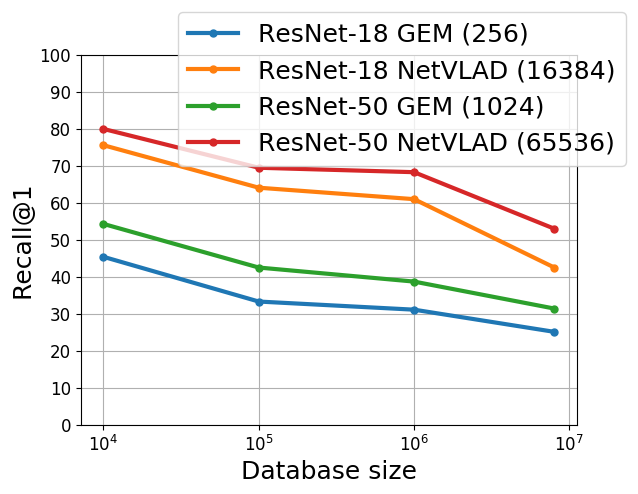}
  \caption{\textbf{Scaling datasets with distractors.} The plot shows the effects of exponentially increasing the size of the database up to 8M. In the legend, the descriptors dimensionality is shown between parentheses.}
  \label{fig:distractors}
\end{figure}

\myparagraph{Discussion}
Results are shown in Fig. \ref{fig:distractors}.
Unsurprisingly, we see that results steadily decrease as the size of the database increases, proving that the task is still far from solved.

%%%%%%%%%%%%%%%%%%%%%%%%%%%%%%%% ALL-DATASETS BENCHMARK
%%%%%%%%%%%%%%%%%%%%%%%%%%%%%%%% ALL-DATASETS BENCHMARK
%%%%%%%%%%%%%%%%%%%%%%%%%%%%%%%% ALL-DATASETS BENCHMARK
\subsection{Testing on an ensemble of datasets}
\label{sec:supp_all_dataset}
The experiment presented in \cref{sec:supp_distractors} investigates how VG methods perform on a large-scale database built with 8M distractors, but with the queries all taken from a single dataset. 
However, in practice, the VG model may be tasked to geolocalize queries coming from different data distributions and geographical areas (e.g., Tokyo, San Francisco, \etc). 
To investigate how VG models fare in this situation, the benchmark also supports experiments considering an ensemble dataset that combines all the test queries and databases considered so far, \ie Pitts30k, MSLS, Tokyo 24/7, R-SF, Eynsham, and St Lucia.
In particular, here we report the results achieved matching the queries only to their respective database (\textit{Single DB}) and matching the queries to all six databases merged (\textit{Multi DB}).

\begin{table}[t!]
\centering
  \resizebox{\columnwidth}{!}{
\begin{tabular}{lcccc}
\toprule
\multirow{2}{*}{Model} &
\multicolumn{2}{c}{Trained on Pitts30k} & \multicolumn{2}{c@{}}{Trained on MSLS} \\
\cmidrule(lr){2-3} \cmidrule(l){4-5}
& R@1 Single DB & R@1 Multi DB & R@1 Single DB & R@1 Multi DB \\
\midrule
ResNet-18 + GeM & 57.9 & 42.2 (-27.1\%) & 74.4 & 65.1 (-12.5\%)  \\
ResNet-50 + GeM & 60.9 & 53.4 (-12.3\%) & 79.4 & 71.6 (-9.82\%)  \\
ResNet-18 + NetVLAD & 70.0 & 67.4 (-3.7\%)  & 83.0 & 79.0 (-4.1\%) \\
ResNet-50 + NetVLAD & 71.4 & 68.7 (-3.8\%)  & 83.2 & 79.0 (-5.0\%) \\
\bottomrule
\end{tabular}
}
\caption{
\textbf{All-data benchmark.}
Using all queries from the six datasets, \emph{Single DB} indicates the average result from matching the queries only to their respective database, \emph{Multi DB} refers to matching the the queries to all six databases merged.
}
\label{tab:all_data_benchmark}
\vspace{-0.6cm}
\end{table}

\myparagraph{Discussion}
The results in \cref{tab:all_data_benchmark} report the R@1 achieved by various methods both in the Single DB and Multi DB settings. In the Multi DB setting there is a clear decrease in recall with respect to averaging across the different datasets tested separately (Single DB), which demonstrates the difficulty of this scenario. Overall, the models trained on MSLS achieve better results than those trained on Pitts30k, which confirms that the larger number and variety of images of MSLS has a notable impact on the generalization capability of the model. We also observe that the percentile drop on the R@1 when going from the Single DB to the Multi DB setting is higher with GeM than with NetVLAD.

%%%%%%%%%%%%%%%%%%%%%%%%%%%%%%%% PRETRAINING
%%%%%%%%%%%%%%%%%%%%%%%%%%%%%%%% PRETRAINING
%%%%%%%%%%%%%%%%%%%%%%%%%%%%%%%% PRETRAINING

\subsection{Pretraining the backbone on other datasets}
\label{sec:supp_pretrain}
In this section, we investigated whether pretraining the backbone of our VG system on datasets different from ImageNet can be beneficial for the training of the model. The datasets used for this purpose were Places 365 \cite{Zhou-2017}, a dataset for scene recognition, and Google Landmark v2 (GLDv2) \cite{Noh-2017, Weyand-2020}, a recent large-scale landmark retrieval/recognition dataset released by Google.
The networks were trained with a standard classification approach for Places and using the ArcFace Loss \cite{Deng-2019} on GLDv2, following the idea proposed by \cite{Yokoo-2020}.

\begin{table*}[!ht]
  \centering
  \resizebox{\textwidth}{!}{
  \begin{tabular}{llllllllll}
    \toprule
    \multirow{2}{*}{Backbone} & 
    \begin{tabular}[c]{@{}l@{}}Aggregation\\ Method\end{tabular} & \begin{tabular}[c]{@{}l@{}}Dataset\end{tabular} &
    \begin{tabular}[c]{@{}l@{}}Training \\ Dataset\end{tabular} &
    \begin{tabular}[c]{@{}l@{}}R@1\\ Pitts30k\end{tabular} & 
    \begin{tabular}[c]{@{}l@{}}R@1\\ MSLS\end{tabular} & 
    \begin{tabular}[c]{@{}l@{}}R@1 \\ Tokyo 24/7\end{tabular} &
    \begin{tabular}[c]{@{}l@{}}R@1\\ R-SF\end{tabular} &
    \begin{tabular}[c]{@{}l@{}}R@1\\ Eynsham\end{tabular} &
    \begin{tabular}[c]{@{}l@{}}R@1\\ St Lucia\end{tabular} \\
    \midrule
    ResNet-18  &GeM      &ImageNet    &Pitts30k & 77.8  \footnotesize{$\pm$ 0.2}&35.3  \footnotesize{$\pm$ 0.5}&{\bf35.3}  \footnotesize{$\pm$ 1.1}&{\bf34.2}  \footnotesize{$\pm$ 1.7}&64.3  \footnotesize{$\pm$ 1.2}&46.2  \footnotesize{$\pm$ 0.4}\\
ResNet-18  &GeM      &GLDv2       &Pitts30k & 74.2  \footnotesize{$\pm$ 0.4}&30.9  \footnotesize{$\pm$ 0.6}&22.3  \footnotesize{$\pm$ 1.9}&20.4  \footnotesize{$\pm$ 1.7}&55.0  \footnotesize{$\pm$ 2.0}&43.3  \footnotesize{$\pm$ 0.7}\\
ResNet-18  &GeM      &Places 365  &Pitts30k & {\bf78.1}  \footnotesize{$\pm$ 1.0}&{\bf36.2}  \footnotesize{$\pm$ 0.9}&31.8  \footnotesize{$\pm$ 0.7}&32.8  \footnotesize{$\pm$ 1.6}&{\bf65.0}  \footnotesize{$\pm$ 2.1}&{\bf48.8}  \footnotesize{$\pm$ 2.1}\\
\midrule
ResNet-18  &NetVLAD  &ImageNet    &Pitts30k & {\bf 86.4}  \footnotesize{$\pm$ 0.3}&{\bf 47.4}  \footnotesize{$\pm$ 1.2}&{\bf 63.4}  \footnotesize{$\pm$ 1.2}& {\bf 61.4}  \footnotesize{$\pm$ 1.5}&76.8  \footnotesize{$\pm$ 1.2}&{\bf 57.6}  \footnotesize{$\pm$ 3.3}\\
ResNet-18  &NetVLAD  &GLDv2       &Pitts30k & 83.3  \footnotesize{$\pm$ 0.5}&39.9  \footnotesize{$\pm$ 0.9}&54.2  \footnotesize{$\pm$ 2.3}&41.1  \footnotesize{$\pm$ 3.6}&71.4  \footnotesize{$\pm$ 2.6}&46.8  \footnotesize{$\pm$ 1.9}\\
ResNet-18  &NetVLAD  &Places 365  &Pitts30k & 85.9  \footnotesize{$\pm$ 0.4}&{\bf 47.4}  \footnotesize{$\pm$ 0.6}&57.9  \footnotesize{$\pm$ 1.4}&59.9  \footnotesize{$\pm$ 3.2}&{\bf 78.7}  \footnotesize{$\pm$ 0.7}&50.4  \footnotesize{$\pm$ 1.0}\\
\midrule
ResNet-50  &GeM      &ImageNet    &Pitts30k & 82.0  \footnotesize{$\pm$ 0.3}&38.0  \footnotesize{$\pm$ 0.1}&{\bf41.5}  \footnotesize{$\pm$ 1.8}&{\bf45.4}  \footnotesize{$\pm$ 2.0}&66.3  \footnotesize{$\pm$ 2.5}&59.0  \footnotesize{$\pm$ 1.4}\\
ResNet-50  &GeM      &GLDv2       &Pitts30k & 77.9  \footnotesize{$\pm$ 0.5}&35.2  \footnotesize{$\pm$ 0.8}&27.6  \footnotesize{$\pm$ 2.1}&37.2  \footnotesize{$\pm$ 1.0}&62.7  \footnotesize{$\pm$ 1.6}&48.4  \footnotesize{$\pm$ 1.7}\\
ResNet-50  &GeM      &Places 365  &Pitts30k & {\bf82.5}  \footnotesize{$\pm$ 0.4}&{\bf40.8}  \footnotesize{$\pm$ 0.3}&41.3  \footnotesize{$\pm$ 0.7}&45.3  \footnotesize{$\pm$ 0.6}&{\bf66.9}  \footnotesize{$\pm$ 1.3}&{\bf 60.8}  \footnotesize{$\pm$ 1.6}\\
\midrule
ResNet-50  &NetVLAD  &ImageNet    &Pitts30k & 86.0  \footnotesize{$\pm$ 0.1}&{\bf 50.7}  \footnotesize{$\pm$ 2.0}&{\bf 69.8}  \footnotesize{$\pm$ 0.8}&{\bf 67.1}  \footnotesize{$\pm$ 2.3}&{\bf 77.7}  \footnotesize{$\pm$ 0.4}&{\bf60.2}  \footnotesize{$\pm$ 1.6}\\
ResNet-50  &NetVLAD  &GLDv2       &Pitts30k & 81.7  \footnotesize{$\pm$ 0.6}&43.5  \footnotesize{$\pm$ 1.0}&56.7  \footnotesize{$\pm$ 0.9}&54.1  \footnotesize{$\pm$ 1.8}&71.4  \footnotesize{$\pm$ 0.6}&42.3  \footnotesize{$\pm$ 2.5}\\
ResNet-50  &NetVLAD  &Places 365  &Pitts30k & {\bf 86.2}  \footnotesize{$\pm$ 0.5}&49.9  \footnotesize{$\pm$ 2.0}&66.3  \footnotesize{$\pm$ 3.3}&59.7  \footnotesize{$\pm$ 3.5}&75.4  \footnotesize{$\pm$ 2.0}&57.2  \footnotesize{$\pm$ 5.5}\\
\midrule
\midrule
ResNet-18  &GeM      &ImageNet    &MSLS     & {\bf71.6}  \footnotesize{$\pm$ 0.1}&{\bf65.3}  \footnotesize{$\pm$ 0.2}&{\bf42.8}  \footnotesize{$\pm$ 1.1}&{\bf30.5}  \footnotesize{$\pm$ 0.8}&{\bf80.3}  \footnotesize{$\pm$ 0.1}&{\bf83.2}   \footnotesize{$\pm$ 0.9}\\
ResNet-18  &GeM      &GLDv2       &MSLS     & 60.7  \footnotesize{$\pm$ 0.5}&64.5  \footnotesize{$\pm$ 0.7}&30.9  \footnotesize{$\pm$ 3.3}&21.5  \footnotesize{$\pm$ 0.8}&79.2  \footnotesize{$\pm$ 0.6}&78.1   \footnotesize{$\pm$ 1.0}\\
ResNet-18  &GeM      &Places 365  &MSLS     & {\bf71.6}  \footnotesize{$\pm$ 0.9}&64.8  \footnotesize{$\pm$ 1.1}&36.6  \footnotesize{$\pm$ 2.2}&25.5  \footnotesize{$\pm$ 0.3}&80.1  \footnotesize{$\pm$ 0.5}&82.4   \footnotesize{$\pm$ 0.6}\\
\midrule
ResNet-18  &NetVLAD  &ImageNet    &MSLS     &  {\bf 81.6}  \footnotesize{$\pm$ 0.5}&{\bf 75.8}  \footnotesize{$\pm$ 0.1}&{\bf 62.3}  \footnotesize{$\pm$ 1.6}&{\bf 55.1}  \footnotesize{$\pm$ 0.9}&{\bf 87.1}  \footnotesize{$\pm$ 0.2}& {\bf 92.1}   \footnotesize{$\pm$ 0.7}\\
ResNet-18  &NetVLAD  &GLDv2       &MSLS     & 73.3  \footnotesize{$\pm$ 0.6}&75.3  \footnotesize{$\pm$ 0.3}&53.4  \footnotesize{$\pm$ 1.3}&40.7  \footnotesize{$\pm$ 2.9}&86.1  \footnotesize{$\pm$ 0.1}&87.6   \footnotesize{$\pm$ 0.9}\\
ResNet-18  &NetVLAD  &Places 365  &MSLS     & 79.7  \footnotesize{$\pm$ 0.5}&75.6  \footnotesize{$\pm$ 0.2}&61.5  \footnotesize{$\pm$ 0.7}&48.6  \footnotesize{$\pm$ 1.5}&86.5  \footnotesize{$\pm$ 0.1}&90.4   \footnotesize{$\pm$ 0.4}\\
\midrule
ResNet-50  &GeM      &ImageNet    &MSLS     & 77.4  \footnotesize{$\pm$ 0.6}&72.0  \footnotesize{$\pm$ 0.5}&{\bf55.4}  \footnotesize{$\pm$ 2.5}&{\bf45.7}  \footnotesize{$\pm$ 1.0}&83.9  \footnotesize{$\pm$ 0.6}&{\bf91.2}   \footnotesize{$\pm$ 0.7}\\
ResNet-50  &GeM      &GLDv2       &MSLS     & 71.1  \footnotesize{$\pm$ 1.7}&72.4  \footnotesize{$\pm$ 0.2}&47.6  \footnotesize{$\pm$ 0.4}&35.8  \footnotesize{$\pm$ 1.6}&84.0  \footnotesize{$\pm$ 0.4}&86.1   \footnotesize{$\pm$ 1.1}\\
ResNet-50  &GeM      &Places 365  &MSLS     & {\bf78.2}  \footnotesize{$\pm$ 1.1}&{\bf72.7}  \footnotesize{$\pm$ 0.6}&51.8  \footnotesize{$\pm$ 2.7}&41.8  \footnotesize{$\pm$ 2.2}&{\bf84.4}  \footnotesize{$\pm$ 0.2}&89.3   \footnotesize{$\pm$ 0.8}\\
\midrule
ResNet-50  &NetVLAD  &ImageNet    &MSLS     & {\bf 80.9}  \footnotesize{$\pm$ 0.0}&76.9  \footnotesize{$\pm$ 0.2}&{\bf 62.8}  \footnotesize{$\pm$ 0.9}&{\bf 51.5}  \footnotesize{$\pm$ 1.2}&{\bf 87.2}  \footnotesize{$\pm$ 0.3}&{\bf 93.8}   \footnotesize{$\pm$ 0.2}\\
ResNet-50  &NetVLAD  &GLDv2       &MSLS     & 74.7  \footnotesize{$\pm$ 1.0}&{\bf 77.4}  \footnotesize{$\pm$ 0.4}&55.0  \footnotesize{$\pm$ 1.7}&45.4  \footnotesize{$\pm$ 1.5}&85.1  \footnotesize{$\pm$ 0.5}&87.7   \footnotesize{$\pm$ 0.8}\\
ResNet-50  &NetVLAD  &Places 365  &MSLS     & 80.0  \footnotesize{$\pm$ 1.1}&75.6  \footnotesize{$\pm$ 0.1}&51.3  \footnotesize{$\pm$ 3.3}&44.8  \footnotesize{$\pm$ 2.3}&86.9  \footnotesize{$\pm$ 0.1}&91.3   \footnotesize{$\pm$ 0.2}\\
    \bottomrule
  \end{tabular}}
    \caption{\textbf{Pretraining the backbone on other datasets.}}
  \label{tab:t5_pretrain}
\end{table*}

\myparagraph{Discussion}
From the substantial number of experiments reported in Tab. \ref{tab:t5_pretrain} the evidence is that in the vast majority of the cases it is not convenient to choose a dataset other than ImageNet to pretrain the backbone; however, the scores are quite close except in the cases of R-SF and Tokyo 24/7 datasets, where using a different pretrain dataset in place of ImageNet on average leads to a drop in performance.
Even in the few cases where GLDv2 or Places365 achieves the highest score, the gap is, in practice, negligible. 
Furthermore, if we take into account the off-the-shelf availability of ImageNet pretrained backbones with respect to the far less common alternatives, it is even more clear that the former is the more advantageous choice.

%%%%%%%%%%%%%%%%%%%%%%%%%%%%%%%% METRICS
%%%%%%%%%%%%%%%%%%%%%%%%%%%%%%%% METRICS
%%%%%%%%%%%%%%%%%%%%%%%%%%%%%%%% METRICS

\subsection{Metrics}
\label{sec:supp_metrics}
In previous experiments, we showed results based on recall@1, with a positive distance of 25 meters. In this section, we explore the use of different metrics.
Specifically, we show how results would change if the heading is taken into account or setting the success threshold to other values than 25 meters. Moreover, we use various values of recalls.
We compute these results using the models trained for Tab. \ref{tab:t2_aggregation_methods_complete} on Pitts30k and changing only the final metric at test time.

\subsubsection{Taking heading/yaw/compass into account}
\label{sec:supp_heading}
\begin{figure}[t]
  \centering
  \includegraphics[width=.3\textwidth]{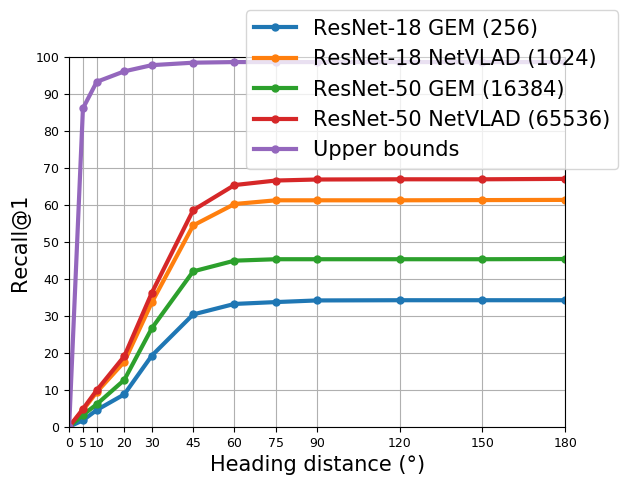}
  \caption{\textbf{Taking heading into account.} Recall@1 on R-SF when heading is taken into account. Different degrees of headings are taken into account on the x-axis. The "Upper bounds" curve refers to the percentage of queries that have at least one database image closer than 25 meters with a difference in heading lower than the given threshold. This corresponds to the upper bound of the recall@1.}
  \label{fig:heading_sf}
\end{figure} 
As many datasets commonly used in Visual Geo-localization only have labels for GPS coordinates, it is often impossible to assess the difference between a query's heading and its predictions.
While one might assume that a positive prediction is likely to have the same heading as its query, this might not always be the case, as in cities it is possible to find places that are self-similar in multiple directions (\eg buildings facing each other with similar architectures).
Moreover, two images representing the same scene might be taken from a very different viewpoint, within 25 meters from each other.
To shed some light on this question, we compute recall@1 on the San Francisco dataset, for which heading labels are available, considering positives with a variable difference in heading from the query (Fig. \ref{fig:heading_sf}). The distance threshold is fixed at 25 meters.

\myparagraph{Discussion}
We can see from Fig. \ref{fig:heading_sf} that roughly all (99.8\%) correct predictions' heading are within 90° from the query's heading, 98\% are within 60°, 90\% within 45°, 57\% within 30°, and only roughly 14\% are within 10°.
Moreover, we see that these results are pretty stable across all models.
The figure clearly shows that post-processing techniques must be considered when an accurate pose estimation is needed.

\subsubsection{Changing the positives' threshold distance}
\label{sec:supp_threshold_distance}
\begin{figure*}[ht!]
    \centering
    \begin{minipage}{.32\textwidth}
        \begin{subfigure}{\textwidth}
            \includegraphics[width=\textwidth]{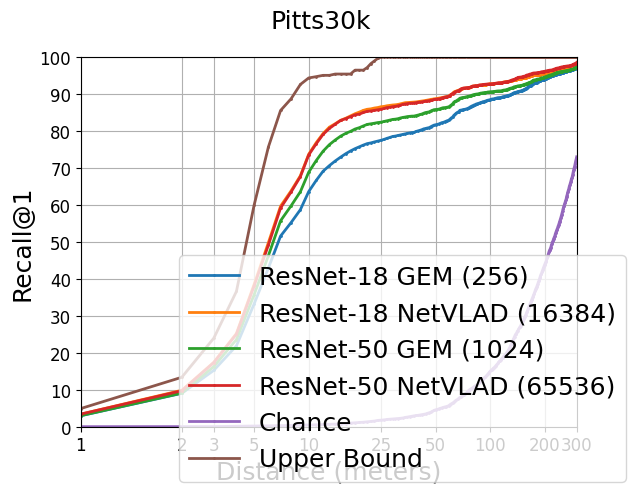}
        \end{subfigure}
        \begin{subfigure}{\textwidth}
            \includegraphics[width=\textwidth]{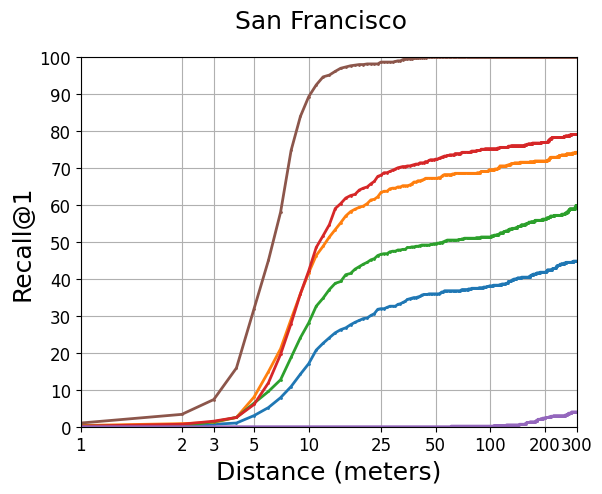} 
        \end{subfigure}
    \end{minipage}
    \begin{minipage}{.32\textwidth}
        \begin{subfigure}{\textwidth}
            \includegraphics[width=\textwidth]{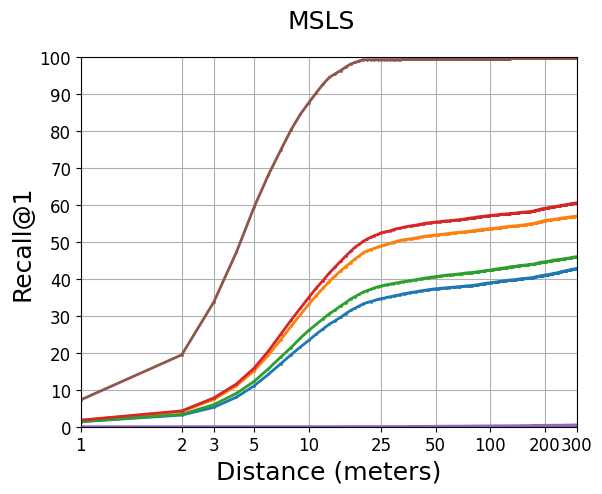} 
        \end{subfigure}
        \begin{subfigure}{\textwidth}
        \includegraphics[width=\textwidth]{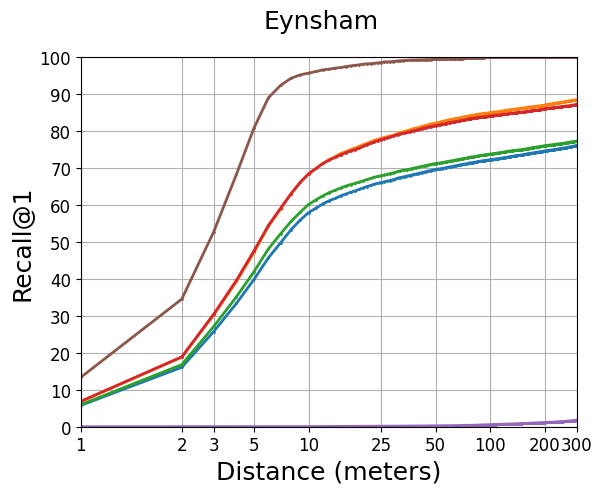} 
        \end{subfigure}
    \end{minipage}
    \begin{minipage}{.32\textwidth}
        \begin{subfigure}{\textwidth}
            \includegraphics[width=\textwidth]{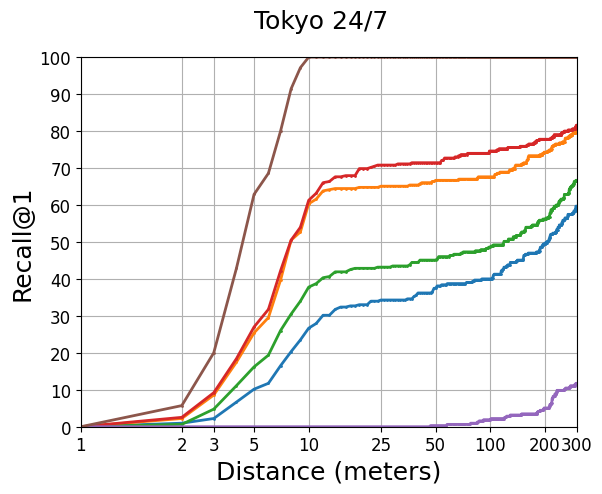} 
        \end{subfigure}
        \begin{subfigure}{\textwidth}
            \includegraphics[width=\textwidth]{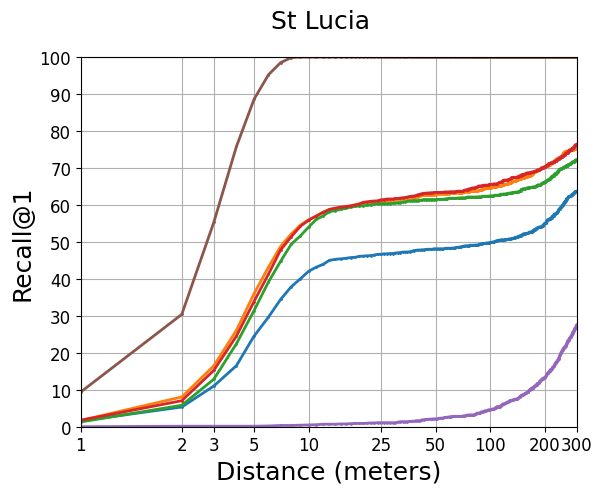} 
        \end{subfigure}
    \end{minipage}
    \caption{\textbf{Changing the positives’ threshold distance.} Plots showing how Recall@1 changes when changing the positives' threshold (x-axis), expressed in meters. Moreover, we also show the upper bound (some queries might not have positives within a given distance) and chance, computed by choosing random predictions for each query. All models are trained on the Pitts30k dataset.}
    \label{fig:distances}
\end{figure*}

Although in most VG works \cite{Arandjelovic-2018, Kim-2017} the distance within which a database image is considered a positive is 25 meters, in the real world, one might require more or less accurate positions, depending on the task and final goal.
Results are shown in Fig. \ref{fig:distances}, where we consider thresholds from 1 to 100 meters.

\myparagraph{Discussion}
Results are consistent across all models: as the threshold distance grows, we can see a fast rise in recall@1. Depending on the dataset, this rapid growth slows down somewhere between 10 and 25 meters. Recall@1 does not reach 90\% for any dataset, even as the threshold grows as high as 50 meters.
\\
The plots also give interesting insights into the datasets: looking at the Upper Bound line it is possible to understand which datasets are denser than others.
For example, St Lucia has an upper bound of 88\% at 5 meters, meaning that for 88\% of the queries there is at least one positive within 5 a meters threshold.
From the plots we see that a distance of 25 meters provides a reliable threshold for evaluation on all the considered datasets, as it ensures that close to 100\% of queries have a relevant database image, and that random chance leads to recalls close to zero.

\subsubsection{Other values of recall}
\label{sec:supp_other_recalls}
\begin{figure*}[th!]
    \centering
    \begin{minipage}{.32\textwidth}
        \begin{subfigure}{\textwidth}
            \includegraphics[width=\textwidth]{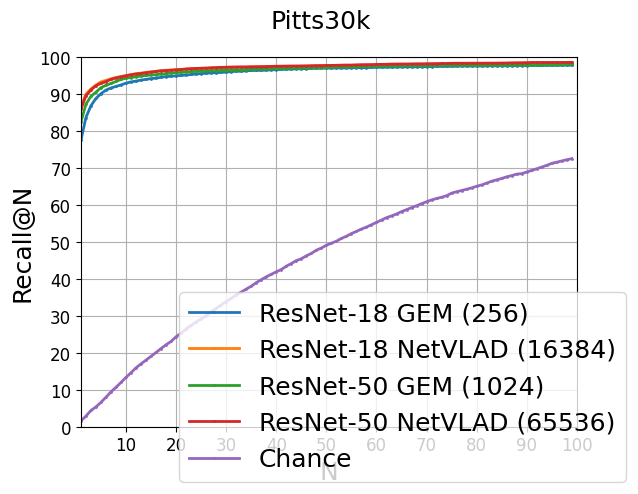}
        \end{subfigure}
        \begin{subfigure}{\textwidth}
            \includegraphics[width=\textwidth]{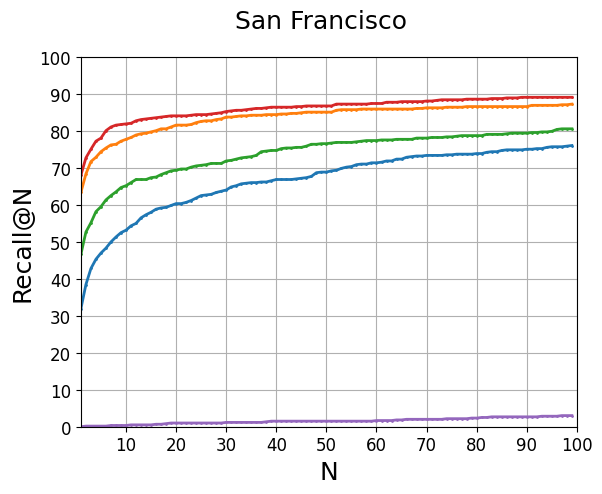}
            \subcaption{}
        \end{subfigure}
    \end{minipage}
    \begin{minipage}{.32\textwidth}
        \begin{subfigure}{\textwidth}
            \includegraphics[width=\textwidth]{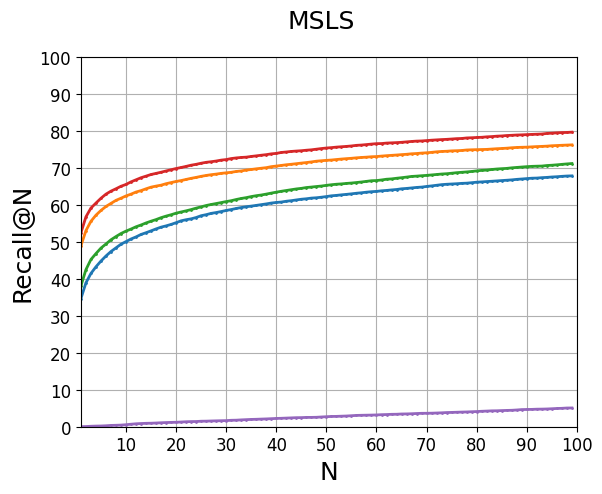}
        \end{subfigure}
        \begin{subfigure}{\textwidth}
            \includegraphics[width=\textwidth]{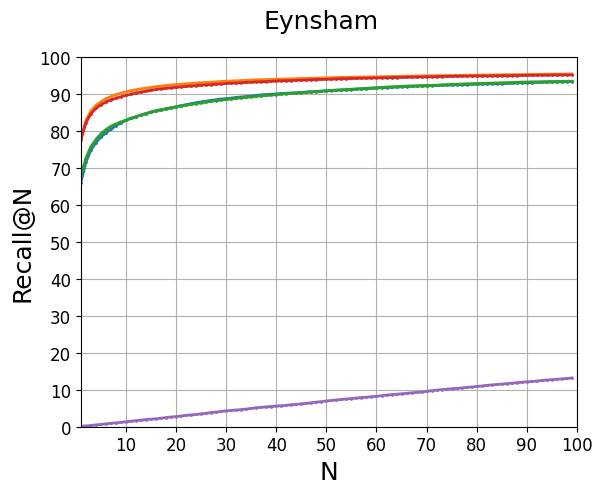}
            \subcaption{}
        \end{subfigure}
    \end{minipage}
    \begin{minipage}{.32\textwidth}
        \begin{subfigure}{\textwidth}
            \includegraphics[width=\textwidth]{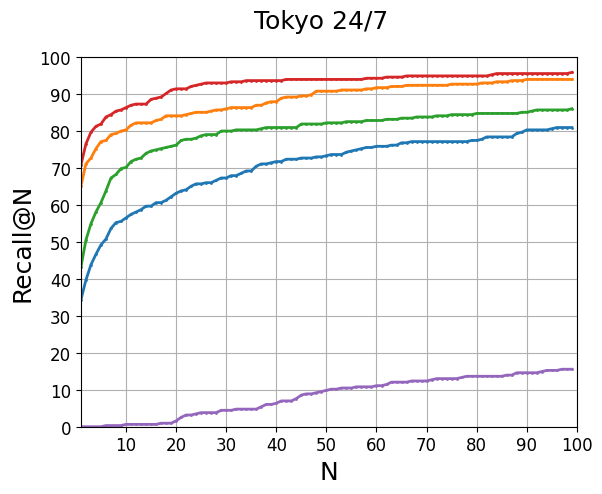}
        \end{subfigure}
        \begin{subfigure}{\textwidth}
            \includegraphics[width=\textwidth]{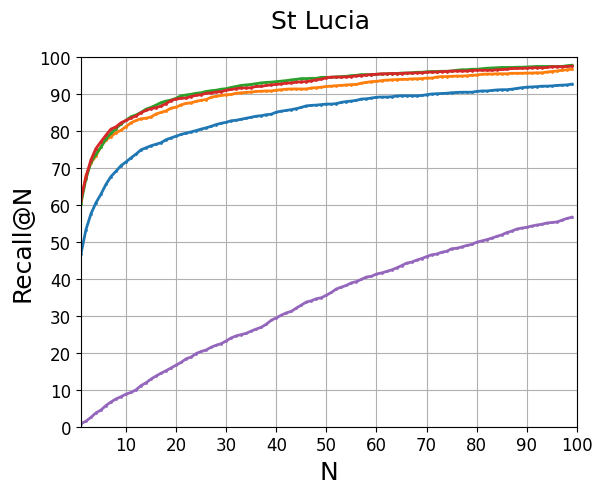}
            \subcaption{}
        \end{subfigure}
    \end{minipage}
    \caption{\textbf{Other values of recall.} Plots showing different values of recalls. On the x-axis is N $\in$ \{1, 2, 3... 100\}, and on the y-axis the recall N.}
    \label{fig:recallsN}
\end{figure*}

In this section, we experiment using other values of N for the recall@N.
Plots with recalls up to 100 are shown in Fig. \ref{fig:recallsN}.

\begin{figure*}[ht!]
    \centering
    \begin{minipage}{.32\textwidth}
        \begin{subfigure}{\textwidth}
            \includegraphics[width=\textwidth]{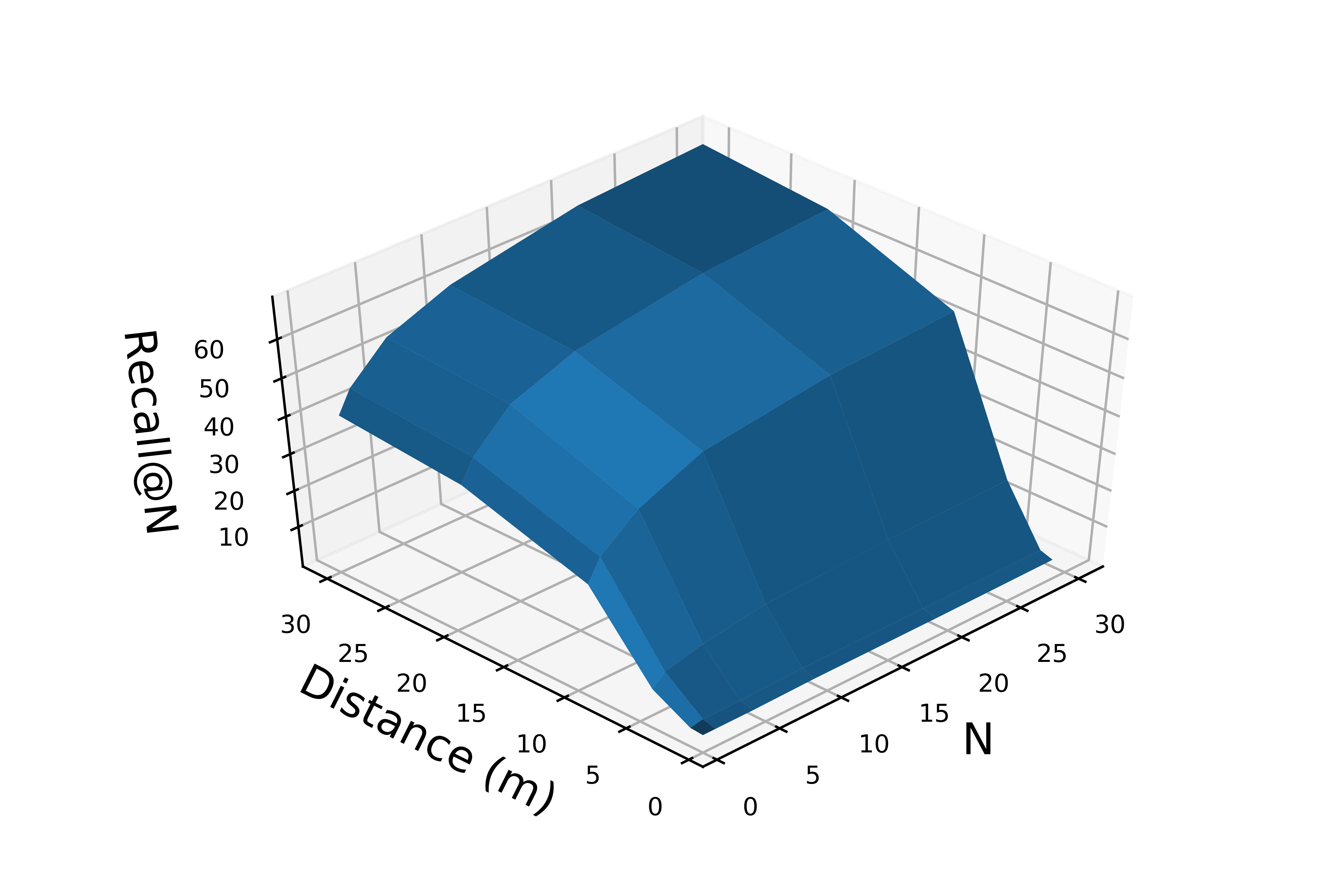}
            \subcaption{}
            \end{subfigure}
    \end{minipage}
    \begin{minipage}{.32\textwidth}
        \begin{subfigure}{\textwidth}
            \includegraphics[width=\textwidth]{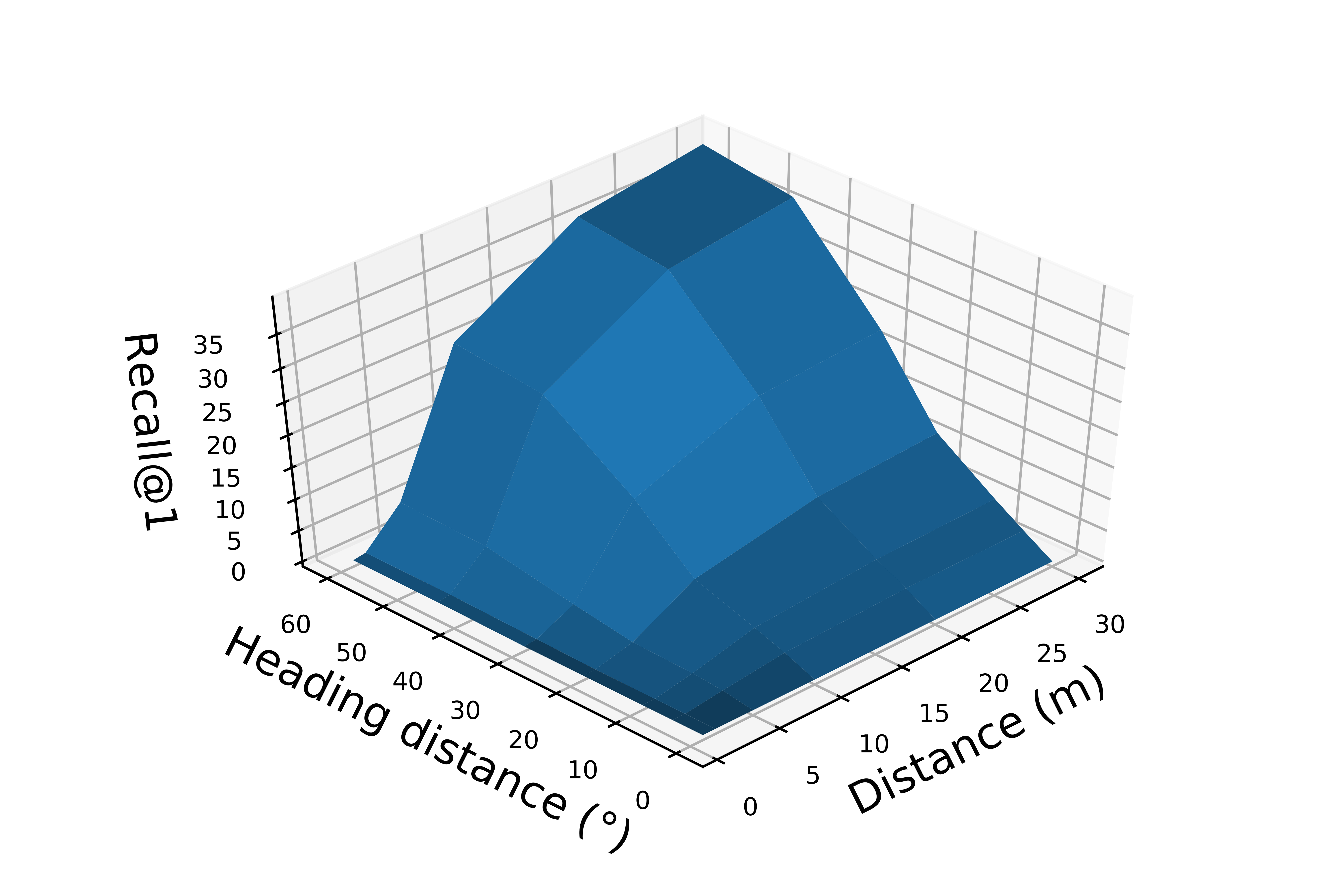}
            \subcaption{}
        \end{subfigure}
    \end{minipage}
    \begin{minipage}{.32\textwidth}
        \begin{subfigure}{\textwidth}
            \includegraphics[width=\textwidth]{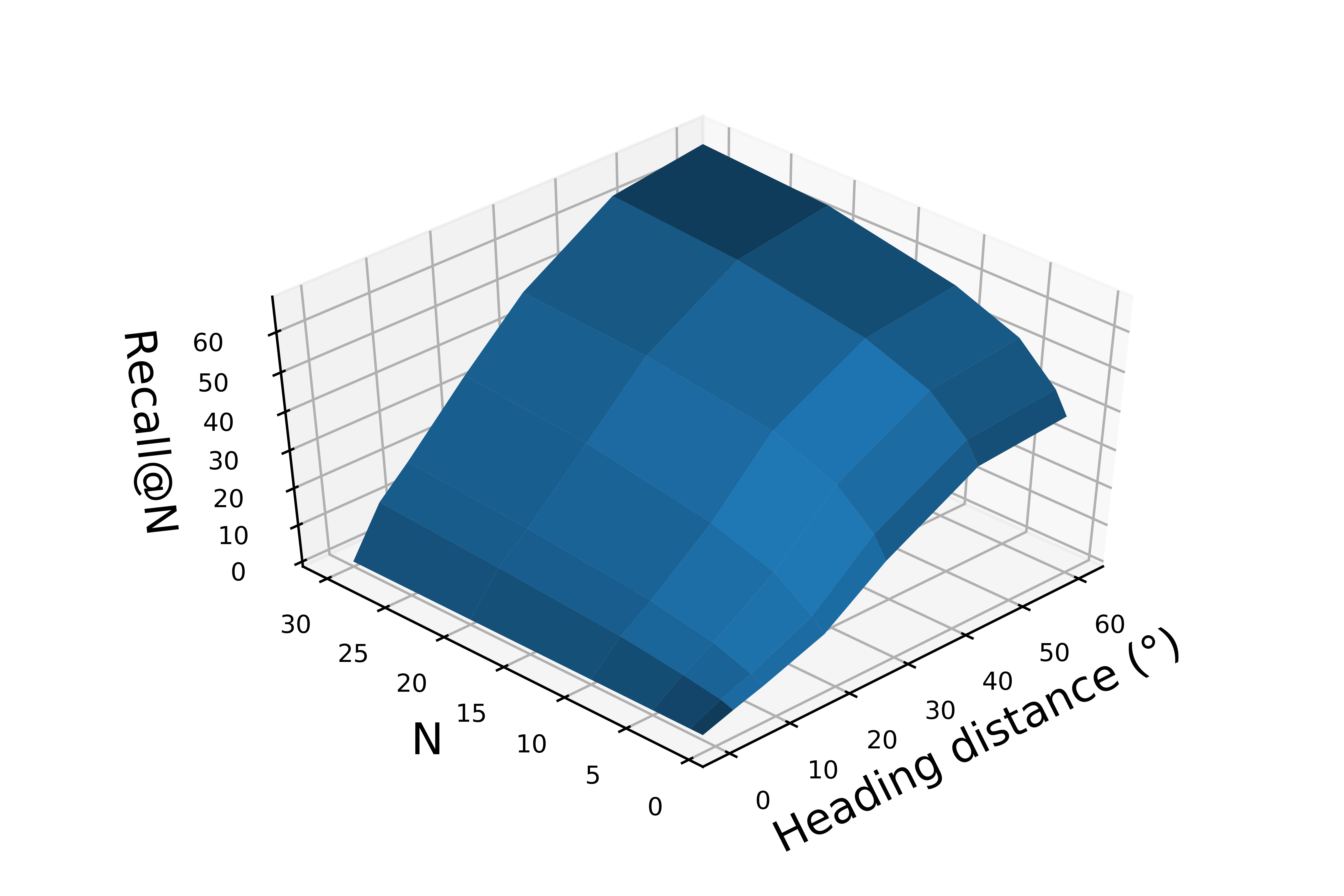}
            \subcaption{}
        \end{subfigure}
    \end{minipage}
    \caption{\textbf{The relation between threshold distances, values of the recall and heading distance.} These 3D plots show how the three factors interact with each others. a) shows recall@N while changing the distances and values of the recall (N), and keeping heading distance at 180°; b) shows recall@1 varying heading angle and distance; c) keeping threshold distance at 25 meters, and varying the two other factors, shows the recall@N.}
    \label{fig:3d_plots}
    \vspace{-0.2cm}
\end{figure*}

\myparagraph{Discussion}
From Fig. \ref{fig:recallsN} we can extract interesting insights:
we see that if a query's location is not found within 5 predictions, chances are rather low (35\% on average over all methods on all datasets) that it is found within 20 predictions.
This number is even lower for more challenging datasets: 19\% for MSLS, 26\% for San Francisco.
Similarly, if a query's location is not found within one prediction, chances are 75\% on average that it is found within 100 (53\% for MSLS, 64\% for San Francisco).
\\
Moreover, the plot shows the upper bound for re-ranking methods like \cite{Hausler-2021, Sarlin-2020}: these methods compute re-ranking over a limited number of predictions (usually 100) as the time complexity grows linearly with such number.
For example, if 100 predictions are considered for re-ranking, the resulting recall@1 cannot be higher than the initial recall@100. These plots suggest that re-ranking over the top 20/30 predictions would give a similar performance at a much lower cost.

\subsubsection{The relation between threshold distances, values of the recall and heading distance}
\label{sec:supp_3d_plots}
In previous sections, we display results while changing one of the three factors between threshold distances, values of the recall, and heading distance at a time, here we investigate the relationships between any given pair of them. 
Fig. \ref{fig:3d_plots} shows how these factors interact with each other.
To compute the results, we used a ResNet-50 + GeM trained on Pitts30k, and the recalls in the plots refer to the R-SF dataset (as it is the only one with heading labels).
\section{Ethical implications}
The technology of Visual Geo-localization can potentially be used to implement invasive forms of surveillance or social media monitoring, thus raising privacy concerns.
The benchmark proposed in this manuscript is just a tool whose purpose is to provide a systematic and standardized approach to testing and comparing different Visual Geo-localization algorithms. As such, it cannot offer guarantees on the final use of the algorithms that it will help to evaluate.
Therefore, we urge all researchers using this tool to be mindful of the potential misuses of their algorithms.
For what concerns data, the framework relies only on pre-existing and publicly available datasets that are widely used in the community and focus on places rather than humans, and thus are considered safe to use.

\end{document}